\documentclass{article}

\PassOptionsToPackage{numbers, compress}{natbib}

\usepackage[final]{neurips_data_2022}

\usepackage[utf8]{inputenc} %
\usepackage[T1]{fontenc}    %
\usepackage{hyperref}       %
\usepackage{url}            %
\usepackage{booktabs}       %
\usepackage{amsfonts}       %
\usepackage{nicefrac}       %
\usepackage{microtype}      %
\usepackage{xcolor}         %

\usepackage{subcaption}
\usepackage{graphicx}
\usepackage{amsmath}
\usepackage{xcolor}
\usepackage{makecell}
\usepackage{comment}
\usepackage{placeins}
\usepackage{longtable}

\usepackage{caption}
\usepackage{subcaption}

\newcommand{\twoPicSideBySide}[4]{
\begin{minipage}{.54\textwidth}
  \centering
  \includegraphics[width=.99\linewidth]{#1}
  \captionof{figure}{#2}
  \label{fig:#1}
\end{minipage}%
\hfill
\begin{minipage}{.44\textwidth}
  \centering
  \includegraphics[width=.99\linewidth]{#3}
  \captionof{figure}{#4}
  \label{fig:#3}
\end{minipage}
}

\newcommand{\picTwo}[4]{
	\begin{figure*}[tb]
		\centering
		\begin{subfigure}[c]{0.46\textwidth}
			\centering
			\includegraphics[width=0.9\textwidth]{#1-0}
			\subcaption{#2}
			\label{fig:#1-0}
		\end{subfigure}
		\begin{subfigure}[c]{0.52\textwidth}
			\centering
			\includegraphics[width=0.9\textwidth]{#1-1}
			\subcaption{#3}
			\label{fig:#1-1}
		\end{subfigure}
		\caption{#4}
		\label{fig:#1}
	\end{figure*}
}

\newcommand{\picTwoExtend}[4]{
	\begin{figure*}[tb]
		\centering
		\begin{subfigure}[c]{0.66\textwidth}
			\centering
			\includegraphics[width=0.9\textwidth]{#1-0}
			\subcaption{#2}
			\label{fig:#1-0}
		\end{subfigure}
		\begin{subfigure}[c]{0.33\textwidth}
			\centering
			\includegraphics[width=0.9\textwidth]{#1-1}
			\subcaption{#3}
			\label{fig:#1-1}
		\end{subfigure}
		\caption{#4}
		\label{fig:#1}
	\end{figure*}
}

\newcommand{\picThree}[5]{
	\begin{figure*}[tb]
		\centering
		\begin{subfigure}[c]{0.325\textwidth}
			\centering
			\includegraphics[width=0.9\textwidth]{#1-0}
			\subcaption{#2}
				\label{fig:#1-0}
		\end{subfigure}
		\begin{subfigure}[c]{0.325\textwidth}
			\centering
			\includegraphics[width=0.9\textwidth]{#1-1}
			\subcaption{#3}
				\label{fig:#1-1}
		\end{subfigure}
		\begin{subfigure}[c]{0.325\textwidth}
			\centering
			\includegraphics[width=0.9\textwidth]{#1-2}
			\subcaption{#4}
				\label{fig:#1-2}
		\end{subfigure}

		\caption{#5}
		\label{fig:#1}
	\end{figure*}
}

\newcommand{\picThreePlus}[5]{
	\begin{figure*}[tb]
		\centering
		\begin{subfigure}[c]{0.95\textwidth}
			\centering
			\includegraphics[width=0.9\textwidth]{#1-legend}
			\label{fig:#1-legend}
		\end{subfigure}
		\begin{subfigure}[c]{0.325\textwidth}
			\centering
			\includegraphics[width=0.9\textwidth]{#1-0}
			\subcaption{#2}
				\label{fig:#1-0}
		\end{subfigure}
		\begin{subfigure}[c]{0.325\textwidth}
			\centering
			\includegraphics[width=0.9\textwidth]{#1-1}
			\subcaption{#3}
				\label{fig:#1-1}
		\end{subfigure}
		\begin{subfigure}[c]{0.325\textwidth}
			\centering
			\includegraphics[width=0.9\textwidth]{#1-2}
			\subcaption{#4}
				\label{fig:#1-2}
		\end{subfigure}

		\caption{#5}
		\label{fig:#1}
	\end{figure*}
}

\newcommand{\picFourteenMice}[2]{
	\begin{figure*}[htb]
		\begin{subfigure}[c]{0.13\textwidth}
		    \centering
			\includegraphics[width=\textwidth]{#1-1}
				\subcaption{g, fibers \newline \newline }
		\end{subfigure}
		\begin{subfigure}[c]{0.13\textwidth}
			\includegraphics[width=\textwidth]{#1-2}
			\subcaption{g, large fibers \newline }
		\end{subfigure}
		\begin{subfigure}[c]{0.13\textwidth}
			\includegraphics[width=\textwidth]{#1-3}
				\subcaption{g ug, direction is detectable}
		\end{subfigure}
		\begin{subfigure}[c]{0.13\textwidth}
			\includegraphics[width=\textwidth]{#1-4}
			\subcaption{g, ug, interconnected fibers}
		\end{subfigure}
		\begin{subfigure}[c]{0.13\textwidth}
			\includegraphics[width=\textwidth]{#1-5}
				\subcaption{nr, directional plane \newline }
		\end{subfigure}
		\begin{subfigure}[c]{0.13\textwidth}
			\includegraphics[width=\textwidth]{#1-6}
				\subcaption{nr, noise \newline \newline }
		\end{subfigure}
		\begin{subfigure}[c]{0.13\textwidth}
			\includegraphics[width=\textwidth]{#1-7}
				\subcaption{ug rough fiber \newline }
		\end{subfigure}
		\begin{subfigure}[c]{0.13\textwidth}
			\includegraphics[width=\textwidth]{#1-8}
				\subcaption{nr ug, rough outline mixed fivers}
		\end{subfigure}\hfill%
		\begin{subfigure}[c]{0.13\textwidth}
			\includegraphics[width=\textwidth]{#1-9}
				\subcaption{ug, because of side \newline }
		\end{subfigure}\hfill%
		\begin{subfigure}[c]{0.13\textwidth}
			\includegraphics[width=\textwidth]{#1-10}
				\subcaption{ug, fizzy fibers \newline }
		\end{subfigure}\hfill%
		\begin{subfigure}[c]{0.13\textwidth}
			\includegraphics[width=\textwidth]{#1-11}
				\subcaption{ug, many crossed fibers \newline }
		\end{subfigure}\hfill%
		\begin{subfigure}[c]{0.13\textwidth}
			\includegraphics[width=\textwidth]{#1-12}
				\subcaption{ug nr \newline \newline }
		\end{subfigure}\hfill%
		\begin{subfigure}[c]{0.13\textwidth}
			\includegraphics[width=\textwidth]{#1-13}
				\subcaption{ug nr, rough fibers \newline }
		\end{subfigure}\hfill%
		\begin{subfigure}[c]{0.13\textwidth}
			\includegraphics[width=\textwidth]{#1-14}
				\subcaption{g, dominant direction \newline }
		\end{subfigure}\hfill%

		\caption{#2}
		\label{fig:#1}

	\end{figure*}
}

\usepackage{tabularx}
\usepackage{multirow}
\usepackage{booktabs}  
\usepackage{colortbl}  
\definecolor{LightGray}{rgb}{0.95,0.95,0.95}

\newcommand{\tblDatasets}{
\begin{table}[tb]
	\caption{
		Overview of the used datasets -- 
		\# is an abbreviation for number.
		The class imbalance is given as the percentage of the smallest and largest class with regard to the complete dataset.
		The agreement is the percentage of annotations that agree with the majority vote.
		The scores ACC and $\hat{ACC}$ are given for the supervised baseline across three test folds.
		The access describes if the (raw) data is available openly, requires permission (restricted) or was not previously available (N/A). 
		In the last column, datasets with modifications to the original data are marked with X. 
		A modification might be adding more annotations or crop images to a region of interest.
	}
	\label{tbl:datasets}
	\resizebox{\textwidth}{!}{
		\begin{tabular}{l c c c c c c c c c c c c c}
			\toprule
			Name & \# classes & Input size $\left[ px \right]$ & \# Images & \multicolumn{2}{c}{Class Imbalance $\left[ \% \right]$ } & Agreement $\left[ \% \right]$ & \# Annotations & ACC $\left[ \% \right]$ & $\hat{ACC}$ $\left[ \% \right]$ & Access & Updated\\

			\cmidrule(r){5-6}
			
			& & &  &  Smallest & Largest  & Mean $\pm$ STD & Mean $\pm$ STD & Mean $\pm$ STD & Mean $\pm$ STD \\
			
			\midrule
			
		    Benthic & 10 & 112$\times$112 & 4867 & 2.31 & 39.66 & 82.61 $\pm$ 19.67 & 4.54 $\pm$ 2.01 & 64.17 $\pm$ 0.63 & 83.36 $\pm$ 0.47 & Restricted & X\\
			CIFAR-10H & 10 & 32$\times$32 & 10000 & 9.88 & 10.16  & 95.44 $\pm$ 8.91 & 51.10 $\pm$ 1.54  & 90.75 $\pm$ 0.39 & 95.72 $\pm$ 0.12 & Open & \\
			Mice Bone & 3 & 224$\times$224 & 7240 & 14.75 & 70.48 & 85.06 $\pm$ 17.52 & 15.30 $\pm$ 21.90 & 61.88 $\pm$ 9.44 & 78.39 $\pm$ 1.95 & Restricted & X\\
			Pig & 4 & 96$\times$96 & 10237 &  7.82 & 41.23 & 65.32 $\pm$ 19.50 & 7.26 $\pm$ 2.29 & 35.97 $\pm$ 3.61 & 64.77 $\pm$ 0.79 & N/A & X   \\
			Plankton & 10 & 96$\times$96 & 12280 & 4.16 & 30.37 & 93.26 $\pm$ 13.60 & 24.38 $\pm$ 44.17 & 89.89 $\pm$ 0.82 & 92.41 $\pm$ 0.41 & Restricted \\
			Quality MRI & 2 & 224$\times$224 & 310 & 34.84 & 64.16 & 71.56 $\pm$ 12.27 & 99.94 $\pm$ 13.44  & 66.62 $\pm$ 3.55 & 75.81 $\pm$ 0.17 & Restricted & X\\
			Synthetic & 6 & 224$\times$224 & 15000 & 16.17 & 17.57 & 74.41 $\pm$ 24.28 & 98.86 $\pm$ 0.99 & 87.85 $\pm$ 0.48 & 74.65 $\pm$ 0.34 & N/A & X\\
			Treeversity\#1 & 6 & 224$\times$224 & 9489 & 9.98 &  30.67 & 88.60 $\pm$ 16.13 & 14.78 $\pm$ 7.06 & 79.50 $\pm$ 1.53 & 89.20 $\pm$ 0.31 & Open & X\\
			Treeversity\#6 & 6 & 224$\times$224 & 9826 & 8.77 & 31.26 & 66.53 $\pm$ 19.48 & 35.45 $\pm$ 11.47 & 56.71 $\pm$ 4.89 & 68.88 $\pm$ 0.72 & Open & X\\
			Turkey & 3 & 192$\times$192 & 8040 & 10.88 & 75.95 & 91.56 $\pm$ 13.82 & 14.85 $\pm$ 20.95 & 75.51 $\pm$ 2.80 & 86.89 $\pm$ 1.03 & Restricted & X\\

		\end{tabular}
	}
		
\end{table}
}

\newcommand{\tblMethods}{
\begin{table}[tb]
	\caption{
		Overview  of used methods grouped into supervised, semi-supervised and self-supervised.
		The second to fifth column describe if the method uses unlabeled data, makes noise estimation, what pretraining is the used input of the initialized dataset are hard or soft labels, respectively.
		The initialization schemes columns describe which schemes were evaluated for individual methods.
		The average runtime of the labeling phase is given in the last column.	
	}
	\label{tbl:methods}
	\resizebox{\textwidth}{!}{
		\begin{tabular}{l c c c c c c c c c c c c c c c c}
			\toprule
			Name & \multirow{2}{*}{\makecell{Unlabeled\\Data}} &  \multirow{2}{*}{\makecell{Noise\\Estimation}} & Pretraining & Labels & \multicolumn{4}{c}{Initialization Schemes} & \multirow{2}{*}{\makecell{Avg. Runtime\\$\left[ h \right]$}} \\

			\cmidrule(r){6-9}
			
			& & & & &  SL & SL+ & SSL & SSL+ &  \\
			
			\midrule
			
			Baseline & & & & Soft & X& X& X& X& 0.00 \\
    		 Heteroscedastic (Het) \cite{correlated_label_noise} & & X & & Hard &    X& X& X& X& 0.50\\
    		 SNGP \cite{liu2020sngp} & & X & & Hard &  X& X& X& X& 0.29\\
    		 ELR+ \cite{Liu2020elr} & & X & ImageNet & Hard &  X& X& X& X& 0.09\\
    		 
		    \midrule	 
	    
	    Mean-Teacher (Mean) \cite{mean-teacher} &  X &  &  & Hard  &X& & X& & 1.08\\
	    Mean-Teacher (Mean+DC3) \cite{dc3} & X &  &  & Hard & X& & X& & 1.20\\
	    $\pi$-Model ($\pi$) \cite{temporal-ensembling}& X &  &  & Hard  & X& & X& & 1.03\\
	    $\pi$-Model ($\pi$+DC3) \cite{dc3}& X &  &  & Hard  & X& & X& & 1.15\\
	    FixMatch \cite{fixmatch}& X &  &  & Hard &  X& & X& & 4.53\\
	    FixMatch +DC3 \cite{dc3}&  X &  &  & Hard &  X& & X& & 4.01\\
	    Pseudo-Label (Pseudo v1) \cite{pseudolabel} & X &  &  & Hard   & X& & X& & 1.10\\
	    Pseudo-Label (Pseudo v1 +DC3) \cite{dc3} & X &  &  & Hard  & X& & X& & 1.40\\
	    Pseudo-Label (Pseudo v2 hard) \cite{pseudolabel} & X &  & ImageNet & Hard & X& X& X& X& 0.16\\
	    Pseudo-Label (Pseudo v2 soft) \cite{pseudolabel} & X &  & ImageNet & Soft  & X&X & X& X & 0.12\\
	    Pseudo-Label (Pseudo v2 not) \cite{pseudolabel} & X &  &  & Soft & X&X & X& X & 0.12\\
	    DivideMix \cite{divide-mix} & X & X &  ImageNet & Hard  & X& X & X& X & 1.39\\
	    	\midrule
	     BYOL \cite{byol} & X &  & Self-Supervised & Hard & X& & X& & 2.59\\
	     MOCOv2 \cite{mocov2} & X &  & Self-Supervised  & Hard &  X& & X& & 7.94\\
	     SimCLR \cite{simclr} & X &  & Self-Supervised  & Hard & X& & X& & 5.89\\ 
	     SWAV \cite{swav} & X &  & Self-Supervised  & Hard & X& & X& & 4.17\\

		\end{tabular}
	}
		
\end{table}
}

\newcommand{\tblMoreAnnotationsZero}{
\begin{table*}[tb]
		\caption{
		Full table to \autoref{fig:1_more_annotations-0}
	}
	\label{tbl:more0}
	\resizebox{\textwidth}{!}{
		\begin{tabular}{l c c c c c c c c c c c c c c c c}
			\toprule
			
			\multirow{2}{*}{Budget}       & \multicolumn{2}{c}{10\%}                 & \multicolumn{2}{c}{20\%}                 & \multicolumn{2}{c}{50\%}                  & \multicolumn{2}{c}{100\%}                 & \multicolumn{2}{c}{300\%}                 & \multicolumn{2}{c}{500\%}                 & \multicolumn{2}{c}{1000\%}                \\
              \cmidrule(r){2-3} \cmidrule(r){4-5}  \cmidrule(r){6-7}  \cmidrule(r){8-9} \cmidrule(r){10-11} \cmidrule(r){12-13} \cmidrule(r){14-15}
                     
                     & Median & Mean $\pm$ SEM & Median & Mean $\pm$ SEM & Median & Mean $\pm$ SEM & Median & Mean $\pm$ SEM & Median & Mean $\pm$ SEM & Median & Mean $\pm$ SEM & Median & Mean $\pm$ SEM \\

                     \midrule
                     
$KL$                 & 0.80 & 0.91 $\pm$ 0.03        & 0.67 & 0.83 $\pm$ 0.03        & 0.53 & 0.64 $\pm$ 0.02        & 0.48 & 0.60 $\pm$ 0.02        & 0.46 & 0.55 $\pm$ 0.04        & 0.42 & 0.54 $\pm$ 0.03        & 0.45 & 0.56 $\pm$ 0.03       \\
$ECE$                & 0.12 & 0.15 $\pm$ 0.00        & 0.10 & 0.13 $\pm$ 0.00        & 0.08 & 0.11 $\pm$ 0.00        & 0.09 & 0.11 $\pm$ 0.00        & 0.08 & 0.09 $\pm$ 0.01        & 0.07 & 0.09 $\pm$ 0.00        & 0.07 & 0.09 $\pm$ 0.01       \\
$ACC$                & 0.55 & 0.56 $\pm$ 0.01        & 0.58 & 0.60 $\pm$ 0.01        & 0.65 & 0.66 $\pm$ 0.01        & 0.69 & 0.68 $\pm$ 0.01        & 0.70 & 0.70 $\pm$ 0.01        & 0.71 & 0.71 $\pm$ 0.01        & 0.72 & 0.71 $\pm$ 0.01       \\
$F1$                 & 0.52 & 0.53 $\pm$ 0.01        & 0.57 & 0.57 $\pm$ 0.01        & 0.62 & 0.63 $\pm$ 0.01        & 0.65 & 0.65 $\pm$ 0.01        & 0.69 & 0.68 $\pm$ 0.01        & 0.70 & 0.70 $\pm$ 0.01        & 0.70 & 0.70 $\pm$ 0.01       \\
$\kappa$             & 0.39 & 0.39 $\pm$ 0.01        & 0.47 & 0.45 $\pm$ 0.01        & 0.56 & 0.53 $\pm$ 0.01        & 0.66 & 0.65 $\pm$ 0.01        & 0.68 & 0.69 $\pm$ 0.01        & 0.69 & 0.70 $\pm$ 0.01        & 0.70 & 0.71 $\pm$ 0.01       \\

		\end{tabular}
	}
	
\end{table*}
}

\newcommand{\tblMoreAnnotationsOne}{
\begin{table*}[tb]
		\caption{
		Full table to \autoref{fig:1_more_annotations-1}
	}
	\label{tbl:more1}
	\resizebox{\textwidth}{!}{
		\begin{tabular}{l c c c c c c c c c c c c c c c c}
			\toprule
              \multirow{2}{*}{Budget}       &  \multicolumn{2}{c}{100\%}                 & \multicolumn{2}{c}{300\%}                 & \multicolumn{2}{c}{500\%}                 & \multicolumn{2}{c}{1000\%}                \\
              \cmidrule(r){2-3} \cmidrule(r){4-5}  \cmidrule(r){6-7}  \cmidrule(r){8-9} \cmidrule(r){10-11} \cmidrule(r){12-13} \cmidrule(r){14-15}
                     
                &     Median & Mean $\pm$ SEM & Median & Mean $\pm$ SEM & Median & Mean $\pm$ SEM & Median & Mean $\pm$ SEM \\
                     
                     \midrule
                     
Results for  $KL$ \\
Difficult datasets   & 0.00 & 0.00 $\pm$ 0.02        & -0.01 & -0.03 $\pm$ 0.06       & -0.04 & -0.04 $\pm$ 0.04       & -0.01 & -0.06 $\pm$ 0.04      \\
Easy datasets        & 0.00 & -0.00 $\pm$ 0.01       & -0.01 & -0.06 $\pm$ 0.03       & -0.02 & -0.08 $\pm$ 0.03       & -0.01 & -0.02 $\pm$ 0.02      \\
Results for  $ECE$\\
Difficult datasets   & 0.00 & 0.00 $\pm$ 0.00        & -0.02 & -0.02 $\pm$ 0.01       & -0.02 & -0.02 $\pm$ 0.01       & -0.01 & -0.02 $\pm$ 0.01      \\
Easy datasets        & 0.00 & 0.00 $\pm$ 0.00        & -0.00 & -0.01 $\pm$ 0.01       & -0.00 & -0.01 $\pm$ 0.00       & 0.00 & -0.00 $\pm$ 0.00      \\
Results for  $ACC$\\
Difficult datasets   & 0.00 & -0.00 $\pm$ 0.00       & 0.02 & 0.02 $\pm$ 0.01        & 0.03 & 0.02 $\pm$ 0.01        & 0.03 & 0.04 $\pm$ 0.01       \\
Easy datasets        & 0.00 & 0.00 $\pm$ 0.00        & 0.01 & 0.01 $\pm$ 0.01        & 0.01 & 0.02 $\pm$ 0.00        & 0.01 & 0.01 $\pm$ 0.00       \\
Results for  $F1$\\
Difficult datasets   & 0.00 & -0.00 $\pm$ 0.00       & 0.03 & 0.02 $\pm$ 0.01        & 0.04 & 0.04 $\pm$ 0.01        & 0.05 & 0.05 $\pm$ 0.01       \\
Easy datasets        & 0.00 & -0.00 $\pm$ 0.00       & 0.01 & 0.02 $\pm$ 0.01        & 0.02 & 0.03 $\pm$ 0.01        & 0.01 & 0.02 $\pm$ 0.00       \\
Results for  $\kappa$\\
Difficult datasets   & 0.00 & -0.00 $\pm$ 0.00       & 0.02 & 0.02 $\pm$ 0.01        & 0.03 & 0.04 $\pm$ 0.01        & 0.04 & 0.06 $\pm$ 0.01       \\
Easy datasets        & 0.00 & -0.00 $\pm$ 0.00       & 0.02 & 0.03 $\pm$ 0.01        & 0.03 & 0.04 $\pm$ 0.01        & 0.02 & 0.04 $\pm$ 0.01       \\

		\end{tabular}
	}
	
\end{table*}
}

\newcommand{\tblMoreAnnotationsTwo}{
\begin{table*}[tb]
		\caption{
		Full table to \autoref{fig:1_more_annotations-2}
	}
	\label{tbl:more2}
	\resizebox{\textwidth}{!}{
		\begin{tabular}{l c c c c c c c c c c c c c c c c}
			\toprule
             \multirow{2}{*}{Budget}       & \multicolumn{2}{c}{100\%}                 & \multicolumn{2}{c}{300\%}                 & \multicolumn{2}{c}{500\%}                 & \multicolumn{2}{c}{1000\%}                \\
              \cmidrule(r){2-3} \cmidrule(r){4-5}  \cmidrule(r){6-7}  \cmidrule(r){8-9} \cmidrule(r){10-11} \cmidrule(r){12-13} \cmidrule(r){14-15}
                     
                 &    Median & Mean $\pm$ SEM & Median & Mean $\pm$ SEM & Median & Mean $\pm$ SEM & Median & Mean $\pm$ SEM \\
                     
                     \midrule
                     
Results for  $KL$\\
Using soft labels    & 0.00 & -0.00 $\pm$ 0.02       & -0.11 & -0.22 $\pm$ 0.05       & -0.13 & -0.25 $\pm$ 0.04       & -0.16 & -0.23 $\pm$ 0.04      \\
Using hard labels    & 0.00 & 0.00 $\pm$ 0.01        & 0.01 & 0.06 $\pm$ 0.05        & 0.01 & 0.06 $\pm$ 0.03        & 0.02 & 0.07 $\pm$ 0.02       \\
Results for  $ECE$\\
Using soft labels    & -0.00 & -0.00 $\pm$ 0.00       & -0.02 & -0.04 $\pm$ 0.01       & -0.03 & -0.04 $\pm$ 0.01       & -0.03 & -0.04 $\pm$ 0.01      \\
Using hard labels    & 0.00 & 0.00 $\pm$ 0.00        & -0.00 & -0.00 $\pm$ 0.00       & -0.00 & -0.00 $\pm$ 0.00       & 0.00 & -0.00 $\pm$ 0.00      \\
Results for  $ACC$\\
Using soft labels    & 0.00 & 0.00 $\pm$ 0.01        & 0.02 & 0.03 $\pm$ 0.01        & 0.02 & 0.03 $\pm$ 0.01        & 0.03 & 0.03 $\pm$ 0.01       \\
Using hard labels    & 0.00 & 0.00 $\pm$ 0.00        & 0.01 & 0.00 $\pm$ 0.00        & 0.01 & 0.02 $\pm$ 0.00        & 0.01 & 0.02 $\pm$ 0.00       \\
Results for  $F1$\\
Using soft labels    & 0.00 & 0.00 $\pm$ 0.01        & 0.02 & 0.04 $\pm$ 0.01        & 0.04 & 0.04 $\pm$ 0.01        & 0.03 & 0.04 $\pm$ 0.01       \\
Using hard labels    & 0.00 & -0.00 $\pm$ 0.00       & 0.01 & 0.01 $\pm$ 0.01        & 0.02 & 0.03 $\pm$ 0.00        & 0.02 & 0.03 $\pm$ 0.01       \\
Results for  $\kappa$\\
Using soft labels    & 0.00 & 0.00 $\pm$ 0.01        & 0.01 & 0.03 $\pm$ 0.01        & 0.03 & 0.04 $\pm$ 0.01        & 0.01 & 0.04 $\pm$ 0.01       \\
Using hard labels    & 0.00 & -0.00 $\pm$ 0.00       & 0.02 & 0.02 $\pm$ 0.01        & 0.03 & 0.04 $\pm$ 0.01        & 0.04 & 0.06 $\pm$ 0.01       \\

		\end{tabular}
	}
	
\end{table*}
}

\newcommand{\tblTransQualZero}{
\begin{table*}[tb]
		\caption{
		Full table to \autoref{fig:trans_qual-0}
	}
	\label{tbl:trans_qual0}
	\resizebox{\textwidth}{!}{
		\begin{tabular}{l c c c c c c c c c c c c c c c c}
			\toprule
			
			\multirow{2}{*}{Budget}       & \multicolumn{2}{c}{10\%}                 & \multicolumn{2}{c}{20\%}                 & \multicolumn{2}{c}{50\%}                  & \multicolumn{2}{c}{100\%}                 & \multicolumn{2}{c}{300\%}                 & \multicolumn{2}{c}{500\%}                 & \multicolumn{2}{c}{1000\%}                \\
              \cmidrule(r){2-3} \cmidrule(r){4-5}  \cmidrule(r){6-7}  \cmidrule(r){8-9} \cmidrule(r){10-11} \cmidrule(r){12-13} \cmidrule(r){14-15}
                     
                     & Median & Mean $\pm$ SEM & Median & Mean $\pm$ SEM & Median & Mean $\pm$ SEM & Median & Mean $\pm$ SEM & Median & Mean $\pm$ SEM & Median & Mean $\pm$ SEM & Median & Mean $\pm$ SEM \\

                     \midrule

$KL$                 & 0.80 & 0.91 $\pm$ 0.03        & 0.67 & 0.83 $\pm$ 0.03        & 0.53 & 0.64 $\pm$ 0.02        & 0.48 & 0.60 $\pm$ 0.02        & 0.46 & 0.55 $\pm$ 0.04        & 0.42 & 0.54 $\pm$ 0.03        & 0.45 & 0.56 $\pm$ 0.03       \\
$\hat{KL}$           & 1.48 & 1.70 $\pm$ 0.05        & 1.21 & 1.52 $\pm$ 0.05        & 0.89 & 1.21 $\pm$ 0.04        & 0.74 & 1.08 $\pm$ 0.04        & 0.72 & 1.02 $\pm$ 0.06        & 0.73 & 0.96 $\pm$ 0.06        & 0.65 & 0.95 $\pm$ 0.07       \\
$ACC$                & 0.55 & 0.56 $\pm$ 0.01        & 0.58 & 0.60 $\pm$ 0.01        & 0.65 & 0.66 $\pm$ 0.01        & 0.69 & 0.68 $\pm$ 0.01        & 0.70 & 0.70 $\pm$ 0.01        & 0.71 & 0.71 $\pm$ 0.01        & 0.72 & 0.71 $\pm$ 0.01       \\
$\hat{ACC}$          & 0.54 & 0.54 $\pm$ 0.01        & 0.57 & 0.58 $\pm$ 0.01        & 0.65 & 0.63 $\pm$ 0.01        & 0.67 & 0.66 $\pm$ 0.01        & 0.70 & 0.69 $\pm$ 0.01        & 0.71 & 0.70 $\pm$ 0.01        & 0.72 & 0.71 $\pm$ 0.01       \\
$ECE$                & 0.12 & 0.15 $\pm$ 0.00        & 0.10 & 0.13 $\pm$ 0.00        & 0.08 & 0.11 $\pm$ 0.00        & 0.09 & 0.11 $\pm$ 0.00        & 0.08 & 0.09 $\pm$ 0.01        & 0.07 & 0.09 $\pm$ 0.00        & 0.07 & 0.09 $\pm$ 0.01       \\
$\hat{ECE}$          & 0.22 & 0.23 $\pm$ 0.01        & 0.19 & 0.21 $\pm$ 0.01        & 0.15 & 0.17 $\pm$ 0.01        & 0.12 & 0.15 $\pm$ 0.00        & 0.09 & 0.13 $\pm$ 0.01        & 0.08 & 0.12 $\pm$ 0.01        & 0.09 & 0.13 $\pm$ 0.01       \\

		\end{tabular}
	}
	
\end{table*}
}

\newcommand{\tblTransQualOne}{
\begin{table*}[tb]
		\caption{
		Full table to \autoref{fig:trans_qual-1}
	}
	\label{tbl:trans_qual1}
	\resizebox{\textwidth}{!}{
		\begin{tabular}{l c c c c c c c c c c c c c c c c}
			\toprule
			
			\multirow{2}{*}{Budget}       & \multicolumn{2}{c}{10-0.10}                 & \multicolumn{2}{c}{05-0.20}                 & \multicolumn{2}{c}{02-0.50}                  & \multicolumn{2}{c}{01-1.00}                     \\
			
              \cmidrule(r){2-3} \cmidrule(r){4-5}  \cmidrule(r){6-7}  \cmidrule(r){8-9}
                     
                     & Median & Mean $\pm$ SEM & Median & Mean $\pm$ SEM & Median & Mean $\pm$ SEM & Median & Mean $\pm$ SEM  \\
                     
                     \midrule
Results for  $KL$\\
Pseudo v2 hard       & 0.84 & 0.95 $\pm$ 0.12        & 0.77 & 0.98 $\pm$ 0.15        & 0.61 & 0.67 $\pm$ 0.05        & 0.44 & 0.51 $\pm$ 0.05       \\
Pseudo v2 soft       & 0.45 & 0.56 $\pm$ 0.07        & 0.40 & 0.47 $\pm$ 0.05        & 0.55 & 0.52 $\pm$ 0.05        & 0.43 & 0.52 $\pm$ 0.05       \\
Results for  $ACC$\\
Pseudo v2 hard       & 0.65 & 0.63 $\pm$ 0.03        & 0.64 & 0.64 $\pm$ 0.03        & 0.71 & 0.67 $\pm$ 0.03        & 0.75 & 0.71 $\pm$ 0.03       \\
Pseudo v2 soft       & 0.67 & 0.64 $\pm$ 0.03        & 0.66 & 0.68 $\pm$ 0.03        & 0.72 & 0.69 $\pm$ 0.02        & 0.72 & 0.71 $\pm$ 0.03       \\
Results for  $ECE$\\
Pseudo v2 hard       & 0.12 & 0.13 $\pm$ 0.02        & 0.13 & 0.14 $\pm$ 0.02        & 0.10 & 0.11 $\pm$ 0.02        & 0.08 & 0.10 $\pm$ 0.01       \\
Pseudo v2 soft       & 0.07 & 0.09 $\pm$ 0.01        & 0.04 & 0.07 $\pm$ 0.01        & 0.07 & 0.09 $\pm$ 0.01        & 0.10 & 0.10 $\pm$ 0.01       \\

		\end{tabular}
	}
	
\end{table*}
}

\newcommand{\tblTransQualTwo}{
\begin{table*}[tb]
			\caption{
		Full table to \autoref{fig:trans_qual-2}
	}
	\label{tbl:trans_qual2}
	\resizebox{\textwidth}{!}{
		\begin{tabular}{l c c c c c c c c c c c c c c c c}
			\toprule
			
			\multirow{2}{*}{Budget}       & \multicolumn{2}{c}{10\%}                 & \multicolumn{2}{c}{20\%}                 & \multicolumn{2}{c}{50\%}                  & \multicolumn{2}{c}{100\%}                 & \multicolumn{2}{c}{300\%}                 & \multicolumn{2}{c}{500\%}                 & \multicolumn{2}{c}{1000\%}                \\
              \cmidrule(r){2-3} \cmidrule(r){4-5}  \cmidrule(r){6-7}  \cmidrule(r){8-9} \cmidrule(r){10-11} \cmidrule(r){12-13} \cmidrule(r){14-15}
                     
                     & Median & Mean $\pm$ SEM & Median & Mean $\pm$ SEM & Median & Mean $\pm$ SEM & Median & Mean $\pm$ SEM & Median & Mean $\pm$ SEM & Median & Mean $\pm$ SEM & Median & Mean $\pm$ SEM \\
                     
                     \midrule
                     
Results for  $KL$\\
Baseline             & 0.83 & 1.39 $\pm$ 0.25        & 0.63 & 0.98 $\pm$ 0.17        & 0.52 & 0.70 $\pm$ 0.08        & 0.52 & 0.69 $\pm$ 0.09        & 0.34 & 0.37 $\pm$ 0.04        & 0.37 & 0.39 $\pm$ 0.04        & 0.29 & 0.33 $\pm$ 0.04       \\
Pseudo v2 soft       & 0.77 & 0.99 $\pm$ 0.20        & 0.72 & 0.90 $\pm$ 0.13        & 0.61 & 0.64 $\pm$ 0.06        & 0.43 & 0.52 $\pm$ 0.05        & 0.37 & 0.39 $\pm$ 0.04        & 0.34 & 0.36 $\pm$ 0.03        & 0.30 & 0.33 $\pm$ 0.04       \\
DivideMix            & 0.59 & 0.60 $\pm$ 0.05        & 0.53 & 0.55 $\pm$ 0.04        & 0.45 & 0.51 $\pm$ 0.04        & 0.44 & 0.52 $\pm$ 0.04        & 0.47 & 0.56 $\pm$ 0.06        & 0.41 & 0.54 $\pm$ 0.06        & 0.45 & 0.64 $\pm$ 0.08       \\
ELR+                 & 0.78 & 0.80 $\pm$ 0.07        & 0.57 & 0.59 $\pm$ 0.06        & 0.45 & 0.51 $\pm$ 0.07        & 0.47 & 0.52 $\pm$ 0.08        & 0.48 & 0.55 $\pm$ 0.09        & 0.44 & 0.70 $\pm$ 0.15        & 0.50 & 0.57 $\pm$ 0.06       \\
Results for  $ACC$\\
Baseline             & 0.52 & 0.57 $\pm$ 0.04        & 0.60 & 0.63 $\pm$ 0.03        & 0.65 & 0.68 $\pm$ 0.03        & 0.71 & 0.71 $\pm$ 0.03        & 0.72 & 0.74 $\pm$ 0.03        & 0.75 & 0.74 $\pm$ 0.03        & 0.75 & 0.74 $\pm$ 0.03       \\
Pseudo v2 soft       & 0.61 & 0.60 $\pm$ 0.02        & 0.63 & 0.63 $\pm$ 0.03        & 0.69 & 0.68 $\pm$ 0.03        & 0.72 & 0.71 $\pm$ 0.03        & 0.73 & 0.74 $\pm$ 0.02        & 0.77 & 0.74 $\pm$ 0.03        & 0.78 & 0.75 $\pm$ 0.03       \\
DivideMix            & 0.62 & 0.62 $\pm$ 0.03        & 0.68 & 0.66 $\pm$ 0.03        & 0.72 & 0.72 $\pm$ 0.03        & 0.73 & 0.73 $\pm$ 0.03        & 0.74 & 0.74 $\pm$ 0.03        & 0.76 & 0.75 $\pm$ 0.03        & 0.75 & 0.74 $\pm$ 0.03       \\
ELR+                 & 0.53 & 0.53 $\pm$ 0.04        & 0.59 & 0.60 $\pm$ 0.04        & 0.66 & 0.65 $\pm$ 0.04        & 0.72 & 0.68 $\pm$ 0.04        & 0.73 & 0.67 $\pm$ 0.04        & 0.72 & 0.68 $\pm$ 0.04        & 0.73 & 0.69 $\pm$ 0.04       \\
Results for  $ECE$ \\
Baseline             & 0.16 & 0.19 $\pm$ 0.03        & 0.14 & 0.17 $\pm$ 0.02        & 0.12 & 0.14 $\pm$ 0.02        & 0.13 & 0.13 $\pm$ 0.01        & 0.04 & 0.07 $\pm$ 0.01        & 0.05 & 0.07 $\pm$ 0.01        & 0.03 & 0.07 $\pm$ 0.01       \\
Pseudo v2 soft       & 0.11 & 0.14 $\pm$ 0.02        & 0.11 & 0.14 $\pm$ 0.02        & 0.11 & 0.11 $\pm$ 0.01        & 0.10 & 0.10 $\pm$ 0.01        & 0.06 & 0.08 $\pm$ 0.01        & 0.06 & 0.08 $\pm$ 0.01        & 0.05 & 0.08 $\pm$ 0.01       \\
DivideMix            & 0.12 & 0.12 $\pm$ 0.01        & 0.09 & 0.11 $\pm$ 0.01        & 0.08 & 0.10 $\pm$ 0.01        & 0.06 & 0.09 $\pm$ 0.01        & 0.06 & 0.09 $\pm$ 0.01        & 0.05 & 0.09 $\pm$ 0.02        & 0.05 & 0.09 $\pm$ 0.02       \\
ELR+                 & 0.09 & 0.15 $\pm$ 0.02        & 0.07 & 0.10 $\pm$ 0.02        & 0.06 & 0.09 $\pm$ 0.01        & 0.05 & 0.09 $\pm$ 0.02        & 0.05 & 0.09 $\pm$ 0.02        & 0.05 & 0.09 $\pm$ 0.02        & 0.05 & 0.09 $\pm$ 0.02       \\

		\end{tabular}
	}

\end{table*}
}

\newcommand{\tblImprovementAcc}{
\begin{table*}[tb]
		\caption{
		Improvements in ACC across methods
	}
	\label{tbl:improve_acc}
	\resizebox{\textwidth}{!}{
		\begin{tabular}{l c c c c c c c c c c c c c c c c}
			\toprule
			
			\multirow{2}{*}{Budget}       & \multicolumn{2}{c}{10\%}                 &  \multicolumn{2}{c}{100\%}                 &  \multicolumn{2}{c}{1000\%}                \\
              \cmidrule(r){2-3} \cmidrule(r){4-5}  \cmidrule(r){6-7}  
                     
                     & Median & Mean $\pm$ SEM & Median & Mean $\pm$ SEM & Median & Mean $\pm$ SEM  \\
                     
                     \midrule
                     
Baseline             & 0.00 & 0.00 $\pm$ 0.01        & 0.00 & 0.00 $\pm$ 0.01        & -0.00 & -0.00 $\pm$ 0.01      \\
ELR+                 & -0.02 & -0.03 $\pm$ 0.01       & 0.01 & -0.03 $\pm$ 0.03       & -0.01 & -0.05 $\pm$ 0.02      \\
Het                  & -0.06 & -0.10 $\pm$ 0.02       & -0.00 & -0.00 $\pm$ 0.01       & -0.00 & -0.01 $\pm$ 0.01      \\
SGNP                 & -0.03 & -0.05 $\pm$ 0.02       & -0.01 & -0.01 $\pm$ 0.01       & -0.01 & -0.01 $\pm$ 0.01      \\
DivideMix            & 0.05 & 0.05 $\pm$ 0.02        & 0.03 & 0.02 $\pm$ 0.02        & 0.00 & -0.00 $\pm$ 0.01      \\
Fixmatch             & 0.05 & 0.05 $\pm$ 0.01        & 0.01 & 0.01 $\pm$ 0.01        & N/A                 \\
Fixmatch + DC3      & 0.03 & 0.02 $\pm$ 0.01        & -0.00 & -0.00 $\pm$ 0.01       & N/A                 \\
Mean                 & 0.02 & 0.03 $\pm$ 0.02        & 0.02 & -0.01 $\pm$ 0.02       & N/A                 \\
Mean+DC3            & 0.05 & 0.03 $\pm$ 0.02        & 0.00 & 0.00 $\pm$ 0.01        & N/A                 \\
$\pi$                & 0.05 & 0.04 $\pm$ 0.02        & 0.02 & 0.01 $\pm$ 0.01        & N/A                 \\
$\pi$+DC3           & 0.04 & 0.03 $\pm$ 0.02        & 0.02 & 0.02 $\pm$ 0.01        & N/A                 \\
Pseudo v1            & 0.07 & 0.03 $\pm$ 0.02        & 0.01 & 0.03 $\pm$ 0.01        & N/A                 \\
Pseudo v1 + DC3     & 0.03 & 0.02 $\pm$ 0.02        & 0.02 & 0.03 $\pm$ 0.01        & N/A                 \\
Pseudo v2 hard       & 0.05 & 0.04 $\pm$ 0.03        & 0.01 & 0.00 $\pm$ 0.01        & -0.03 & -0.02 $\pm$ 0.02      \\
Pseudo v2 soft       & 0.01 & 0.04 $\pm$ 0.03        & -0.00 & -0.00 $\pm$ 0.01       & -0.00 & 0.01 $\pm$ 0.01       \\
Pseudo v2 not        & -0.18 & -0.18 $\pm$ 0.04       & -0.07 & -0.14 $\pm$ 0.03       & -0.12 & -0.17 $\pm$ 0.03      \\
BYOL                 & -0.05 & -0.06 $\pm$ 0.04       & -0.11 & -0.14 $\pm$ 0.03       & N/A                 \\
MOCOv2               & -0.04 & -0.05 $\pm$ 0.02       & -0.10 & -0.12 $\pm$ 0.02       & N/A                 \\
SimCLR               & 0.01 & 0.01 $\pm$ 0.01        & -0.04 & -0.04 $\pm$ 0.01       & N/A                 \\
SWAV                 & -0.04 & -0.06 $\pm$ 0.02       & -0.12 & -0.13 $\pm$ 0.02       & N/A                 \\

		\end{tabular}
	}
	
\end{table*}
}

\newcommand{\tblImprovementKl}{
\begin{table*}[tb]
		\caption{
		Improvements in KL across methods
	}
	\label{tbl:improve_kl}
	\resizebox{\textwidth}{!}{
		\begin{tabular}{l c c c c c c c c c c c c c c c c}
			\toprule
			
			\multirow{2}{*}{Budget}       & \multicolumn{2}{c}{10\%}                 &  \multicolumn{2}{c}{100\%}                 &  \multicolumn{2}{c}{1000\%}                \\
              \cmidrule(r){2-3} \cmidrule(r){4-5}  \cmidrule(r){6-7}  
                     
                     & Median & Mean $\pm$ SEM & Median & Mean $\pm$ SEM & Median & Mean $\pm$ SEM  \\
                     
                     \midrule
                     
Baseline             & -0.00 & -0.00 $\pm$ 0.07       & 0.00 & -0.00 $\pm$ 0.03       & 0.00 & 0.00 $\pm$ 0.01       \\
ELR+                 & -0.16 & -0.59 $\pm$ 0.22       & -0.13 & -0.17 $\pm$ 0.06       & 0.15 & 0.24 $\pm$ 0.06       \\
Het                  & -0.06 & -0.02 $\pm$ 0.08       & -0.01 & 0.01 $\pm$ 0.02        & 0.22 & 0.30 $\pm$ 0.05       \\
SGNP                 & -0.26 & -0.59 $\pm$ 0.27       & 0.00 & -0.15 $\pm$ 0.08       & 0.20 & 0.26 $\pm$ 0.04       \\
DivideMix            & -0.21 & -0.78 $\pm$ 0.25       & -0.03 & -0.17 $\pm$ 0.08       & 0.16 & 0.31 $\pm$ 0.07       \\
Fixmatch             & -0.22 & -0.43 $\pm$ 0.14       & -0.05 & -0.17 $\pm$ 0.05       & N/A                 \\
Fixmatch + DC3      & -0.29 & -0.48 $\pm$ 0.17       & -0.06 & -0.13 $\pm$ 0.04       & N/A                 \\
Mean                 & -0.14 & -0.59 $\pm$ 0.22       & -0.12 & -0.22 $\pm$ 0.08       & N/A                 \\
Mean+DC3            & -0.20 & -0.51 $\pm$ 0.17       & -0.08 & -0.15 $\pm$ 0.07       & N/A                 \\
$\pi$                & -0.33 & -0.63 $\pm$ 0.19       & -0.06 & -0.17 $\pm$ 0.05       & N/A                 \\
$\pi$+DC3           & -0.32 & -0.62 $\pm$ 0.20       & -0.10 & -0.22 $\pm$ 0.06       & N/A                 \\
Pseudo v1            & -0.25 & -0.47 $\pm$ 0.19       & -0.07 & -0.17 $\pm$ 0.04       & N/A                 \\
Pseudo v1 + DC3     & -0.24 & -0.56 $\pm$ 0.22       & -0.07 & 0.05 $\pm$ 0.23        & N/A                 \\
Pseudo v2 hard       & -0.30 & -0.39 $\pm$ 0.16       & -0.03 & -0.19 $\pm$ 0.08       & 0.19 & 0.28 $\pm$ 0.04       \\
Pseudo v2 soft       & -0.28 & -0.40 $\pm$ 0.19       & -0.04 & -0.17 $\pm$ 0.06       & 0.01 & 0.00 $\pm$ 0.02       \\
Pseudo v2 not        & 0.12 & -0.11 $\pm$ 0.34       & 0.10 & 0.22 $\pm$ 0.19        & 0.28 & 0.46 $\pm$ 0.13       \\
BYOL                 & -0.20 & -0.47 $\pm$ 0.21       & -0.08 & 0.02 $\pm$ 0.09        & N/A                 \\
MOCOv2               & -0.29 & -0.63 $\pm$ 0.22       & -0.08 & -0.13 $\pm$ 0.10       & N/A                 \\
SimCLR               & -0.07 & -0.40 $\pm$ 0.22       & 0.01 & 0.04 $\pm$ 0.06        & N/A                 \\
SWAV                 & -0.05 & -0.43 $\pm$ 0.22       & 0.19 & 0.17 $\pm$ 0.08        & N/A                 \\

		\end{tabular}
	}
	
\end{table*}
}

\newcommand{\tblResults}{
\begin{table*}[tb]
    \caption{
		Results for the best performing methods -- The best metric is marked bold while the 2nd and 3rd best are italic. Only methods with at least one top3 ranking across the budgets are presented. The full results are in the supplementary. (a) show the relative improvement over the baseline. (b) are detailed results for the budget of 100\% across all datasets.
	}
	\label{tbl:results}
	\begin{subtable}{.4\linewidth}
		\caption{Improvements}
		\label{tbl:results-0}
	\label{tbl:improves}
	
	\resizebox{\textwidth}{!}{
		\begin{tabular}{l c c c c c c c c c c c c c c c c}
			\toprule
			
			\multirow{2}{*}{Budget}       & \multicolumn{2}{c}{10\%}                 &  \multicolumn{2}{c}{100\%}                 &  \multicolumn{2}{c}{1000\%}                \\
              \cmidrule(r){2-3} \cmidrule(r){4-5}  \cmidrule(r){6-7}  
                     
                     & Median & Mean $\pm$ SEM & Median & Mean $\pm$ SEM & Median & Mean $\pm$ SEM  \\
                     
                     \midrule

ELR+                 & -0.16 & -0.59 $\pm$ 0.22       & \textbf{-0.13} & -0.17 $\pm$ 0.06       & \textit{0.15} & \textit{0.24 $\pm$ 0.06}       \\
SGNP                 & -0.26 & -0.59 $\pm$ 0.27       & 0.00 & -0.15 $\pm$ 0.08       & 0.20 & \textit{0.26 $\pm$ 0.04}       \\
DivideMix            & -0.21 & \textbf{-0.78 $\pm$ 0.25}       & -0.03 & -0.17 $\pm$ 0.08       & \textit{0.16} & 0.31 $\pm$ 0.07       \\
Mean                 & -0.14 & -0.59 $\pm$ 0.22       & \textit{-0.12} & \textbf{-0.22 $\pm$ 0.08}       & N/A                 \\
$\pi$                & \textbf{-0.33} & -0.63 $\pm$ 0.19       & -0.06 & -0.17 $\pm$ 0.05       & N/A                 \\
$\pi$+DC3           & \textit{-0.32} & -0.62 $\pm$ 0.20       & \textit{-0.10} & \textbf{-0.22 $\pm$ 0.06}       & N/A                 \\
Pseudo v2 hard       & \textit{-0.30} & -0.39 $\pm$ 0.16       & -0.03 & \textit{-0.19 $\pm$ 0.08}       & 0.19 & 0.28 $\pm$ 0.04       \\
Pseudo v2 soft       & -0.28 & -0.40 $\pm$ 0.19       & -0.04 & -0.17 $\pm$ 0.06       & \textbf{0.01} & \textbf{0.00 $\pm$ 0.02}       \\
MOCOv2               & -0.29 & \textit{-0.63 $\pm$ 0.22}       & -0.08 & -0.13 $\pm$ 0.10       & N/A                 \\

		\end{tabular}
	}
	
	\end{subtable}
	\begin{subtable}{.6\linewidth}
	
			\caption{
		Details 100\% Budget
	}
	\label{tbl:results-1}
	
	\resizebox{\textwidth}{!}{
		\begin{tabular}{l c c c c c c c c c c c c c c c c}
			\toprule
			
			Dataset      & Benthic              & CIFAR10H             & MiceBone  & Pig             & Plankton             & QualityMRI           & Synthetic            & Treeversity\#1       & Treeversity\#6       & Turkey               \\
           \\
                     \midrule
                     
Baseline             &  1.17  $\pm$ 0.04        &  0.41  $\pm$ 0.02        &  0.55  $\pm$ 0.06  & 0.75 $\pm$    0.05  &  0.34  $\pm$ 0.02        &  1.73  $\pm$ 0.48        &  \textbf{0.08  $\pm$ 0.01}        &  0.49  $\pm$ 0.04        &  1.02  $\pm$ 0.03        &  \textit{0.40  $\pm$ 0.08}       \\
ELR+                 &  \textbf{0.70  $\pm$ 0.01}        &  \textit{0.29  $\pm$ 0.02}  &  \textbf{0.29  $\pm$ 0.01}   & \textit{0.62 $\pm$ 0.09 }           &  \textbf{0.24  $\pm$ 0.03}        &  1.44  $\pm$ 0.83        &  0.18  $\pm$ 0.02        &  \textbf{0.46  $\pm$ 0.01}        &  \textbf{0.47  $\pm$ 0.05}        &  0.52  $\pm$ 0.05       \\

SGNP                 &  1.11  $\pm$ 0.05        &  0.38  $\pm$ 0.01        &  0.56  $\pm$ 0.14    & 0.77 $\pm$  0.07  &  0.33  $\pm$ 0.02        &  \textbf{0.25  $\pm$ 0.14}        &  0.10  $\pm$ 0.00        &  \textbf{0.46  $\pm$ 0.01}        &  1.07  $\pm$ 0.05        &  \textit{0.42  $\pm$ 0.08}       \\
DivideMix            &  0.87  $\pm$ 0.07        &  0.36  $\pm$ 0.02        &  \textit{0.38  $\pm$ 0.05}   & 0.95 $\pm$ 0.04     &  0.34  $\pm$ 0.01        &  0.46  $\pm$ 0.26        &  0.33  $\pm$ 0.00        &  0.47  $\pm$ 0.01        &  0.62  $\pm$ 0.05        &  0.43  $\pm$ 0.08       \\

Mean                 &  0.80  $\pm$ 0.06        &  \textbf{0.28  $\pm$ 0.02}        &  \textit{0.38  $\pm$ 0.02 }   & \textit{0.64 $\pm$ 0.03}    &  0.32  $\pm$ 0.01        &  \textit{0.35  $\pm$ 0.11}        &  0.09  $\pm$ 0.01        &  0.61  $\pm$ 0.01        &  0.52  $\pm$ 0.03        &  0.75  $\pm$ 0.07       \\
$\pi$                &  \textit{0.71  $\pm$ 0.03}        &  0.33  $\pm$ 0.02        &  \textit{0.38  $\pm$ 0.00}      & 0.83 $\pm$ 0.18   &  \textit{0.30  $\pm$ 0.02}        &  0.98  $\pm$ 0.04        &  \textbf{0.08  $\pm$ 0.01}        &  0.52  $\pm$ 0.01        &  \textit{0.51  $\pm$ 0.03}        &  0.55  $\pm$ 0.01       \\
$\pi$+DC3           &  \textit{0.72  $\pm$ 0.06}        &  \textit{0.30  $\pm$ 0.02}        &  0.39  $\pm$ 0.04      & 0.67 $\pm$ 0.06  &  \textit{0.29  $\pm$ 0.01}        &  0.88  $\pm$ 0.34        &  \textbf{0.08  $\pm$ 0.00 }       &  0.48  $\pm$ 0.01        &  \textit{0.49  $\pm$ 0.00 }       &  0.44  $\pm$ 0.08       \\

Pseudo v2 hard       &  0.97 $\pm$ 0.10        &  0.43 $\pm$ 0.02        &  0.45 $\pm$ 0.10        &  0.67 $\pm$ 0.04        &  0.31 $\pm$ 0.03        &  0.29 $\pm$ 0.05        &  0.10 $\pm$ 0.01        &  0.46 $\pm$ 0.03        &  0.99 $\pm$ 0.09        &  0.39 $\pm$ 0.05       \\

Pseudo v2 soft       &  1.00  $\pm$ 0.08        &  0.41  $\pm$ 0.02        &  0.40  $\pm$ 0.08      & 0.70 $\pm$ 0.06   &  0.32  $\pm$ 0.06        &  0.62  $\pm$ 0.16        &  0.10  $\pm$ 0.01        &  \textbf{0.46  $\pm$ 0.01}        &  0.83  $\pm$ 0.13        &  \textbf{0.37  $\pm$ 0.08}       \\

MOCOv2               &  0.91  $\pm$ 0.05        &  0.98  $\pm$ 0.02        & \textit{ 0.37  $\pm$ 0.04}   & \textbf{0.56 $\pm$ 0.01}     &  0.52  $\pm$ 0.01        &  \textit{0.29  $\pm$ 0.11}        &  0.13  $\pm$ 0.01        &  0.80  $\pm$ 0.03        &  0.61  $\pm$ 0.01        &  \textit{0.42  $\pm$ 0.08}       \\

		\end{tabular}
	}

	\end{subtable}
\end{table*}
}

\newcommand{\tblDetailsTenAcc}{
\begin{table*}[tb]
		\caption{
		Details 10\% Budget Acc
	}
	\label{tbl:detailsTenAcc}
	\resizebox{\textwidth}{!}{
		\begin{tabular}{l c c c c c c c c c c c c c c c c}
			\toprule
			
			Dataset      & Benthic              & CIFAR10H             & MiceBone    & Pig         & Plankton             & QualityMRI           & Synthetic            & Treeversity\#1       & Treeversity\#6       & Turkey               \\
           \\
                     \midrule
                     
Baseline             &  0.64 $\pm$ 0.01        &  0.91 $\pm$ 0.00        &  0.62 $\pm$ 0.09        &  0.36 $\pm$ 0.04        &  0.90 $\pm$ 0.01        &  0.67 $\pm$ 0.04        &  0.88 $\pm$ 0.00        &  0.80 $\pm$ 0.02        &  0.57 $\pm$ 0.05        &  0.76 $\pm$ 0.03       \\
ELR+                 &  0.69 $\pm$ 0.01        &  0.89 $\pm$ 0.01        &  0.56 $\pm$ 0.05        &  0.39 $\pm$ 0.01        &  0.92 $\pm$ 0.01        &  0.50 $\pm$ 0.00        &  0.92 $\pm$ 0.01        &  0.80 $\pm$ 0.00        &  0.75 $\pm$ 0.02        &  0.36 $\pm$ 0.02       \\
Het                  &  0.63 $\pm$ 0.01        &  0.90 $\pm$ 0.01        &  0.64 $\pm$ 0.01        &  0.36 $\pm$ 0.01        &  0.90 $\pm$ 0.01        & N/A                  &  0.84 $\pm$ 0.02        &  0.80 $\pm$ 0.01        &  0.60 $\pm$ 0.01        &  0.72 $\pm$ 0.01       \\
SGNP                 &  0.64 $\pm$ 0.00        &  0.90 $\pm$ 0.01        &  0.61 $\pm$ 0.03        &  0.33 $\pm$ 0.07        &  0.91 $\pm$ 0.01        &  0.63 $\pm$ 0.11        &  0.84 $\pm$ 0.03        &  0.80 $\pm$ 0.01        &  0.56 $\pm$ 0.03        &  0.75 $\pm$ 0.05       \\
DivideMix            &  0.69 $\pm$ 0.04        &  0.90 $\pm$ 0.01        &  0.66 $\pm$ 0.02        &  0.43 $\pm$ 0.02        &  0.92 $\pm$ 0.01        &  0.52 $\pm$ 0.02        &  0.93 $\pm$ 0.00        &  0.82 $\pm$ 0.01        &  0.77 $\pm$ 0.02        &  0.68 $\pm$ 0.02       \\
Fixmatch             &  0.69 $\pm$ 0.04        &  0.91 $\pm$ 0.01        &  0.72 $\pm$ 0.02        &  0.29 $\pm$ 0.05        &  0.92 $\pm$ 0.01        &  0.72 $\pm$ 0.06        &  0.83 $\pm$ 0.03        &  0.78 $\pm$ 0.03        &  0.64 $\pm$ 0.02        &  0.69 $\pm$ 0.06       \\
Fixmatch + DC3      &  0.69 $\pm$ 0.01        &  0.90 $\pm$ 0.00        &  0.59 $\pm$ 0.08        &  0.33 $\pm$ 0.03        &  0.92 $\pm$ 0.01        &  0.63 $\pm$ 0.07        &  0.81 $\pm$ 0.13        &  0.79 $\pm$ 0.01        &  0.67 $\pm$ 0.01        &  0.76 $\pm$ 0.03       \\
Mean                 &  0.69 $\pm$ 0.01        &  0.90 $\pm$ 0.01        &  0.64 $\pm$ 0.03        &  0.39 $\pm$ 0.02        &  0.91 $\pm$ 0.01        &  0.64 $\pm$ 0.06        &  0.90 $\pm$ 0.01        &  0.71 $\pm$ 0.01        &  0.66 $\pm$ 0.01        &  0.52 $\pm$ 0.01       \\
Mean+DC3            &  0.71 $\pm$ 0.01        &  0.90 $\pm$ 0.02        &  0.66 $\pm$ 0.00        &  0.35 $\pm$ 0.03        &  0.91 $\pm$ 0.01        &  0.62 $\pm$ 0.07        &  0.89 $\pm$ 0.01        &  0.75 $\pm$ 0.00        &  0.66 $\pm$ 0.01        &  0.66 $\pm$ 0.06       \\
$\pi$                &  0.71 $\pm$ 0.02        &  0.90 $\pm$ 0.00        &  0.65 $\pm$ 0.01        &  0.34 $\pm$ 0.01        &  0.91 $\pm$ 0.01        &  0.71 $\pm$ 0.01        &  0.92 $\pm$ 0.01        &  0.75 $\pm$ 0.01        &  0.66 $\pm$ 0.00        &  0.63 $\pm$ 0.02       \\
$\pi$+DC3           &  0.70 $\pm$ 0.02        &  0.90 $\pm$ 0.01        &  0.64 $\pm$ 0.05        &  0.37 $\pm$ 0.04        &  0.92 $\pm$ 0.00        &  0.68 $\pm$ 0.06        &  0.91 $\pm$ 0.01        &  0.77 $\pm$ 0.00        &  0.66 $\pm$ 0.00        &  0.69 $\pm$ 0.03       \\
Pseudo v1            &  0.65 $\pm$ 0.05        &  0.91 $\pm$ 0.00        &  0.71 $\pm$ 0.02        &  0.34 $\pm$ 0.08        &  0.91 $\pm$ 0.01        &  0.68 $\pm$ 0.05        &  0.92 $\pm$ 0.01        &  0.80 $\pm$ 0.01        &  0.69 $\pm$ 0.01        &  0.80 $\pm$ 0.04       \\
Pseudo v1 + DC3     &  0.64 $\pm$ 0.04        &  0.90 $\pm$ 0.02        &  0.73 $\pm$ 0.01        &  0.30 $\pm$ 0.04        &  0.91 $\pm$ 0.01        &  0.75 $\pm$ 0.03        &  0.90 $\pm$ 0.02        &  0.82 $\pm$ 0.01        &  0.69 $\pm$ 0.00        &  0.79 $\pm$ 0.04       \\
Pseudo v2 hard       &  0.66 $\pm$ 0.02        &  0.84 $\pm$ 0.01        &  0.64 $\pm$ 0.05        &  0.35 $\pm$ 0.01        &  0.89 $\pm$ 0.02        &  0.73 $\pm$ 0.09        &  0.87 $\pm$ 0.01        &  0.80 $\pm$ 0.01        &  0.60 $\pm$ 0.03        &  0.78 $\pm$ 0.04       \\
Pseudo v2 soft       &  0.65 $\pm$ 0.02        &  0.85 $\pm$ 0.01        &  0.68 $\pm$ 0.01        &  0.31 $\pm$ 0.04        &  0.88 $\pm$ 0.04        &  0.74 $\pm$ 0.06        &  0.87 $\pm$ 0.01        &  0.79 $\pm$ 0.01        &  0.58 $\pm$ 0.04        &  0.71 $\pm$ 0.05       \\
Pseudo v2 not        &  0.58 $\pm$ 0.01        &  0.59 $\pm$ 0.04        &  0.64 $\pm$ 0.04        &  0.33 $\pm$ 0.05        &  0.39 $\pm$ 0.14        &  0.60 $\pm$ 0.06        &  0.62 $\pm$ 0.10        &  0.69 $\pm$ 0.07        &  0.53 $\pm$ 0.10        &  0.68 $\pm$ 0.05       \\
BYOL                 &  0.64 $\pm$ 0.01        &  0.38 $\pm$ 0.06        &  0.54 $\pm$ 0.03        &  0.34 $\pm$ 0.02        &  0.80 $\pm$ 0.02        &  0.50 $\pm$ 0.02        &  0.68 $\pm$ 0.00        &  0.59 $\pm$ 0.03        &  0.59 $\pm$ 0.02        &  0.62 $\pm$ 0.04       \\
MOCOv2               &  0.58 $\pm$ 0.01        &  0.64 $\pm$ 0.01        &  0.52 $\pm$ 0.03        &  0.35 $\pm$ 0.01        &  0.79 $\pm$ 0.01        &  0.56 $\pm$ 0.02        &  0.89 $\pm$ 0.02        &  0.61 $\pm$ 0.01        &  0.58 $\pm$ 0.01        &  0.40 $\pm$ 0.05       \\
SimCLR               &  0.58 $\pm$ 0.03        &  0.87 $\pm$ 0.01        &  0.65 $\pm$ 0.05        &  0.35 $\pm$ 0.07        &  0.88 $\pm$ 0.01        &  0.63 $\pm$ 0.11        &  0.79 $\pm$ 0.00        &  0.69 $\pm$ 0.01        &  0.54 $\pm$ 0.00        &  0.66 $\pm$ 0.05       \\
SWAV                 &  0.24 $\pm$ 0.02        &  0.74 $\pm$ 0.01        &  0.53 $\pm$ 0.04        &  0.35 $\pm$ 0.04        &  0.86 $\pm$ 0.01        &  0.60 $\pm$ 0.08        &  0.66 $\pm$ 0.03        &  0.65 $\pm$ 0.01        &  0.60 $\pm$ 0.03        &  0.56 $\pm$ 0.10       \\

		\end{tabular}
	}
	
\end{table*}
}

\newcommand{\tblDetailsTenKl}{
\begin{table*}[tb]
		\caption{
		Details 10\% Budget KL
	}
	\label{tbl:detailsTenKl}
	\resizebox{\textwidth}{!}{
		\begin{tabular}{l c c c c c c c c c c c c c c c c}
			\toprule
			
			Dataset      & Benthic              & CIFAR10H             & MiceBone   & Pig          & Plankton             & QualityMRI           & Synthetic            & Treeversity\#1       & Treeversity\#6       & Turkey               \\
           \\
                     \midrule
                     
Baseline             &  1.17 $\pm$ 0.04        &  0.41 $\pm$ 0.02        &  0.55 $\pm$ 0.06        &  0.75 $\pm$ 0.05        &  0.34 $\pm$ 0.02        &  1.73 $\pm$ 0.48        &  0.08 $\pm$ 0.01        &  0.49 $\pm$ 0.04        &  1.02 $\pm$ 0.03        &  0.40 $\pm$ 0.08       \\
ELR+                 &  0.70 $\pm$ 0.01        &  0.29 $\pm$ 0.02        &  0.29 $\pm$ 0.01        &  0.62 $\pm$ 0.09        &  0.24 $\pm$ 0.03        &  1.44 $\pm$ 0.83        &  0.18 $\pm$ 0.02        &  0.46 $\pm$ 0.01        &  0.47 $\pm$ 0.05        &  0.52 $\pm$ 0.05       \\
Het                  &  1.16 $\pm$ 0.15        &  0.39 $\pm$ 0.01        &  0.62 $\pm$ 0.18        &  0.70 $\pm$ 0.10        &  0.34 $\pm$ 0.03        & N/A                  &  0.10 $\pm$ 0.01        &  0.46 $\pm$ 0.01        &  0.99 $\pm$ 0.02        &  0.52 $\pm$ 0.07       \\
SGNP                 &  1.11 $\pm$ 0.05        &  0.38 $\pm$ 0.01        &  0.56 $\pm$ 0.14        &  0.77 $\pm$ 0.07        &  0.33 $\pm$ 0.02        &  0.25 $\pm$ 0.14        &  0.10 $\pm$ 0.00        &  0.46 $\pm$ 0.01        &  1.07 $\pm$ 0.05        &  0.42 $\pm$ 0.08       \\
DivideMix            &  0.87 $\pm$ 0.07        &  0.36 $\pm$ 0.02        &  0.38 $\pm$ 0.05        &  0.95 $\pm$ 0.04        &  0.34 $\pm$ 0.01        &  0.46 $\pm$ 0.26        &  0.33 $\pm$ 0.00        &  0.47 $\pm$ 0.01        &  0.62 $\pm$ 0.05        &  0.43 $\pm$ 0.08       \\
Fixmatch             &  0.88 $\pm$ 0.24        &  0.37 $\pm$ 0.02        &  0.41 $\pm$ 0.01        &  0.72 $\pm$ 0.05        &  0.27 $\pm$ 0.02        &  0.99 $\pm$ 0.30        &  0.11 $\pm$ 0.02        &  0.48 $\pm$ 0.06        &  0.51 $\pm$ 0.01        &  0.48 $\pm$ 0.05       \\
Fixmatch + DC3      &  0.82 $\pm$ 0.02        &  0.34 $\pm$ 0.03        &  0.54 $\pm$ 0.12        &  0.71 $\pm$ 0.04        &  0.27 $\pm$ 0.02        &  1.41 $\pm$ 0.15        &  0.18 $\pm$ 0.15        &  0.44 $\pm$ 0.01        &  0.46 $\pm$ 0.01        &  0.47 $\pm$ 0.03       \\
Mean                 &  0.80 $\pm$ 0.06        &  0.28 $\pm$ 0.02        &  0.38 $\pm$ 0.02        &  0.64 $\pm$ 0.03        &  0.32 $\pm$ 0.01        &  0.35 $\pm$ 0.11        &  0.09 $\pm$ 0.01        &  0.61 $\pm$ 0.01        &  0.52 $\pm$ 0.03        &  0.75 $\pm$ 0.07       \\
Mean+DC3            &  0.73 $\pm$ 0.02        &  0.28 $\pm$ 0.04        &  0.37 $\pm$ 0.00        &  0.71 $\pm$ 0.11        &  0.30 $\pm$ 0.02        &  1.42 $\pm$ 0.93        &  0.11 $\pm$ 0.02        &  0.51 $\pm$ 0.01        &  0.48 $\pm$ 0.02        &  0.55 $\pm$ 0.10       \\
$\pi$                &  0.71 $\pm$ 0.03        &  0.33 $\pm$ 0.02        &  0.38 $\pm$ 0.00        &  0.83 $\pm$ 0.18        &  0.30 $\pm$ 0.02        &  0.98 $\pm$ 0.04        &  0.08 $\pm$ 0.01        &  0.52 $\pm$ 0.01        &  0.51 $\pm$ 0.03        &  0.55 $\pm$ 0.01       \\
$\pi$+DC3           &  0.72 $\pm$ 0.06        &  0.30 $\pm$ 0.02        &  0.39 $\pm$ 0.04        &  0.67 $\pm$ 0.06        &  0.29 $\pm$ 0.01        &  0.88 $\pm$ 0.34        &  0.08 $\pm$ 0.00        &  0.48 $\pm$ 0.01        &  0.49 $\pm$ 0.00        &  0.44 $\pm$ 0.08       \\
Pseudo v1            &  1.10 $\pm$ 0.13        &  0.35 $\pm$ 0.02        &  0.33 $\pm$ 0.01        &  0.71 $\pm$ 0.07        &  0.32 $\pm$ 0.02        &  1.25 $\pm$ 0.35        &  0.05 $\pm$ 0.00        &  0.42 $\pm$ 0.01        &  0.42 $\pm$ 0.01        &  0.33 $\pm$ 0.03       \\
Pseudo v1 + DC3     &  1.10 $\pm$ 0.12        &  0.33 $\pm$ 0.03        &  0.34 $\pm$ 0.02        &  2.87 $\pm$ 3.09        &  0.31 $\pm$ 0.01        &  1.22 $\pm$ 0.51        &  0.06 $\pm$ 0.00        &  0.41 $\pm$ 0.01        &  0.43 $\pm$ 0.00        &  0.40 $\pm$ 0.04       \\
Pseudo v2 hard       &  0.97 $\pm$ 0.10        &  0.43 $\pm$ 0.02        &  0.45 $\pm$ 0.10        &  0.67 $\pm$ 0.04        &  0.31 $\pm$ 0.03        &  0.29 $\pm$ 0.05        &  0.10 $\pm$ 0.01        &  0.46 $\pm$ 0.03        &  0.99 $\pm$ 0.09        &  0.39 $\pm$ 0.05       \\
Pseudo v2 soft       &  1.00 $\pm$ 0.08        &  0.41 $\pm$ 0.02        &  0.40 $\pm$ 0.08        &  0.70 $\pm$ 0.06        &  0.32 $\pm$ 0.06        &  0.62 $\pm$ 0.16        &  0.10 $\pm$ 0.01        &  0.46 $\pm$ 0.01        &  0.83 $\pm$ 0.13        &  0.37 $\pm$ 0.08       \\
Pseudo v2 not        &  1.05 $\pm$ 0.02        &  1.12 $\pm$ 0.04        &  0.46 $\pm$ 0.10        &  0.79 $\pm$ 0.09        &  2.70 $\pm$ 0.89        &  0.12 $\pm$ 0.02        &  1.00 $\pm$ 0.57        &  0.68 $\pm$ 0.17        &  0.67 $\pm$ 0.11        &  0.57 $\pm$ 0.06       \\
BYOL                 &  0.74 $\pm$ 0.03        &  1.61 $\pm$ 0.22        &  0.46 $\pm$ 0.02        &  0.64 $\pm$ 0.01        &  0.51 $\pm$ 0.05        &  1.07 $\pm$ 0.18        &  0.36 $\pm$ 0.01        &  0.82 $\pm$ 0.09        &  0.58 $\pm$ 0.03        &  0.34 $\pm$ 0.04       \\
MOCOv2               &  0.91 $\pm$ 0.05        &  0.98 $\pm$ 0.02        &  0.37 $\pm$ 0.04        &  0.56 $\pm$ 0.01        &  0.52 $\pm$ 0.01        &  0.29 $\pm$ 0.11        &  0.13 $\pm$ 0.01        &  0.80 $\pm$ 0.03        &  0.61 $\pm$ 0.01        &  0.42 $\pm$ 0.08       \\
SimCLR               &  1.19 $\pm$ 0.10        &  0.36 $\pm$ 0.02        &  0.48 $\pm$ 0.05        &  0.75 $\pm$ 0.09        &  0.29 $\pm$ 0.02        &  1.83 $\pm$ 1.01        &  0.38 $\pm$ 0.03        &  0.68 $\pm$ 0.00        &  0.94 $\pm$ 0.04        &  0.43 $\pm$ 0.06       \\
SWAV                 &  1.56 $\pm$ 0.16        &  0.70 $\pm$ 0.01        &  0.56 $\pm$ 0.09        &  1.33 $\pm$ 0.98        &  0.38 $\pm$ 0.01        &  1.90 $\pm$ 0.31        &  0.36 $\pm$ 0.03        &  0.71 $\pm$ 0.02        &  0.61 $\pm$ 0.03        &  0.51 $\pm$ 0.04       \\

		\end{tabular}
	}
	
\end{table*}
}

\newcommand{\tblDetailsHundretAcc}{
\begin{table*}[tb]
		\caption{
		Details 100\% Budget Acc
	}
	\label{tbl:detailsHundretAcc}
	\resizebox{\textwidth}{!}{
		\begin{tabular}{l c c c c c c c c c c c c c c c c}
			\toprule
			
			Dataset      & Benthic              & CIFAR10H             & MiceBone & Pig            & Plankton             & QualityMRI           & Synthetic            & Treeversity\#1       & Treeversity\#6       & Turkey               \\
           \\
                     \midrule
                     
Baseline             &  0.64 $\pm$ 0.01        &  0.91 $\pm$ 0.00        &  0.62 $\pm$ 0.09        &  0.36 $\pm$ 0.04        &  0.90 $\pm$ 0.01        &  0.67 $\pm$ 0.04        &  0.88 $\pm$ 0.00        &  0.80 $\pm$ 0.02        &  0.57 $\pm$ 0.05        &  0.76 $\pm$ 0.03       \\
ELR+                 &  0.69 $\pm$ 0.01        &  0.89 $\pm$ 0.01        &  0.56 $\pm$ 0.05        &  0.39 $\pm$ 0.01        &  0.92 $\pm$ 0.01        &  0.50 $\pm$ 0.00        &  0.92 $\pm$ 0.01        &  0.80 $\pm$ 0.00        &  0.75 $\pm$ 0.02        &  0.36 $\pm$ 0.02       \\
Het                  &  0.63 $\pm$ 0.01        &  0.90 $\pm$ 0.01        &  0.64 $\pm$ 0.01        &  0.36 $\pm$ 0.01        &  0.90 $\pm$ 0.01        & N/A                  &  0.84 $\pm$ 0.02        &  0.80 $\pm$ 0.01        &  0.60 $\pm$ 0.01        &  0.72 $\pm$ 0.01       \\
SGNP                 &  0.64 $\pm$ 0.00        &  0.90 $\pm$ 0.01        &  0.61 $\pm$ 0.03        &  0.33 $\pm$ 0.07        &  0.91 $\pm$ 0.01        &  0.63 $\pm$ 0.11        &  0.84 $\pm$ 0.03        &  0.80 $\pm$ 0.01        &  0.56 $\pm$ 0.03        &  0.75 $\pm$ 0.05       \\
DivideMix            &  0.69 $\pm$ 0.04        &  0.90 $\pm$ 0.01        &  0.66 $\pm$ 0.02        &  0.43 $\pm$ 0.02        &  0.92 $\pm$ 0.01        &  0.52 $\pm$ 0.02        &  0.93 $\pm$ 0.00        &  0.82 $\pm$ 0.01        &  0.77 $\pm$ 0.02        &  0.68 $\pm$ 0.02       \\
Fixmatch             &  0.69 $\pm$ 0.04        &  0.91 $\pm$ 0.01        &  0.72 $\pm$ 0.02        &  0.29 $\pm$ 0.05        &  0.92 $\pm$ 0.01        &  0.72 $\pm$ 0.06        &  0.83 $\pm$ 0.03        &  0.78 $\pm$ 0.03        &  0.64 $\pm$ 0.02        &  0.69 $\pm$ 0.06       \\
Fixmatch + DC3      &  0.69 $\pm$ 0.01        &  0.90 $\pm$ 0.00        &  0.59 $\pm$ 0.08        &  0.33 $\pm$ 0.03        &  0.92 $\pm$ 0.01        &  0.63 $\pm$ 0.07        &  0.81 $\pm$ 0.13        &  0.79 $\pm$ 0.01        &  0.67 $\pm$ 0.01        &  0.76 $\pm$ 0.03       \\
Mean                 &  0.69 $\pm$ 0.01        &  0.90 $\pm$ 0.01        &  0.64 $\pm$ 0.03        &  0.39 $\pm$ 0.02        &  0.91 $\pm$ 0.01        &  0.64 $\pm$ 0.06        &  0.90 $\pm$ 0.01        &  0.71 $\pm$ 0.01        &  0.66 $\pm$ 0.01        &  0.52 $\pm$ 0.01       \\
Mean+DC3            &  0.71 $\pm$ 0.01        &  0.90 $\pm$ 0.02        &  0.66 $\pm$ 0.00        &  0.35 $\pm$ 0.03        &  0.91 $\pm$ 0.01        &  0.62 $\pm$ 0.07        &  0.89 $\pm$ 0.01        &  0.75 $\pm$ 0.00        &  0.66 $\pm$ 0.01        &  0.66 $\pm$ 0.06       \\
$\pi$                &  0.71 $\pm$ 0.02        &  0.90 $\pm$ 0.00        &  0.65 $\pm$ 0.01        &  0.34 $\pm$ 0.01        &  0.91 $\pm$ 0.01        &  0.71 $\pm$ 0.01        &  0.92 $\pm$ 0.01        &  0.75 $\pm$ 0.01        &  0.66 $\pm$ 0.00        &  0.63 $\pm$ 0.02       \\
$\pi$+DC3           &  0.70 $\pm$ 0.02        &  0.90 $\pm$ 0.01        &  0.64 $\pm$ 0.05        &  0.37 $\pm$ 0.04        &  0.92 $\pm$ 0.00        &  0.68 $\pm$ 0.06        &  0.91 $\pm$ 0.01        &  0.77 $\pm$ 0.00        &  0.66 $\pm$ 0.00        &  0.69 $\pm$ 0.03       \\
Pseudo v1            &  0.65 $\pm$ 0.05        &  0.91 $\pm$ 0.00        &  0.71 $\pm$ 0.02        &  0.34 $\pm$ 0.08        &  0.91 $\pm$ 0.01        &  0.68 $\pm$ 0.05        &  0.92 $\pm$ 0.01        &  0.80 $\pm$ 0.01        &  0.69 $\pm$ 0.01        &  0.80 $\pm$ 0.04       \\
Pseudo v1 + DC3     &  0.64 $\pm$ 0.04        &  0.90 $\pm$ 0.02        &  0.73 $\pm$ 0.01        &  0.30 $\pm$ 0.04        &  0.91 $\pm$ 0.01        &  0.75 $\pm$ 0.03        &  0.90 $\pm$ 0.02        &  0.82 $\pm$ 0.01        &  0.69 $\pm$ 0.00        &  0.79 $\pm$ 0.04       \\
Pseudo v2 hard       &  0.66 $\pm$ 0.02        &  0.84 $\pm$ 0.01        &  0.64 $\pm$ 0.05        &  0.35 $\pm$ 0.01        &  0.89 $\pm$ 0.02        &  0.73 $\pm$ 0.09        &  0.87 $\pm$ 0.01        &  0.80 $\pm$ 0.01        &  0.60 $\pm$ 0.03        &  0.78 $\pm$ 0.04       \\
Pseudo v2 soft       &  0.65 $\pm$ 0.02        &  0.85 $\pm$ 0.01        &  0.68 $\pm$ 0.01        &  0.31 $\pm$ 0.04        &  0.88 $\pm$ 0.04        &  0.74 $\pm$ 0.06        &  0.87 $\pm$ 0.01        &  0.79 $\pm$ 0.01        &  0.58 $\pm$ 0.04        &  0.71 $\pm$ 0.05       \\
Pseudo v2 not        &  0.58 $\pm$ 0.01        &  0.59 $\pm$ 0.04        &  0.64 $\pm$ 0.04        &  0.33 $\pm$ 0.05        &  0.39 $\pm$ 0.14        &  0.60 $\pm$ 0.06        &  0.62 $\pm$ 0.10        &  0.69 $\pm$ 0.07        &  0.53 $\pm$ 0.10        &  0.68 $\pm$ 0.05       \\
BYOL                 &  0.64 $\pm$ 0.01        &  0.38 $\pm$ 0.06        &  0.54 $\pm$ 0.03        &  0.34 $\pm$ 0.02        &  0.80 $\pm$ 0.02        &  0.50 $\pm$ 0.02        &  0.68 $\pm$ 0.00        &  0.59 $\pm$ 0.03        &  0.59 $\pm$ 0.02        &  0.62 $\pm$ 0.04       \\
MOCOv2               &  0.58 $\pm$ 0.01        &  0.64 $\pm$ 0.01        &  0.52 $\pm$ 0.03        &  0.35 $\pm$ 0.01        &  0.79 $\pm$ 0.01        &  0.56 $\pm$ 0.02        &  0.89 $\pm$ 0.02        &  0.61 $\pm$ 0.01        &  0.58 $\pm$ 0.01        &  0.40 $\pm$ 0.05       \\
SimCLR               &  0.58 $\pm$ 0.03        &  0.87 $\pm$ 0.01        &  0.65 $\pm$ 0.05        &  0.35 $\pm$ 0.07        &  0.88 $\pm$ 0.01        &  0.63 $\pm$ 0.11        &  0.79 $\pm$ 0.00        &  0.69 $\pm$ 0.01        &  0.54 $\pm$ 0.00        &  0.66 $\pm$ 0.05       \\
SWAV                 &  0.24 $\pm$ 0.02        &  0.74 $\pm$ 0.01        &  0.53 $\pm$ 0.04        &  0.35 $\pm$ 0.04        &  0.86 $\pm$ 0.01        &  0.60 $\pm$ 0.08        &  0.66 $\pm$ 0.03        &  0.65 $\pm$ 0.01        &  0.60 $\pm$ 0.03        &  0.56 $\pm$ 0.10       \\

		\end{tabular}
	}
	
\end{table*}
}

\newcommand{\tblDetailsHundretKl}{
\begin{table*}[tb]
		\caption{
		Details 100\% Budget Kl
	}
	\label{tbl:detailsHundretKl}
	\resizebox{\textwidth}{!}{
		\begin{tabular}{l c c c c c c c c c c c c c c c c}
			\toprule
			
			Dataset      & Benthic              & CIFAR10H             & MiceBone  & Pig           & Plankton             & QualityMRI           & Synthetic            & Treeversity\#1       & Treeversity\#6       & Turkey               \\
           \\
                     \midrule
                     
Baseline             &  1.17 $\pm$ 0.04        &  0.41 $\pm$ 0.02        &  0.55 $\pm$ 0.06        &  0.75 $\pm$ 0.05        &  0.34 $\pm$ 0.02        &  1.73 $\pm$ 0.48        &  0.08 $\pm$ 0.01        &  0.49 $\pm$ 0.04        &  1.02 $\pm$ 0.03        &  0.40 $\pm$ 0.08       \\
ELR+                 &  0.70 $\pm$ 0.01        &  0.29 $\pm$ 0.02        &  0.29 $\pm$ 0.01        &  0.62 $\pm$ 0.09        &  0.24 $\pm$ 0.03        &  1.44 $\pm$ 0.83        &  0.18 $\pm$ 0.02        &  0.46 $\pm$ 0.01        &  0.47 $\pm$ 0.05        &  0.52 $\pm$ 0.05       \\
Het                  &  1.16 $\pm$ 0.15        &  0.39 $\pm$ 0.01        &  0.62 $\pm$ 0.18        &  0.70 $\pm$ 0.10        &  0.34 $\pm$ 0.03        & N/A                  &  0.10 $\pm$ 0.01        &  0.46 $\pm$ 0.01        &  0.99 $\pm$ 0.02        &  0.52 $\pm$ 0.07       \\
SGNP                 &  1.11 $\pm$ 0.05        &  0.38 $\pm$ 0.01        &  0.56 $\pm$ 0.14        &  0.77 $\pm$ 0.07        &  0.33 $\pm$ 0.02        &  0.25 $\pm$ 0.14        &  0.10 $\pm$ 0.00        &  0.46 $\pm$ 0.01        &  1.07 $\pm$ 0.05        &  0.42 $\pm$ 0.08       \\
DivideMix            &  0.87 $\pm$ 0.07        &  0.36 $\pm$ 0.02        &  0.38 $\pm$ 0.05        &  0.95 $\pm$ 0.04        &  0.34 $\pm$ 0.01        &  0.46 $\pm$ 0.26        &  0.33 $\pm$ 0.00        &  0.47 $\pm$ 0.01        &  0.62 $\pm$ 0.05        &  0.43 $\pm$ 0.08       \\
Fixmatch             &  0.88 $\pm$ 0.24        &  0.37 $\pm$ 0.02        &  0.41 $\pm$ 0.01        &  0.72 $\pm$ 0.05        &  0.27 $\pm$ 0.02        &  0.99 $\pm$ 0.30        &  0.11 $\pm$ 0.02        &  0.48 $\pm$ 0.06        &  0.51 $\pm$ 0.01        &  0.48 $\pm$ 0.05       \\
Fixmatch + DC3      &  0.82 $\pm$ 0.02        &  0.34 $\pm$ 0.03        &  0.54 $\pm$ 0.12        &  0.71 $\pm$ 0.04        &  0.27 $\pm$ 0.02        &  1.41 $\pm$ 0.15        &  0.18 $\pm$ 0.15        &  0.44 $\pm$ 0.01        &  0.46 $\pm$ 0.01        &  0.47 $\pm$ 0.03       \\
Mean                 &  0.80 $\pm$ 0.06        &  0.28 $\pm$ 0.02        &  0.38 $\pm$ 0.02        &  0.64 $\pm$ 0.03        &  0.32 $\pm$ 0.01        &  0.35 $\pm$ 0.11        &  0.09 $\pm$ 0.01        &  0.61 $\pm$ 0.01        &  0.52 $\pm$ 0.03        &  0.75 $\pm$ 0.07       \\
Mean+DC3            &  0.73 $\pm$ 0.02        &  0.28 $\pm$ 0.04        &  0.37 $\pm$ 0.00        &  0.71 $\pm$ 0.11        &  0.30 $\pm$ 0.02        &  1.42 $\pm$ 0.93        &  0.11 $\pm$ 0.02        &  0.51 $\pm$ 0.01        &  0.48 $\pm$ 0.02        &  0.55 $\pm$ 0.10       \\
$\pi$                &  0.71 $\pm$ 0.03        &  0.33 $\pm$ 0.02        &  0.38 $\pm$ 0.00        &  0.83 $\pm$ 0.18        &  0.30 $\pm$ 0.02        &  0.98 $\pm$ 0.04        &  0.08 $\pm$ 0.01        &  0.52 $\pm$ 0.01        &  0.51 $\pm$ 0.03        &  0.55 $\pm$ 0.01       \\
$\pi$+DC3           &  0.72 $\pm$ 0.06        &  0.30 $\pm$ 0.02        &  0.39 $\pm$ 0.04        &  0.67 $\pm$ 0.06        &  0.29 $\pm$ 0.01        &  0.88 $\pm$ 0.34        &  0.08 $\pm$ 0.00        &  0.48 $\pm$ 0.01        &  0.49 $\pm$ 0.00        &  0.44 $\pm$ 0.08       \\
Pseudo v1            &  1.10 $\pm$ 0.13        &  0.35 $\pm$ 0.02        &  0.33 $\pm$ 0.01        &  0.71 $\pm$ 0.07        &  0.32 $\pm$ 0.02        &  1.25 $\pm$ 0.35        &  0.05 $\pm$ 0.00        &  0.42 $\pm$ 0.01        &  0.42 $\pm$ 0.01        &  0.33 $\pm$ 0.03       \\
Pseudo v1 + DC3     &  1.10 $\pm$ 0.12        &  0.33 $\pm$ 0.03        &  0.34 $\pm$ 0.02        &  2.87 $\pm$ 3.09        &  0.31 $\pm$ 0.01        &  1.22 $\pm$ 0.51        &  0.06 $\pm$ 0.00        &  0.41 $\pm$ 0.01        &  0.43 $\pm$ 0.00        &  0.40 $\pm$ 0.04       \\
Pseudo v2 hard       &  0.97 $\pm$ 0.10        &  0.43 $\pm$ 0.02        &  0.45 $\pm$ 0.10        &  0.67 $\pm$ 0.04        &  0.31 $\pm$ 0.03        &  0.29 $\pm$ 0.05        &  0.10 $\pm$ 0.01        &  0.46 $\pm$ 0.03        &  0.99 $\pm$ 0.09        &  0.39 $\pm$ 0.05       \\
Pseudo v2 soft       &  1.00 $\pm$ 0.08        &  0.41 $\pm$ 0.02        &  0.40 $\pm$ 0.08        &  0.70 $\pm$ 0.06        &  0.32 $\pm$ 0.06        &  0.62 $\pm$ 0.16        &  0.10 $\pm$ 0.01        &  0.46 $\pm$ 0.01        &  0.83 $\pm$ 0.13        &  0.37 $\pm$ 0.08       \\
Pseudo v2 not        &  1.05 $\pm$ 0.02        &  1.12 $\pm$ 0.04        &  0.46 $\pm$ 0.10        &  0.79 $\pm$ 0.09        &  2.70 $\pm$ 0.89        &  0.12 $\pm$ 0.02        &  1.00 $\pm$ 0.57        &  0.68 $\pm$ 0.17        &  0.67 $\pm$ 0.11        &  0.57 $\pm$ 0.06       \\
BYOL                 &  0.74 $\pm$ 0.03        &  1.61 $\pm$ 0.22        &  0.46 $\pm$ 0.02        &  0.64 $\pm$ 0.01        &  0.51 $\pm$ 0.05        &  1.07 $\pm$ 0.18        &  0.36 $\pm$ 0.01        &  0.82 $\pm$ 0.09        &  0.58 $\pm$ 0.03        &  0.34 $\pm$ 0.04       \\
MOCOv2               &  0.91 $\pm$ 0.05        &  0.98 $\pm$ 0.02        &  0.37 $\pm$ 0.04        &  0.56 $\pm$ 0.01        &  0.52 $\pm$ 0.01        &  0.29 $\pm$ 0.11        &  0.13 $\pm$ 0.01        &  0.80 $\pm$ 0.03        &  0.61 $\pm$ 0.01        &  0.42 $\pm$ 0.08       \\
SimCLR               &  1.19 $\pm$ 0.10        &  0.36 $\pm$ 0.02        &  0.48 $\pm$ 0.05        &  0.75 $\pm$ 0.09        &  0.29 $\pm$ 0.02        &  1.83 $\pm$ 1.01        &  0.38 $\pm$ 0.03        &  0.68 $\pm$ 0.00        &  0.94 $\pm$ 0.04        &  0.43 $\pm$ 0.06       \\
SWAV                 &  1.56 $\pm$ 0.16        &  0.70 $\pm$ 0.01        &  0.56 $\pm$ 0.09        &  1.33 $\pm$ 0.98        &  0.38 $\pm$ 0.01        &  1.90 $\pm$ 0.31        &  0.36 $\pm$ 0.03        &  0.71 $\pm$ 0.02        &  0.61 $\pm$ 0.03        &  0.51 $\pm$ 0.04       \\

		\end{tabular}
	}
	
\end{table*}
}

\newcommand{\tblDetailsThousandAcc}{
\begin{table*}[tb]
		\caption{
		Details 1000\% Budget Acc
	}
	\label{tbl:detailsThousandAcc}
	\resizebox{\textwidth}{!}{
		\begin{tabular}{l c c c c c c c c c c c c c c c c}
			\toprule
			
			Dataset      & Benthic              & CIFAR10H             & MiceBone       & Pig      & Plankton             & QualityMRI           & Synthetic            & Treeversity\#1       & Treeversity\#6       & Turkey               \\
           \\
                     \midrule
                     
Baseline             &  0.68 $\pm$ 0.01        &  0.91 $\pm$ 0.01        &  0.69 $\pm$ 0.07        &  0.42 $\pm$ 0.04        &  0.92 $\pm$ 0.01        &  0.56 $\pm$ 0.01        &  0.91 $\pm$ 0.01        &  0.82 $\pm$ 0.01        &  0.76 $\pm$ 0.01        &  0.75 $\pm$ 0.07       \\
ELR+                 &  0.68 $\pm$ 0.02        &  0.90 $\pm$ 0.01        &  0.64 $\pm$ 0.05        &  0.38 $\pm$ 0.00        &  0.92 $\pm$ 0.01        &  0.50 $\pm$ 0.00        &  0.93 $\pm$ 0.02        &  0.81 $\pm$ 0.01        &  0.79 $\pm$ 0.01        &  0.38 $\pm$ 0.04       \\
Het                  &  0.66 $\pm$ 0.02        &  0.91 $\pm$ 0.01        &  0.68 $\pm$ 0.02        &  0.39 $\pm$ 0.04        &  0.91 $\pm$ 0.01        & N/A                  &  0.91 $\pm$ 0.01        &  0.82 $\pm$ 0.01        &  0.68 $\pm$ 0.02        &  0.79 $\pm$ 0.05       \\
SGNP                 &  0.66 $\pm$ 0.01        &  0.91 $\pm$ 0.01        &  0.68 $\pm$ 0.02        &  0.36 $\pm$ 0.04        &  0.90 $\pm$ 0.02        &  0.68 $\pm$ 0.08        &  0.92 $\pm$ 0.01        &  0.81 $\pm$ 0.01        &  0.70 $\pm$ 0.01        &  0.74 $\pm$ 0.06       \\
DivideMix            &  0.72 $\pm$ 0.01        &  0.90 $\pm$ 0.01        &  0.68 $\pm$ 0.04        &  0.40 $\pm$ 0.02        &  0.92 $\pm$ 0.01        &  0.57 $\pm$ 0.08        &  0.94 $\pm$ 0.00        &  0.82 $\pm$ 0.01        &  0.78 $\pm$ 0.01        &  0.69 $\pm$ 0.01       \\

Pseudo v2 hard       &  0.67 $\pm$ 0.01        &  0.85 $\pm$ 0.01        &  0.68 $\pm$ 0.07        &  0.34 $\pm$ 0.06        &  0.85 $\pm$ 0.08        &  0.73 $\pm$ 0.08        &  0.87 $\pm$ 0.04        &  0.80 $\pm$ 0.01        &  0.66 $\pm$ 0.05        &  0.74 $\pm$ 0.05       \\
Pseudo v2 soft       &  0.68 $\pm$ 0.01        &  0.84 $\pm$ 0.01        &  0.73 $\pm$ 0.03        &  0.39 $\pm$ 0.08        &  0.90 $\pm$ 0.01        &  0.70 $\pm$ 0.05        &  0.91 $\pm$ 0.01        &  0.81 $\pm$ 0.01        &  0.76 $\pm$ 0.01        &  0.79 $\pm$ 0.07       \\
Pseudo v2 not        &  0.57 $\pm$ 0.05        &  0.62 $\pm$ 0.02        &  0.64 $\pm$ 0.08        &  0.40 $\pm$ 0.04        &  0.35 $\pm$ 0.13        &  0.56 $\pm$ 0.05        &  0.67 $\pm$ 0.09        &  0.68 $\pm$ 0.05        &  0.57 $\pm$ 0.09        &  0.67 $\pm$ 0.01       \\

		\end{tabular}
	}
	
\end{table*}
}

\newcommand{\tblDetailsThousandKl}{
\begin{table*}[tb]
	\caption{
		Details 1000\% Budget Kl
	}
	\label{tbl:detailsThousandKl}
	\resizebox{\textwidth}{!}{
		\begin{tabular}{l c c c c c c c c c c c c c c c c}
			\toprule
			
			Dataset      & Benthic              & CIFAR10H             & MiceBone   & Pig          & Plankton             & QualityMRI           & Synthetic            & Treeversity\#1       & Treeversity\#6       & Turkey               \\
           \\
                     \midrule
                     
Baseline             &  0.75 $\pm$ 0.03        &  0.26 $\pm$ 0.02        &  0.23 $\pm$ 0.02        &  0.54 $\pm$ 0.04        &  0.20 $\pm$ 0.02        &  0.31 $\pm$ 0.14        &  0.04 $\pm$ 0.00        &  0.36 $\pm$ 0.02        &  0.33 $\pm$ 0.01        &  0.24 $\pm$ 0.06       \\
ELR+                 &  0.75 $\pm$ 0.05        &  0.28 $\pm$ 0.02        &  0.34 $\pm$ 0.04        &  0.73 $\pm$ 0.02        &  0.27 $\pm$ 0.03        &  1.26 $\pm$ 0.56        &  0.49 $\pm$ 0.06        &  0.45 $\pm$ 0.01        &  0.64 $\pm$ 0.10        &  0.50 $\pm$ 0.13       \\
Het                  &  1.19 $\pm$ 0.05        &  0.38 $\pm$ 0.02        &  0.47 $\pm$ 0.03        &  0.87 $\pm$ 0.33        &  0.36 $\pm$ 0.04        & N/A                  &  0.26 $\pm$ 0.03        &  0.48 $\pm$ 0.03        &  1.25 $\pm$ 0.07        &  0.39 $\pm$ 0.03       \\
SGNP                 &  1.15 $\pm$ 0.04        &  0.36 $\pm$ 0.02        &  0.48 $\pm$ 0.13        &  1.02 $\pm$ 0.26        &  0.35 $\pm$ 0.04        &  0.25 $\pm$ 0.18        &  0.21 $\pm$ 0.03        &  0.49 $\pm$ 0.01        &  1.06 $\pm$ 0.04        &  0.47 $\pm$ 0.10       \\
DivideMix            &  0.83 $\pm$ 0.05        &  0.37 $\pm$ 0.03        &  0.35 $\pm$ 0.05        &  1.04 $\pm$ 0.25        &  0.34 $\pm$ 0.01        &  1.34 $\pm$ 0.84        &  0.43 $\pm$ 0.04        &  0.46 $\pm$ 0.01        &  0.77 $\pm$ 0.03        &  0.41 $\pm$ 0.07       \\

Pseudo v2 hard       &  0.95 $\pm$ 0.06        &  0.42 $\pm$ 0.01        &  0.45 $\pm$ 0.09        &  0.67 $\pm$ 0.02        &  0.43 $\pm$ 0.17        &  1.02 $\pm$ 0.27        &  0.20 $\pm$ 0.09        &  0.48 $\pm$ 0.04        &  1.00 $\pm$ 0.05        &  0.42 $\pm$ 0.10       \\
Pseudo v2 soft       &  0.70 $\pm$ 0.02        &  0.43 $\pm$ 0.02        &  0.22 $\pm$ 0.02        &  0.61 $\pm$ 0.07        &  0.25 $\pm$ 0.01        &  0.12 $\pm$ 0.04        &  0.06 $\pm$ 0.01        &  0.37 $\pm$ 0.02        &  0.35 $\pm$ 0.01        &  0.21 $\pm$ 0.02       \\
Pseudo v2 not        &  1.04 $\pm$ 0.13        &  1.04 $\pm$ 0.03        &  0.31 $\pm$ 0.05        &  0.57 $\pm$ 0.01        &  2.42 $\pm$ 0.98        &  0.16 $\pm$ 0.03        &  0.72 $\pm$ 0.32        &  0.66 $\pm$ 0.07        &  0.60 $\pm$ 0.09        &  0.35 $\pm$ 0.04       \\

		\end{tabular}
	}
		
\end{table*}
}

\title{Is one annotation enough?\\{\small A data-centric image classification benchmark for noisy and ambiguous label estimation}}

\author{%
Lars Schmarje$^{1}$\thanks{Corresponding Author, \texttt{las@informatik.uni-kiel.de}} \quad Vasco Grossmann$^{1}$ \quad Claudius Zelenka$^1$ \quad Sabine Dippel$^2$ \quad Rainer Kiko$^3$ \\
\textbf{Mariusz Oszust}$^4$ \quad \textbf{Matti Pastell}$^6$ \textbf{Jenny Stracke}$^5$ \quad \textbf{Anna Valros}$^7$ \quad \textbf{Nina Volkmann}$^7$ \\ \textbf{Reinhard Koch}$^1$\\
$^1$MIP, Kiel University \quad $^2$ITT, Friedrich-Loeffler-Institut \quad $^3$LOV, Sorbonne Université \\
$^4$Rzeszow University of Technology  \quad $^5$ITW, University Bonn \quad $^6$University of Helsinki \\
$^7$Luke, Natural Resources Institute Finland \quad   $^8$WING, University of Veterinary Medicine Hannover
}
\begin{document}
	
	\maketitle
	
	\begin{abstract}
		
	High-quality data is necessary for modern machine learning.
    However, the acquisition of such data is difficult due to noisy and ambiguous annotations of humans.
    The aggregation of such annotations to determine the label of an image leads to a lower data quality.
    We propose a data-centric image classification benchmark with ten real-world datasets and multiple annotations per image to allow researchers to investigate and quantify the impact of such data quality issues.
    With the benchmark we can study the impact of annotation costs and (semi-)supervised methods on the data quality for image classification by applying a novel methodology to a range of different algorithms and diverse datasets.
    Our benchmark uses a two-phase approach via a data label improvement method in the first phase and a fixed evaluation model in the second phase.
    Thereby, we give a measure for the relation between the input labeling effort and the performance of (semi-)supervised algorithms to enable a deeper insight into how labels should be created for effective model training.
    Across thousands of experiments, we show that one annotation is not enough and that the inclusion of multiple annotations allows for a better approximation of the real underlying class distribution.
    We identify that hard labels can not capture the ambiguity of the data and this might lead to the common issue of overconfident models.
    Based on the presented datasets, benchmarked methods, and analysis, we create multiple research opportunities for the future directed at the improvement of label noise estimation approaches, data annotation schemes, realistic (semi-)supervised learning, or more reliable image collection.
    \footnote{The source code is available at \url{https://github.com/Emprime/dcic}.\\
    The datasets are available at \url{https://doi.org/10.5281/zenodo.7152309}.}
	\end{abstract}
	
	\section{Introduction}

High-quality data is the fuel of modern machine learning and almost all models improve with higher quality data \cite{are_we_done,relabelImagenet,Northcutt2021pervasiveErrors}.
Therefore, such data are a key component for developing future techniques.
The acquisition of a large amount of data is considered particularly challenging due to the participation of humans in the process.
Their mistakes or subjective interpretations of annotation tasks can lead to \emph{noisy} or \emph{ambiguous} labels, respectively \cite{cifar10h,planktonUncertain,foc,inter-observer-segmentation,mammo-variability,tailception,eye-fuzzy,cancergrading}.
Consequently, the labels suffer from heteroscedastic aleatoric uncertainty which means that the data contains inherent noise, which is class- or even sample-dependent and negatively affects the quality \cite{collier2020simple}.

In \autoref{fig:teaser}, we present the impact of this uncertainty on the class "cat" in the CIFAR-10 dataset \cite{cifar}.
While all images have the same ground truth label in CIFAR-10, humans created agreeing annotations only with varying rates from four to 100 percent \cite{cifar10h}.
This means that individual annotations can be expected to be noisy as they diverge from the majority opinion.
Furthermore, a majority vote across multiple annotations can not capture the ambiguity between different images.
In some extreme cases (red borders), we even see a disagreeing majority vote across all annotators from the expected ground truth class.
We raise the question if all images should be treated equally if human annotations show such varying certainties.
Taking a \emph{data-centric} perspective \cite{data-centric-appraoch,schmarje2021datacentric,Marcu2021datacentricMyths,Tony2012FourthParadigm}, we investigate the data in contrast to only the model for answering this question. 
Specifically, we propose a data-centric image classification (DCIC) benchmark that indirectly measures a method's ability to identify noisy and ambiguous labels and correct them.
DCIC consists of ten real-world datasets of different domains (see \autoref{fig:datasets}) and multiple human annotations for each image.
The benchmark focuses on a data-centric view of the image classification problem by separating the data quality improvement and the classification performance into two tasks.

The main structure of the benchmark is divided into a \emph{Labeling} and an \emph{Evaluation} phase (see \autoref{fig:outline-0}) which is comparable to established Teacher-Student-Approaches \cite{mean-teacher,temporal-ensembling}.
Using this denotation, the benchmarked method will, as a teacher, improve labels during the first phase.
These are then benchmarked in the second phase by analyzing their quality as training input to a student model.
Be aware that we do not allow a knowledge transfer from the second phase to the first phase.

In detail, during the Labeling phase, we use samples from the distribution of the above-mentioned annotations to get different realistic label estimates as an \emph{initialization}.
The task of the benchmarked method is to improve these estimates for better performance of an other classification model in the second phase.
In that phase (Evaluation), the obtained labels are used as input for training a fixed model and its performance is measured on a testing subset of the original data.
In contrast to common model-centric deep learning approaches (see \autoref{fig:outline-1}), we can vary the initialization for the same method and better separate its performance from the data improvement.
 The fixed model is used for the evaluation to facilitate distinguishing between performance gains due to improved input data and better learning of the method itself.

\twoPicSideBySide{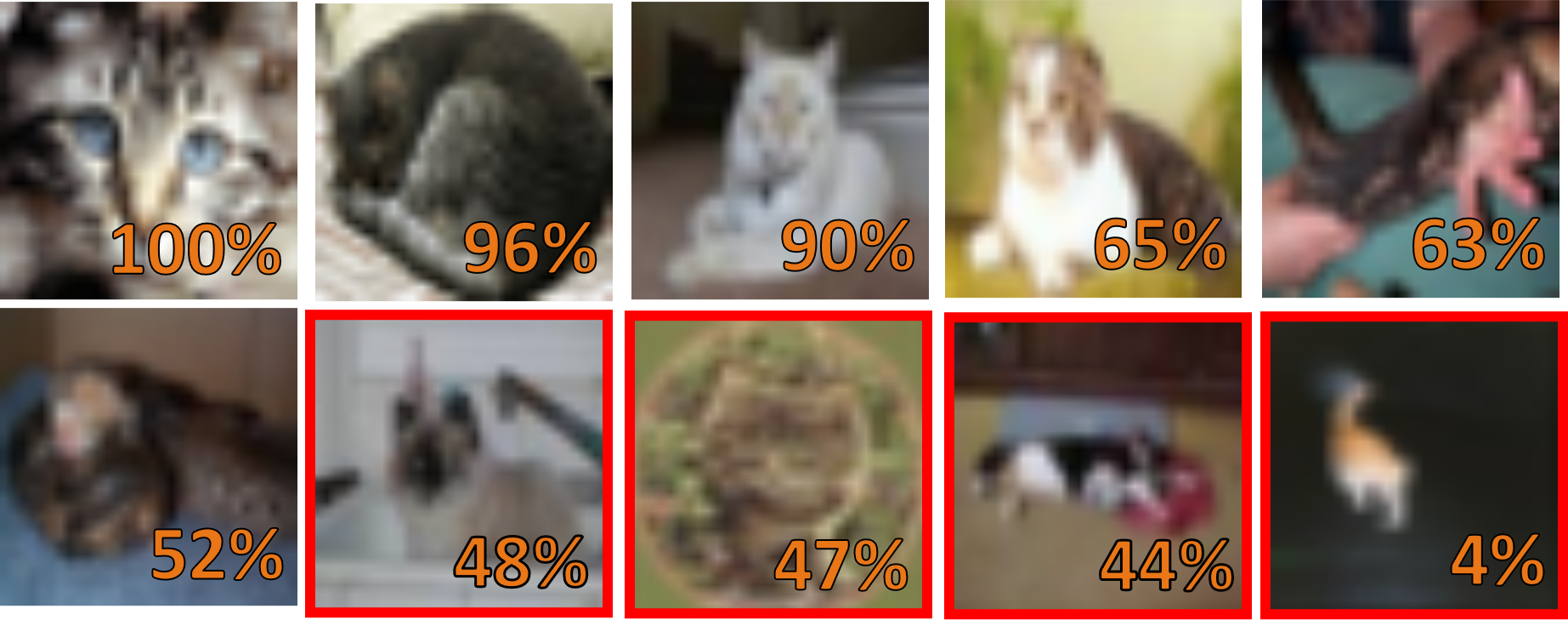}{Are all images showing a cat? --
Based on their ground truth labels from CIFAR-10 \cite{cifar} they should all be cats. 
However, we give the agreement rate with the class cat from \cite{cifar10h} in the lower right corner and see a wide range from four to 100\%.
Based on a majority vote, the last images (red border) would have not to be labeled as a cat but as dog, frog, dog, and deer, respectively.
Based on these observations, we answer in our paper the question of whether all images should be treated equally as cats or if we should use multiple annotations and the resulting soft labels to capture this intrinsic noise and ambiguity.
}{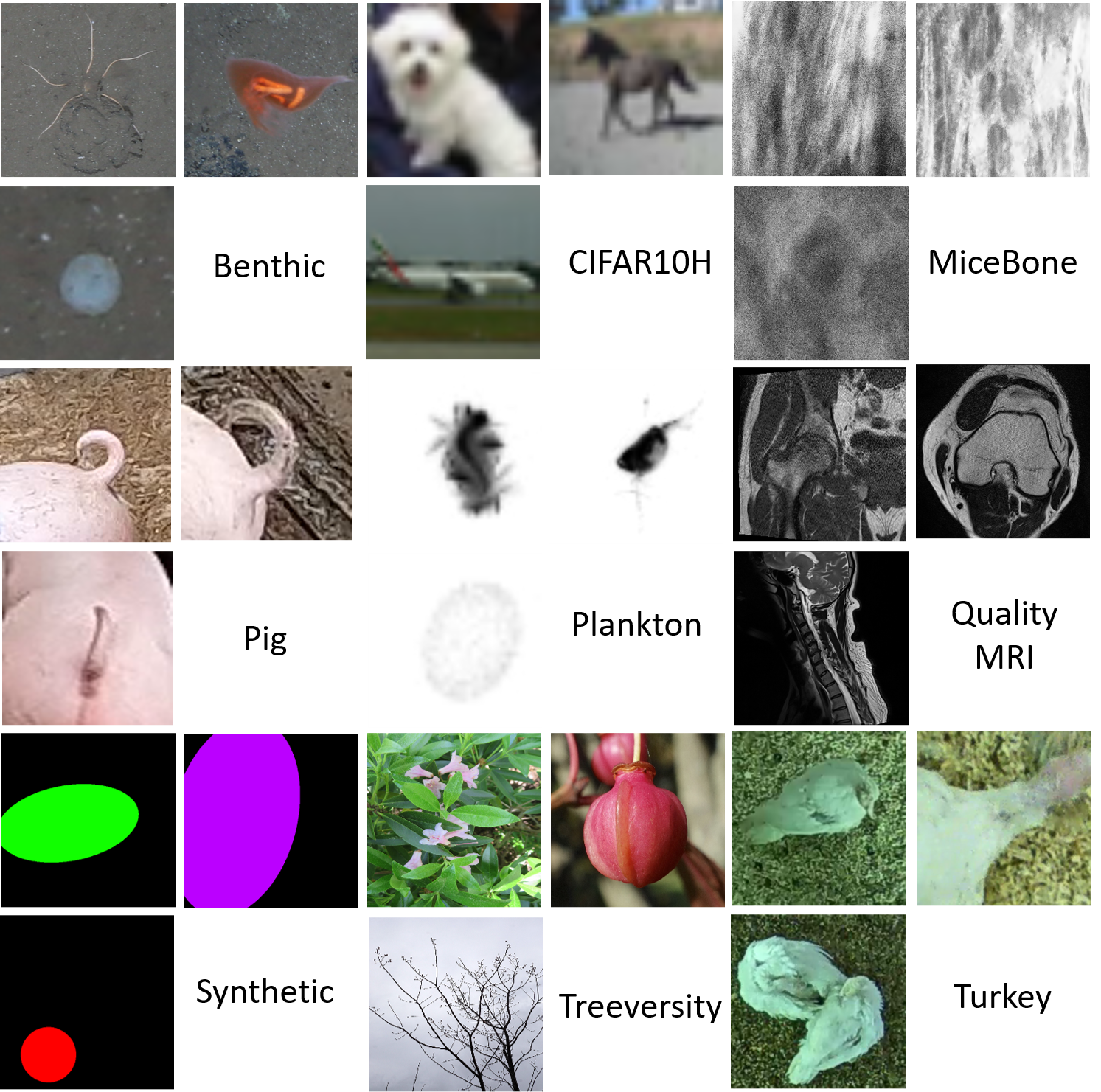}{Three example images for all datasets. Details about the statistics of the dataset are given in \autoref{tbl:datasets}.}

Our benchmark is not only useful to evaluate existing methods, but will support research into algorithms for realistic datasets.
Especially, it can bridge the research between semi-supervised learning and noise estimation based on realistic ambiguous noise patterns. 
We provide multiple algorithms as baselines and support the integration of more algorithms by common dataloaders for the two most popular deep learning frameworks: Tensorflow \cite{tensorflow2016-whitepaper} and Pytorch \cite{pytorch2019Neurips}.
We analyzed thousands of combinations of baseline methods, different initializations, and datasets.
The obtained results confirm that the improvement of data quality leads to performance gains.
Additionally, we investigated factors that influence the data quality and identified trends that lead to better learning of the underlying distribution.

\picTwoExtend{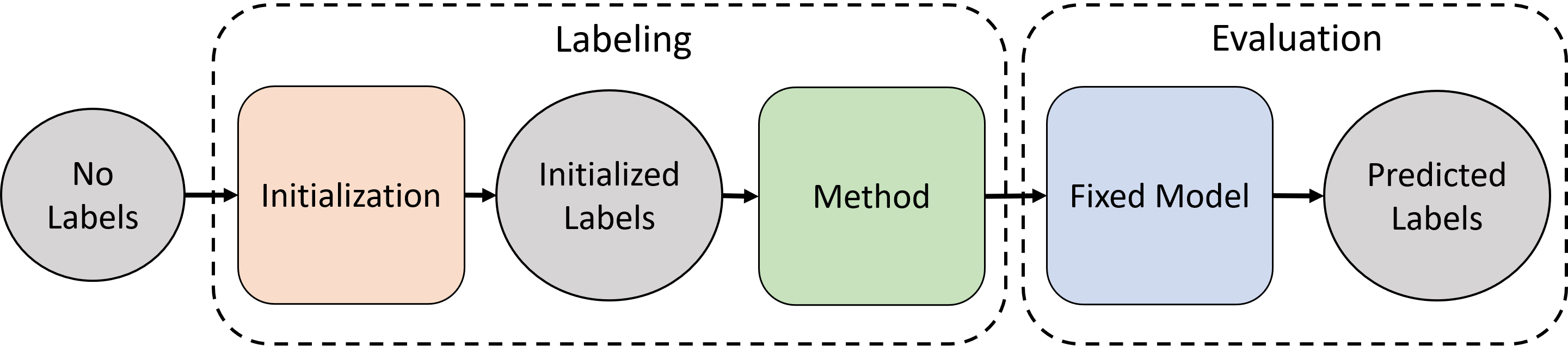}{Data-Centric (Ours)}{Model-Centric (Common)}{Comparison of our data-centric approach with the commonly used model-centric approach.
The circles and arrows represent the available label information in addition to the corresponding images.
The squares represent the methods which generate / change these label.
There are two main differences between our and the common model-centric approach.
Firstly, we also look at how the raw unlabeled data is initialized and thus how many annotations are required.
Secondly, we use a fixed model to evaluate the output of the benchmarked method.
These differences lead to a greater separation of data quality and method performance on the final scores on the predicted labels.}

	Our key contributions are: 
	(1)  We collected and created ten real-world image classification datasets with multiple annotations per image.
    These annotations allow a realistic simulation of noise patterns and will be helpful for future research in machine learning on real world data sets.
    (2) We provide a multi-domain benchmark based on these datasets for noisy and ambiguous label estimation.
        We implemented 20 methods as comparison.
     The benchmark also covers the topic of cost and bridges research between semi-supervised learning and ambiguous label estimation.
    (3) We show that one annotation per image is not enough because model performance improves as more labels are given for each input.
        We identify that the current focus on hard labels for classifications is ill-suited to learn the underlying ground truth distribution.
        A change in data preprocessing especially in annotation protocols could mitigate this and lead to less overconfident models.

	\subsection{Related Work}
	\label{subsec:related}

	Human annotations of one image can differ due to complex reasons.
	Next to mere individual errors, cognitive sciences have shown that human judgement under uncertainty is driven by a subjective bias and the context of the annotation process~\cite{Schustek2018instance}.
	As labeling relies on human perception, data quality problems, including issues with noisy and ambiguous labels, have been broadly discussed in the literature \cite{song2022noisyLabelsSurvey,surveynoise,cifar10h,Wei2021cifar10n,merIssue}.
	While voting strategies have proven to be robust tools to remove outlying annotation errors in a single label scenario, they also eliminate subjective disagreement, even in cases with more than just one valid interpretation~\cite{rajeswar2022multi}.
	The impact of this information loss has been discussed in numerous studies, indicating limitations in capturing ground truth by just one label~\cite{Uma2021SurveyDisagreement,Davani2021BeyondMajority,grossmann2022beyond,Basile2021ConsiderDisagree}. 
	We empirically support these arguments across 9 datasets and 20 methods.
	
As label noise can severely degrade the classification performance~\cite{Northcutt2021pervasiveErrors}, learning with flawed training data has become a substantial field of research in which numerous strategies have been proposed:
while sample selection methods separate clean and noisy data by evaluating small-loss or disagreement~\cite{yao2019searching,yu2019does}, correction methods aim at relabeling wrongly assigned labels by either learning class prototypes~\cite{han2019deep} or by pseudo-labeling strategies that utilize confident predictions~\cite{pseudolabel}.
Multiple methods have been proposed, but are often evaluated on synthetic noise \cite{Liu2020elr,divide-mix}.
However, Wei et al. showed that synthetic noise is different from real noise by humans, which limits the generality of findings \cite{Wei2021cifar10n}.
Gao et al. proposed synthetic annotators with individual labeling behavior instead of random noise
to reduce uncertainty in predictions
~\cite{Gao2022SyntheticAnnotators}. 
We go beyond this by using human annotations to reproduce realistic noise pattern and we do not only look at annotation errors but also at ambiguous annotations from subjective interpretations.
	
Datasets like CIFAR-10H \cite{cifar10h} and CIFAR-10N \cite{Wei2021cifar10n} address the problem of realistic noise by providing multiple annotations per image for example of the original CIFAR-10 dataset. By doing so, both publications demonstrate an improved performance and a higher robustness, while also claiming that further research in dealing with human noise is still inevitable.
The utilization of soft label distributions instead of hard one-hot label encodings enables the detailed representation of subjective disagreement and improves the generalization with ambiguous datasets~\cite{cifar10h,bagherinezhad2018label,krothapalli2020adaptive}.
In our benchmark, we extend this idea to eight diverse datasets apart from CIFAR-10 for a broader evaluation.

If we want to use multiple annotations per image, we need to consider the cost of such annotations to make it feasible in a project.
Current research such as semi-supervised learning \cite{simclrv2,fixmatch,dc3,survey} could be used to analyze only a portion of the data with multiple annotations.
However, approaches which combine noisy labels with semi-supervised learning have not been extended to real-world image classification tasks or do not consider the possibility that one labels is not enough to capture the ambiguity of subjectivity~\cite{Li2014SemisupervisedAmbiguous,Zhang2021Codim,Wang2020seminll}.
We provide with our benchmark the datasets and infrastructure to bridge the research between semi-supervised learning and noise estimation.

Prediction uncertainty can be attributed to unexplainable noise in the given training or test dataset (aleatoric uncertainty) or a wrong model inference (epistemic uncertainty) and it can be difficult to approximate and differentiate between them~\cite{Sambyal2021MedicalUncertainty,Abdar2021surveyUncertainty,Valdenegro-Toro2022ApproximateUncertainty,Kendall2017Uncertainty}.
Several real-world noisy datasets have been utilized as a foundation for classification benchmarks~\cite{li2017webvision,lee2018cleannet,cifar10h}, Song et al. provide a current survey on datasets and methods~\cite{song2022noisyLabelsSurvey}.
Moreover, most robust methods are evaluated based on the test set accuracy \cite{divide-mix,song2022noisyLabelsSurvey,Wei2021cifar10n,santarossa2022medregnet,Zhang2020SevereNoise}.
However, even a small change in the structure or parameters of a method can directly impact its performance, limiting the comparability \cite{revisiting-self}.
Other fields, such as Bayesian Neural Networks, address this issue by comparing results to statistical simulations, for example \cite{izmailov2021bayesianReally}.
A recent benchmark \cite{uncertaintyBaselines} tries to overcome this issue by providing a baseline for noisy labels as a form of uncertainty estimation \cite{collier2020simple}.
However, this benchmark relies on synthetic noise or noisy datasets without knowledge about the underlying ground truth distributions \cite{li2017webvision}. 
We use a data-centric approach to minimize the impact of implementation detail differences and measure the impact of the data indirectly during the Labeling phase by evaluating on a fixed model.

	\section{Benchmark}
	\label{sec:benchmark}
	
	Our benchmark is divided into two major phases: \emph{Labeling} and \emph{Evaluation}.
	In alignment with the Data-Centric Idea \cite{data-centric-appraoch}, we separate the improvement of the data (Labeling) from the improvement of the models (Evaluation).
	The benchmark can be utilized to analyze a variety of research questions, but we focus on evaluating methods that estimate noisy or ambiguous labels.
	
	We use the terms \emph{noisy} and \emph{ambiguous} throughout this work synonymously because we often do not differentiate between their cause during the annotation process. 
	As mentioned above, these causes are errors or mistakes of human annotators which can be recovered for noisy label or noise. 
	Subjective interpretations, imprecise task descriptions or poor image quality lead to ambiguous labels.
	
	In general, we have an image dataset $X$ with $k$ known classes and use human annotations to approximate the image labels.
	Each image $x \in X$ has an often unknown soft ground truth label $\hat{l_x} \in [0,1]^k$.
	Therefore, we use $N$ hard human annotations $a_i \in \{0,1\}^k$ with $i \in {1,...,N}$ as estimates of $\hat{l_x}$.
	We assume that an average of annotations ($l_x = \sum_{i=0}^N \frac{a_i}{N}$) is an approximation of this target label $\hat{l_x}$ as in \cite{dc3}.
	Based on this definition, an annotation $a_i$ or hard label sampled from the distribution $l_x$ are in general of lower quality because they can not capture the aleatoric uncertainty of the soft label $l_x$.
	We split the data equally and randomly in five \emph{folds} and ensure a similar class distribution between the folds as best as possible.
	For one run, we use three folds as training ($X_T$) and one fold each as validation ($X_V$) and test ($X_E$) data, respectively. %
	We call such an assignment of folds to the training, validation and test data \emph{slice}.

     \paragraph{Labeling}
    The Labeling phase consists of two steps. 
    In the first step an initialization is used to get label estimates and in the second step, the benchmarked method $\Theta$ aims to improve these labels.
    As initialization, we acquire $m \in \mathbb{N}$ annotations for $n \in [0,100]$ percent of the training and validation images.
    
     We call the total number of required annotations \emph{budget} $b = m \cdot n$ and report it as proportions of training and validation images ($|X_T \cup X_V|$).
     In general, a classification task gets easier with more annotations or a higher budget.
     Be aware that the same initialization results in the the same budget while the same budget can achieved by different initializations. 
     
     The used initialization schemes per method are defined later in this section.
     We chose fixed initialization schemes for better comparability between the methods.
     How these labels are improved by the method $\Theta$ is not restricted.
     However, annotations aside from the given initialization are not allowed to be used. 
     Since we measure the quality by training a different fixed network in the next phase, a good label would be presumably as close as possible to $l_x$.

     \paragraph{Evaluation}
     \label{subsec:evaluation}
     
     In the Evaluation phase, the model and its hyperparameters are fixed to measure only the impact of the provided labels ($\Theta(x)$).
     The training of this fixed model $\Phi$ is calculated on the provided $\Theta(x)$ with $x \in X_T$.
     The best network parameters during training are selected based on a minimal divergence between $\Phi(x)$ and $\Theta(x)$ with $x \in X_V$.
     The generalization is then tested by measuring the difference between $\Phi(x)$ and $l_x$ for $x \in X_E$.

     \paragraph{Metrics}
     
     Kullback-Leibler divergence ($KL$) \cite{kullback} between $\Phi(x)$ and $l_x$ for $x \in X_E$ has been used as our main metric since it is an established method to measures the difference between two distributions \cite{murphy2012machinelearning}.
     We averaged in a 3-fold cross-validation per dataset for a high reproducibility.
     We used the three slices defined by $X_{V_i} = \{f_{i+1} \}$, $X_{E_i} = \{f_{i+2}\}$ and the rest as training ($X_{T_i} = \{f_{i},f_{((i+2)\%5)+1},f_{((i+3)\%5)+1}, \}$ with $\%$ for modulo) for the folds $f_1, ..., f_5$ with $i \in {1,2,3}$ as the index of the slices. 
      While $KL$ directly allows to measure the desired distribution divergence, we provide additional metrics as comparisons.
     We evaluate the accuracy ($ACC$) and F1-Score ($F1$) between $\Phi(x)$ and $l_x$ for $x \in X_E$ per class and report the mean across the classes, which is commonly called the macro value and allows evaluation even in the presence of class imbalance.
     We used the most likely class based on the evaluated distributions for these metrics.
     We analyze the calibration of the models by reporting the Expected Calibration Error ($ECE$) \cite{guo2017ece}.
     As reference, we provide all of these metrics also on the difference between the proposed label before the second training $\Theta(x)$ and the expected ground-truth $l_x$ for $x \in X_E$. %
     The metrics are noted as $\hat{ACC}, \hat{F1}$ and $\hat{ECE}$.
     We report the Cohen's Kappa Score ($\kappa$) \cite{kappa} as the measurement of the consistency of $\Theta(x)$  between the folds because more consistent labels result in higher model performance.
     
	\paragraph{Datasets}
	\label{subsec:datasets}

	We include ten real-world classification datasets in our benchmark.
	Since we need multiple annotations per image for the evaluation of the quality of labels and  this information is often not available in existing datasets or insufficient for our benchmark,  we collected, adopted, or extended annotations of the following datasets.
	Their details are shortly described below, while their properties and exemplary images are shown in \autoref{tbl:datasets} and   \autoref{fig:datasets}, respectively.
    As presented, the datasets vary across all reported properties, giving an opportunity to comprehensively evaluate considered methods.
    More challenging datasets are characterized by a high-class imbalance, a low average agreement, or a low number of annotations per image.
    Detailed reports about the collection process and remaining dataset specifics are given in the supplementary.
	
	\tblDatasets
	\emph{1. Benthic} depicts images from the seafloor and consists of underwater flora and fauna.
	We used annotations from \cite{schoening2020Megafauna,Langenkamper2020GearStudy} but filtered for at least three annotations per object and cropped the main image to this object.
	We combined classes with too few images in agreement with domain experts.
	\emph{2. CIFAR-10H} is a variant of CIFAR-10 \cite{cifar} introduced in \cite{cifar10h}. 
	Peterson et al. analyzed the underlying class distribution like us by reannotating the CIFAR-10 test set.
	In contrast to other variants like CIFAR-10N \cite{Wei2021cifar10n}, this dataset provides more annotations per image.
	\emph{3. MiceBone} consists of Second-Harmonic-Generation images of collagen fibers in mice \cite{schmarje2019}. 
	The raw images were preprocessed as described in \cite{dc3}.
	Since there is a need for multiple annotations per image, we hired workers to increase their number by a factor of five. %
	\emph{4. Pig} consists cropped tail images from European farms. The annotations were collected by hired workers with high domain knowledge. The goal is the classification of the injury degree of the tail.
	\emph{5. Plankton} is a collection of underwater plankton images with multiple annotations from citizen scientists \cite{foc}.
	We use the preprocessing described in \cite{dc3}.
	\emph{6. QualityMRI} consists of human magnetic resonance images (MRI)  with a varying quality and multiple subjective quality ratings  gathered in tests with radiologists.
	It was introduced and evaluated in \cite{obuchowicz2020qualityMRI,stepien2021cnnQuality}.
	\emph{7. Synthetic} dataset was generated for the purpose of this study.
	It consists of images that contain one blue, red, or green circle or ellipse on a black background. 
	To create ambiguous images, we added color and axis interpolations of these classes.
	\emph{8+9. TreeVersity} is a publicly available crowdsourced dataset of plant images from the Arnold Arboretum of Harvard University\footnote{\url{https://arboretum.harvard.edu/research/data-resources/}}.
	In the crowdsourcing project, the images were tagged with a given set of labels.
	We used a simplified version with six classes where we combined classes with too few images.
    Only images with at least three tags were used. 
    Tags are not the same as class labels, therefore, we provide two subsets of TreeVersity. 
    In TreeVersity\#1, we filtered for exactly one given tag of the six possible ones per user which is similar to a classification. 
    In TreeVersity\#6, we filtered for a maximum of six different tags which means we did not apply any restrictions.
	\emph{10. Turkey} is a dataset with images of turkeys and their injuries \cite{volkmann_turkeys,volkmann2022keypoint}.
	We used the preprocessing described in \cite{dc3} and  extended the original annotations, increasing their number by a factor of five with hired workers.

	\paragraph{Methods}
	\label{subsec:methods}
	
	We compare a variety of recent supervised, self-supervised, and semi-supervised algorithms against our baseline.
	The baseline does not adjust the initialized dataset in the first phase in any way and just forwards these labels to the supervised training of the second phase.
	Thus it is equivalent to supervised learning in a model-centric benchmark.
	We selected the other methods based on their recency, access to authors code or reimplementations and if they are state-of-the-art or commonly used as comparisons in the literature.
	The \emph{supervised} methods are Heteroscedastic \cite{correlated_label_noise}, SNGP \cite{liu2020sngp}, and ELR+ \cite{Liu2020elr}.
	The \emph{semi-supervised} methods are Mean-Teacher \cite{mean-teacher}, $\pi$-Model \cite{temporal-ensembling}, FixMatch \cite{fixmatch}, DC3 \cite{dc3}, Pseudo-Label \cite{pseudolabel}, and DivideMix \cite{divide-mix}.
	The \emph{self-supervised} methods are BYOL \cite{byol}, MOCOv2 \cite{mocov2}, SimCLR \cite{simclr}, and SWAV \cite{swav}.
	Detailed descriptions about most of them are given in \cite{survey} and their key characteristics are presented in \autoref{tbl:methods}.
	We use the reported hyperparameters for Imagenet \cite{imagenet} or Webvision \cite{li2017webvision} by the original authors to ensure a comparison out-of-the-box across different image domains.
	For DC3\cite{dc3}, we investigated the combinations with Mean-Teacher, $\pi$-Model, FixMatch, and Pseudo-Label.
	For Pseudo-Label, we used two different implementations (v1 and v2) and variants with or without pretraining and soft or hard labels as input.
	In total, this results in 20 investigated methods.
	For better referencing, we group them as described above but put methods that \emph{use soft labels} into their own group.
	
	\tblMethods
	
	\paragraph{Initialization Schemes}
	\label{subsec:init}
	
	We investigated a fixed set of initialization schemes and note them $m$-$n$ for $m$ annotations for a subset of data with a relative size $n$.
	For easier reference, we group them as 
	\begin{itemize}
	\item \emph{Supervised Learning (SL)} 01-1.00
	\item \emph{Supervised Learning+ (SL+)} 03-1.00, 05-1.00, 10-1.00
	\item \emph{Semi-Supervised Learning (SSL)} 01-0.10, 01-0.20, 01-0.50
	\item \emph{Semi-Supervised Learning (SSL+)} 10-0.10, 05-0.20, 02-0.50

	\end{itemize}

	\paragraph{Implementation Details}
	\label{subsec:details}
	
	The final results depend on a good set of fixed hyperparameters like learning rate for the model $\Phi$ for each dataset during the evaluation.
    Therefore, we determined them by applying Hyperopt \cite{bergstra2013hyperopt} with 100 search trials across the same grid of parameters for all datasets. 
    The target was the minimization of $KL$ between $\Phi(x)$ and $l_x$ for the baseline experiment with exactly ten annotations per image across one slice.
    We executed these and later experiments on an Nvidia RTX 3090 with 24GB VRAM or comparable hardware.
    Some combinations of models and input sizes could not fit on this hardware and therefore were ignored to keep the needed hardware to a minimum.
    Details about the used parameter grid are given in the supplementary.
	We ensure that all folds are randomly generated, while restrictions about similar images are considered.
	Without these restrictions, similar images, e.g., frames from the same camera might lead to an information leakage between the folds which would negatively influence the interpretability of the results. 
	
	\section{Analysis}
	\label{sec:eval}
	
	The evaluation was conducted across combinations of all datasets, methods, initialization schemes, and slices.
	A complete cross-combination would result in 5400 experiments from which we selected 3456 experiments to save resources since some combinations would not add more insights e.g. due to inferior performance of similar methods.
	A detailed overview of the initialization schemes used per method can be found in \autoref{tbl:methods}.
	As shown in \autoref{tbl:datasets}, we have a large variability between the datasets, especially in $ACC$ and $\hat{ACC}$ that range from 36\% to 96\% for the baseline.
	Due to the fact that the baseline does not adjust the initialization, $\hat{ACC}$ can be seen as the performance of humans in improving the labels for the given budget. 
    The datasets Benthic, MiceBone, Pig, Treeversity\#6, and Turkey have an over 10\% lower $ACC$ than the expected $\hat{ACC}$, which marks them as particularly challenging for the model.
    Moreover, the datasets Benthic, MiceBone, Pig, QualitMRI, and Treeveritsy\#6 have an $\hat{ACC}$ of lower than 85\% which marks them as difficult even for humans.
    The $\hat{ACC}$ of Synthetic is even lower due to the artificially created labels.
    Due to this variability, an average across the scores can be misleading.
    For this reason, we report the median in this paper and report the full results including the standard error of the mean (SEM) in the supplementary.
   
   \paragraph{What metrics should we use?}
    
    We analyzed the correlations between our metrics to determine which contain the same or similar information and which are complementary.
    All calculated Pearson correlation coefficients  have a p-value < 0.01 and the four strongest correlations are $ACC$ vs. $F1$ (0.99), $KL$ vs. $ECE$ (0.68) $F1$ vs. $\kappa$ (0.77) and $ACC$ vs. $\kappa$ (0.77).
    The other correlations are around -0.5.
    Selected correlations are illustrated in \autoref{fig:2_correlations} and additional graphics and analysis are given in the supplementary.
  $F1$ balances the precision and recall but in our experiments we see almost identical values to $ACC$ which we credit to the averaging per class.
  This means we only need to concentrate on $ACC$ as a classification score.
    Overconfident models are a problem of modern machine learning \cite{guo2017ece} and the higher correlation between $KL$ and $ECE$ compared to any of the two with $ACC$, $F1$ or $\kappa$ indicates that our focus on classification metrics like $ACC$ and $F1$ could be the issue.
    Only metrics like $KL$ and $ECE$ consider the complete distribution and which justifies using mainly $KL$ for the evaluation of this benchmark.

      \picThree{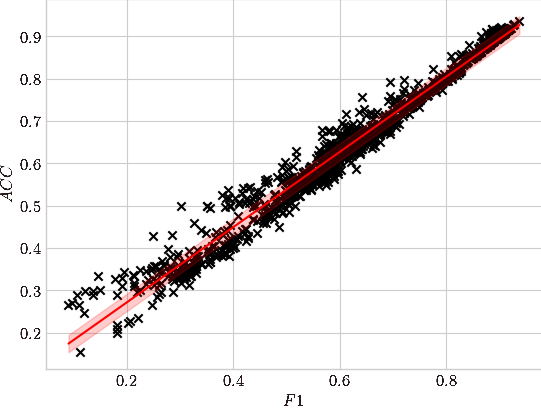}{$ACC$ vs. $F1$}{$KL$ vs. $ECE$}{$ACC$ vs. $ECE$}{Correlations between selected metrics across all experiments. The red line represents the linear regression between the metrics and the light red area the mean absolute error of the regression. 
      }

    \picThreePlus{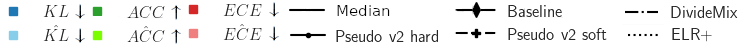}{Transfer Learning}{Quantity vs. Quality}{Method comparison}{Analysis of all or selected methods across different budgets or initialization schemes. For details about the definitions see \autoref{sec:benchmark}. 
    The orange crosses in (c) represent the best performance of Pseudo v2 soft with another initialization scheme see (b).
    The budget is increased by an raised portion of labeled data ($n$) until 1.00 and then increased further by using additionally multiple annotations per image ($m$). }
    
    \paragraph{One annotation is not enough}

    It is to be expected that more and better data should lead to increased performance which we quantify in \autoref{fig:trans_qual-0}.
    It can be seen that all metrics improve with an increased budget in the form of more annotated data or more annotations per image. 
    However, the impact is lower for more annotations per image. 
    For example, $ACC$ increases from around 55\% to 69\% for the full supervision. 
    Up to 10 annotations per image increase the score only to around 72\%. 
    This difference can be explained by the fact that additional annotations are most valuable to improve uncertain labels.
    In alignment with previous research \cite{Uma2021SurveyDisagreement,grossmann2022beyond,Basile2021ConsiderDisagree}, these results across thousands of experiments empirically justify that one annotation is not enough to capture the ground-truth of an item. 
    Some improvements can be gained from correction annotation errors via a majority vote but the high disagreement and low $\hat{ACC}$ of the baseline on some datasets support the hypothesis that ambiguous annotations are a main source for the improvement.
    This ambiguity can not be described with a single hard majority vote and thus highlights the importance of using soft labels.
    Additionally, we see that $KL$ is about half as small as $\hat{KL}$, $ECE$ is around 3-10\% lower than $\hat{ECE}$ and $ACC$ is 1-2\% better than $\hat{ACC}$.
   As shown previously by Hinton et al. \cite{Hinton2015Distillation,simclrv2} the knowledge distillation via a neural network into soft labels can be beneficial for $ACC$.
   We find the impact for metrics like $KL$ and $ECE$ even higher which supports our design of a two-phase benchmark.

    \paragraph{Limits of current state-of-the-art}
    
    To determine the best-performing state-of-the-art method, we gathered their relative improvements over the baseline in \autoref{tbl:results-0}.
    The best algorithm for each type of the soft, semi-supervised, supervised, self-supervised approaches based on the average performance across the budgets of 10\%, 100\%, and 1000\%  are Pseudo v2 soft, DivideMix, ELR+, and Mocov2, respectively.%
    We visualize the best three of them in \autoref{fig:trans_qual-2} and give detailed results across the datasets for the budget 100\% in \autoref{tbl:results-1}.
    The full results can be found in the supplementary.
    All top three methods are pretrained on ImageNet and outperform the rest in the field they were designed for.
    DivideMix is the best during partial supervision (budget < 100\%), ELR+ is more noise robust (budget > 100\%), and Pseudo v2 soft has the lowest $KL$ score (budget > 100\%).
    It is important to note that  ImageNet pretraining leads to improvements on many datasets (see Pseudo v2 not in the supplementary) but also to worse results on others.
    It needs to be investigated if other pretraining such as CLIP \cite{Radford2021Clip} or unsupervised pretraining on larger datasets \cite{simclrv2,byolv2} could improve on these results.
   Overall, the current state-of-the-art methods are insufficient for a label preprocessing across all domains.
    For a high budget any investigated method is worse than the supervised baseline without adjustments of the initialized dataset which shows the lack of appropriate algorithms for such budgets.
    
    Moreover, an average better performance does not mean that the gains are equal across all datasets. 
    For example, ELR+ has the lowest KL at a budget of 100\% for five out of ten datasets but on the QualityMRI dataset, it is among the worst methods.
    This means while some methods might work on some datasets they might not generalize to other datasets. 
    Overall, we see the highest $KL$ for the datasets Benthic, Pig, Quality MRI and Treeversity\#6 which also have the lowest agreement as seen in \autoref{tbl:datasets} except for the synthetic dataset.
    These datasets also show the largest variance in results across the methods. 
    We conclude that the impact of the data is larger than the impact of the current preprocessing of state-of-the-art methods.
    This highlights the importance for investigating the data and label generation more if they a more impactful than the method itself.
    
    While a higher budget leads to improved metrics, it also matters how it is used.
    In \autoref{fig:trans_qual-1}, we investigated the impact on $KL$ and $ACC$ for a budget of 100\% for Pseudo-Label using hard or soft labels. 
    We find that the accuracy is comparable between the methods and increases with a rising percentage of labeled data ($m$).
    For hard labels, the $KL$ improves equally. 
    If we use soft labels for training, we see lower results for 05-0.20 and 10-0.10.
    We conclude that we should investigate more how we distribute our budget if increasing it is not an option.

    \tblResults

	\section{Discussion}
	\label{sec:discussion}
	
	Overall, we can confirm several previous research hypotheses while identifying missing information and thus new research opportunities with our novel datasets and benchmark.
	
	We can demonstrate that data quality positively impacts the classifications scores like $ACC$ and $F1$ and distribution-based scores like $KL$ and $ECE$.
	Knowledge distillation can improve the approximation of the underlying distribution further. 
	We agree with previous research \cite{Uma2021SurveyDisagreement,Davani2021BeyondMajority,grossmann2022beyond,Basile2021ConsiderDisagree} that one annotation is not enough and we need to use soft labels to handle ambiguous data.
    $KL$ and $ECE$ are highly correlated (0.7) and are improved more when using soft labels.
	We believe that focusing on learning the real distribution and thus minimizing $KL$ can lead to less overconfident models. 
	Using soft labels as input seems to be crucial for achieving this since hard labels and classification metrics like $ACC$ lead to models which slightly ignore the real ground truth distribution. 

    Nevertheless, most of the investigated state-of-the-art method do not use soft labels and often interpret noise only as errors in the annotation process. 
    These issues need to be addressed in future research and a simple method like Pseudo v2 soft illustrates how the $KL$ score can be lowered with this approach.
    For the largest budget, the baseline is the best model and even special noise estimation algorithms like ELR+ \cite{Liu2020elr} and SNGP \cite{liu2020sngp} can not achieve better results.
    We see a high variance across the datasets for different methods in our benchmark.
    However, we need methods which work across a variety of domains out-of-the-box to allow an easy application to current research question in other domains.
    Another practical issue is that we need to find solutions for acquiring soft labels even with a limited budget.
    In many research projects, it is difficult to annotate thousands of images with domain experts and annotating them multiple times would only increase the costs further.
    Thus, we need to bridge the research in semi-supervised learning and ambiguous and noise estimation.
    Such combinations could allow the usage of soft labels on a subset of images and simultaneously determine annotation errors.
    Our benchmark and datasets allowed the identification of these issues and thus could also be used to research new methods to solve these issues.
    We are confident that our benchmark and datasets can facilitate the bridged research on the topic of semi-supervised learning and ambiguous image estimation for real world image classification problems.

	\paragraph{Impact}
	\label{subsec:impact}
	
	This work as a benchmark provides ten datasets and a detailed evaluation across 20 algorithms on this benchmark. 
	The provided data can allow the investigation of research questions on the topics of e.g. noise estimation, data annotation scheme, or realistic semi-supervised learning.
	This work is intended to allow and spark future research and thus no direct social impacts are expected.
	However, this basic research is time and resource-consuming.
	For the final evaluation, we conducted experiments with about 5500 GPU hours which equals around 600kg $CO_2$.
	For this reason, we limited the evaluation always to necessary elements when possible in order to not increase the needed GPU hours further.
	
	\paragraph{Limitations}
	\label{subsec:limitations}
	
	The 20 investigated algorithms are only evaluated with one fixed set of hyperparameters across different datasets during the labeling phase. 
	For optimal performance, a tuning per algorithm would have been required. 
	We were interested in the general performance out-of-the-box and therefore neglected this issue due to resource minimization.
	The researched datasets are all below 15,000 images and the unsupervised learning potential on millions of images could not be investigated. 
	We want to provide a detailed analysis in relation to the underlying distribution $l_x$ per image which is only possible with multiple annotations per image.
	For larger datasets, this effort was just not feasible.
	Classification with hundreds or more classes are also not feasible because the annotation costs increase with the number of classes.
	As described above we conducted more than a thousand experiments but we had to select and combine several results in a comprehensive manner in this paper. 
	These aggregations can not capture all details.
	Much more detailed analysis e.g., per dataset would be possible and thus we included all raw results in the supplementary.
	Due to the fixed initialization scheme, we can not investigate active learning approaches.
	However, this restriction is chosen to allow a better comparability and future researchers could decide against such a restriction.
	
	\paragraph{Conclusion}
	
	In alignment with previous research, we show that one annotation is not enough to handle ambiguous and noisy images and their underlying ground truth distribution.
	Multiple annotations and some kind of soft label are required to capture the difference in the images. 
	Future research needs to investigate in more detail how annotations are being created, including annotation costs. 
	We show that current state-of-the-art can help under certain budget or dataset constraints. 
	However, methods with consistent results across a variety of datasets and budgets are missing.
	We release all datasets and the benchmark publicly to enrich future research on these topics.

	\begin{ack}
	
	We thank Mark Collier for his valuable feedback and discussion about the benchmark.
	We thank the annotators Kristina Ahlqvist, Daniel Grundig, Stine Heindorff, Jana Krambeck, Kathrin Körner,  Richard Lange, Miina Tuominen-Brinkas and Emirhan Ustalar for their valuable work. 
	
	We acknowledge funding of L. Schmarje by the ARTEMIS project (Grant number 01EC1908E) funded by the Federal Ministry of Education and Research (BMBF, Germany).
	R. Kiko also acknowledges support via a “Make Our Planet Great Again” grant of the French National Research Agency within the “Programme d’Investissements d’Avenir”; reference “ANR-19-MPGA-0012”.
	V. Grossmann is employed with funds provided by Kiel Marine Science (KMS) and Future Ocean Network (FON) by Kiel University.
	Funds to conduct the PlanktonID project were granted to R. Kiko and R. Koch (CP1733) by the Cluster of Excellence 80 “Future Ocean” within the framework of the Excellence Initiative by the Deutsche Forschungsgemeinschaft (DFG) on behalf of the German federal and state governments.
	Turkey data set was collected as part of the project “RedAlert – detection of pecking injuries in turkeys using neural networks” which was supported by the “Animal Welfare Innovation Award” of the “Initiative Tierwohl”.
	
	\end{ack}
	\bibliography{lib}

\newpage
\section*{Checklist}

\begin{enumerate}

\item For all authors...
\begin{enumerate}
  \item Do the main claims made in the abstract and introduction accurately reflect the paper's contributions and scope?
    \answerYes{All claims are either introduced in Section~\ref{sec:benchmark} or are concluded in Section~\ref{sec:discussion}}
  \item Did you describe the limitations of your work?
    \answerYes{See Section~\ref{subsec:limitations}}
  \item Did you discuss any potential negative societal impacts of your work?
    \answerYes{See Section~\ref{subsec:impact}}
  \item Have you read the ethics review guidelines and ensured that your paper conforms to them?
    \answerYes{All raw data was acquired from open data or in agreement with ethical guidelines.}
\end{enumerate}

\item If you are including theoretical results...
\begin{enumerate}
  \item Did you state the full set of assumptions of all theoretical results?
    \answerNA{No theoretical claims made}
	\item Did you include complete proofs of all theoretical results?
    \answerNA{No theoretical claims made}
\end{enumerate}

\item If you ran experiments (e.g. for benchmarks)...
\begin{enumerate}
  \item Did you include the code, data, and instructions needed to reproduce the main experimental results (either in the supplemental material or as a URL)?
    \answerYes{See Supplementary or the main repository.
    }
  \item Did you specify all the training details (e.g., data splits, hyperparameters, how they were chosen)?
    \answerYes{See Section~\ref{subsec:details}}
	\item Did you report error bars (e.g., with respect to the random seed after running experiments multiple times)?
    \answerYes{See Section~\ref{tbl:datasets}, As explained in \ref{sec:eval} most reported values are median values due to outliers, error bars would clutter the graphics and decrease their interpretability. We give all results (including SEM) in the supplementary for any interested reader.}
	\item Did you include the total amount of compute and the type of resources used (e.g., type of GPUs, internal cluster, or cloud provider)?
    \answerYes{See the implementation details in Section~\ref{subsec:details}}
\end{enumerate}

\item If you are using existing assets (e.g., code, data, models) or curating/releasing new assets...
\begin{enumerate}
  \item If your work uses existing assets, did you cite the creators?
    \answerYes{We credited all authors when we used or extended their datasets. See under datasets in Section~\ref{subsec:datasets}}
  \item Did you mention the license of the assets?
    \answerYes{All license information are in the license file in the Git Repository}
  \item Did you include any new assets either in the supplemental material or as a URL?
    \answerYes{Multiple datasets are created or extended by us, also the source code for the benchmark is new. The source code and datasets are reachable as described in the access section in the supplementary or in the main repository.}
  \item Did you discuss whether and how consent was obtained from people whose data you're using/curating?
    \answerYes{We either used own datasets, public datasets or received personal permission from respective data owners.}
  \item Did you discuss whether the data you are using/curating contains personally identifiable information or offensive content?
    \answerNA{The datasets are images of animals, objects or plants and thus can not identify people. Our data does not contain offensive content. }
\end{enumerate}

\item If you used crowdsourcing or conducted research with human subjects...
\begin{enumerate}
  \item Did you include the full text of instructions given to participants and screenshots, if applicable?
    \answerNA{We either used preliminary existing crowd sourcing materials or used only selected few workers as annotators and gave only oral instructions. We give credit and reference all previous instructions. }
  \item Did you describe any potential participant risks, with links to Institutional Review Board (IRB) approvals, if applicable?
    \answerNA{Not applicable, we see no potential risk for humans by annotating images.}
  \item Did you include the estimated hourly wage paid to participants and the total amount spent on participant compensation?
    \answerYes{We included hourly wages where applicable in the supplementary.}
\end{enumerate}

\end{enumerate}

\appendix

\newpage

\section{Supplementary}

\subsection{Access to datasets and benchmark}
\label{subsec:access}

The source code is available at \url{https://github.com/Emprime/dcic}.
The datasets are available at \url{https://doi.org/10.5281/zenodo.7152309}.

Please check the licenses which vary between datasets and source code.
You also need to credit all sources appropriately if you want to use them.

\FloatBarrier

\subsection{Additional Results}

\subsubsection{More annotations are better}

As stated in the main paper, the datasets Benthic, MiceBone, Pig, QualitMRI, and Treeveritsy\#6 have an $\hat{ACC}$ of lower than 85\% which marks them as difficult even for humans.
Thus, we call them \emph{difficult} and the rest \emph{easy} datasets in the following.

We include in \autoref{fig:1_more_annotations-0} all measured scores over the all calculated budgets. 
We see as in  \autoref{fig:trans_qual-0} an improvement of the scores with increasing budget.
For easier datasets, we see in \autoref{fig:1_more_annotations-1} that the improvements with additional annotations are lower or do not exist, supporting the observation that uncertain labels benefit more from additional annotations.
If we check in \autoref{fig:1_more_annotations-2} which methods benefit the most from additional annotations, we see up to 0.15 lower relative $KL$ and $ECE$ for soft labels. 
We can confirm similar observations on a t-SNE \cite{van2008tsne} visualization of the Plankton dataset in \autoref{fig:tsne}.
We use the method Pseudolabel v2 soft for these experiments.
For a budget of 10\%, the clusters are not as distinct as with 100\% while for a budget of 1000\% we can see  more linear transitions in alignment with the soft ground truth label.
In agreement with \cite{grossmann2022beyond}, we see that more annotated data helps to structure the feature space while linear transitions clarify with the added benefit of soft labels.

 \picThreePlus{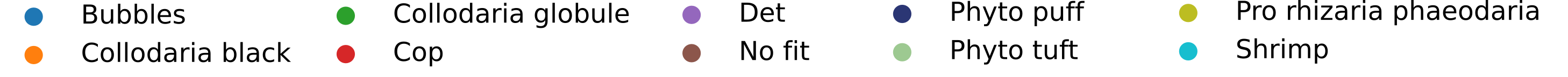}{10\%}{100\%}{1000\%}{T-SNE evaluation on Plankton dataset on a budget of 10\%, 100\% and 1000\% --
      On the top, all images are colored based on the majority class of the ground truth distribution.
      The lower images are zoomed in regions where we color only two classes accordingly to their actual ground truth distribution.
      }
      
\picThreePlus{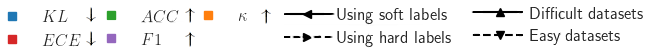}{Overview}{Difficult vs. Easy Datasets}{Soft vs. Hard Labels Used}{
Impact of increased budget across all methods and datasets -- The last two images show the relative changes in comparison to a budget of 100\% from different subgroups of the datasets and methods respectively.}

\subsubsection{Correlations}

Extending on the mentioned correlations in the main paper, we want to highlight to important conclusion from the investigated correlations.
$F1$ balances the precision and recall and is general not the same as $ACC$ but in our experiments we see almost identical values to $ACC$ which we credit to the averaging per class.
    A high correlation also exists between $ACC$ and $\kappa$ which means that higher consistency in the input labels for different folds leads to better classification scores.
    This is a particularly useful result since consistency can be calculated without training a network or knowing anything about the ground truth distribution.
    This discovery allows the opportunity to roughly estimate the $ACC$ of dataset without the knowledge of any labels.

\subsection{Further implementation details and insights}

\subsection{Details about $\hat{ACC}$}

In this section, we describe in detail why $\hat{ACC}$ measures the human performance on the baseline method. 
$\hat{ACC}$ measures the impact of the initialized labels against the ground truth for the baseline method. This is special for the baseline because it does not change the initialised labels. The ground truth is based on human annotations and the initialised labels on the ground truth and thus also human annotations.

The difference is the following: For example, if we have 20 annotations for an image $x$ and the resulting ground truth distribution for the image would be $l_x = [0.8, 0.05, 0.15]$ for three classes.. The initialized labels are sampled from $l_x$. This means for one annotation per image ($m = 1$) defined by an initialization scheme, the initialized label would be [1 0 0] in 80\% of the cases, [0 1 0] in 5\% of the cases and [0 0 1] in 15\% of the cases. In this case, the initialized label often represents the majority vote, but in about 20\% of the cases it would be a hard vote for a clear minority. For $\hat{ACC}$  we only look at the majority vote of the ground truth and thus the expected result would be class 0 for image $x$ even if we feed in 20\% of the cases a different training label for image $x$ in the training. This would of course negatively impact the classification performance as it would confuse the network.

An important detail is that we normally do not use the initialized label directly for the evaluation (it is not allowed to initialized labels on the test set). However, the baseline is designed to forward the sampled initialized labels in the test stage which effectively means that we sample one ($m$=1) or multiple human(s) ($m$>1) from the ground truth distribution. By doing so, the $\hat{ACC}$ measures the performance of the sampled annotation against the complete ground truth. 

In short, $\hat{ACC}$ measure the impact of sample human annotation against the complete ground truth which we interpret as the performance of a human annotator against the combined ground truth. 

\subsection{About the improvement of Semi- and Self-supervised methods}

In this section, we describe how Semi- and Self-supervised methods can be used to improve the labels in the first phase. 
When we talk about self-supervised methods, we mean the self-supervised training on a proxy task (e.g. maximzing agreement in SimCLR) and then a fine-tuning to the given classification task.

The improvement can be credited to two properties.
Firstly, in many experiments the labels are initialized only on a small portion of the dataset and thus predicting a label on the rest of the unlabeled data creates / improves them in contrast to the previously unknown state.
Secondly, when the network generalizes well, errors or ambiguities in the training data should be mitigated. 
For example, the network should predict a similar probability distribution for similar images even if the used training labels diverge.
As seen in \autoref{fig:trans_qual-0}, these two benefits lead to a lower $KL$ score in the second phase in comparison to the first phase $\hat{KL}$.

\subsubsection{Self-supervised methods}

We investigated the four self-supervised methods BYOL \cite{byol}, MOCOv2 \cite{mocov2}, SimCLR \cite{simclr} and SWAV \cite{swav}.
On large datasets like ImageNet \cite{imagenet} these methods have been shown to be comparable to or even surpass semi-supervised methods like FixMatch \cite{fixmatch}.
However, we assume all self-supervised methods to be inferior to most other benchmarked methods due to three main reasons.
Firstly, the self-supervised methods are more difficult to tune.
The models learn in the first phase without any supervision and thus depend on properly selected image augmentations \cite{byol}.
For better comparability, we enforced the same set of hyperparameters including image augmentations across all datasets.
This was problematic for some methods on specific datasets (e.g., BYOL on CIFAR-10H).
Secondly, self-supervised methods are typically trained on large datasets. 
We trained only on a few thousand images and restricted the batch size for a comparable hardware usage. 
These two restrictions are often not considered in the literature and therefore might negatively impact the performance.
Thirdly, many of the compared methods are pretrained on ImageNet. 
If we compare to not-pretrained methods such as \texttt{Pseudo v2 not}, we see that the self-supervised method is performing comparable or better.

\subsubsection{Hyperparameter optimization}

We used the public framework Ray Tune \footnote{https://docs.ray.io/en/latest/tune/index.html} for the hyperparameter optimization of the evaluation phase.
As stated in the main paper we determined them by applying Hyperopt \cite{bergstra2013hyperopt} with 100 search trials across the same grid of parameters for all datasets. 
The target was the minimization of $KL$ between $\Phi(x)$ and $l_x$ for the baseline experiment with exactly ten annotations per image across one fold.
Below we show the used parameter grid.

\begin{verbatim}
config={
    'weights': weights,
    'batch_size': tune.choice([8,32,128]),
    'epochs': tune.choice([20]),
    'lr': tune.choice([1e-1,3e-1,1e-4,1e-5]),
    'dropout': tune.choice([0, 0.5]),
    'network': tune.choice(['wideresnet28-10', 'resnet50v2_large',
      'densenet121', 'incepresv2', 'resnet50v2', 'wideresnet16-8' ]),
    'augmentation': tune.choice([0,1,2,3]),
    'opt': tune.choice(['sgd', 'adam', 'sgdw', 'sgdwr']),
    'input_upsampling': tune.choice([True, False]),
    'weight_decay': tune.choice([5e-4, 1e-3]),
},
\end{verbatim}

As initialization, we used a heuristically determined set of hyperparameters from our preliminary studies. 
The details about these parameters can be found in the source code.

\FloatBarrier

\subsection{Detailed dataset descriptions}

In this section, we describe in detail from where and how we collected the raw data, how we preprocessed the data and the labels, and give additional information per dataset. 
For all datasets, we split the data into five folds.
We ensured during the splitting that the class distribution and the total size are similar between the folds.
Moreover, we ensured that very similar images, e.g. consecutive slices in MiceBone are in the same fold.
In multiple cases, we simplified the class labels due to very long tail distributions or missing data.
In these cases, we give the mapping as \texttt{'original\_label' : 'our\_label'}. 
If our label is \texttt{None}, we ignored the class completely.

\subsubsection{Benthic}

The raw data was provided to us by the authors of \cite{schoening2020Megafauna,Langenkamper2020GearStudy}.
In their studies, multiple annotators labeled any objects they could detect on seafloor images. 
One example image without and with annotations is given in \autoref{fig:benthic_raw2}.

\picTwo{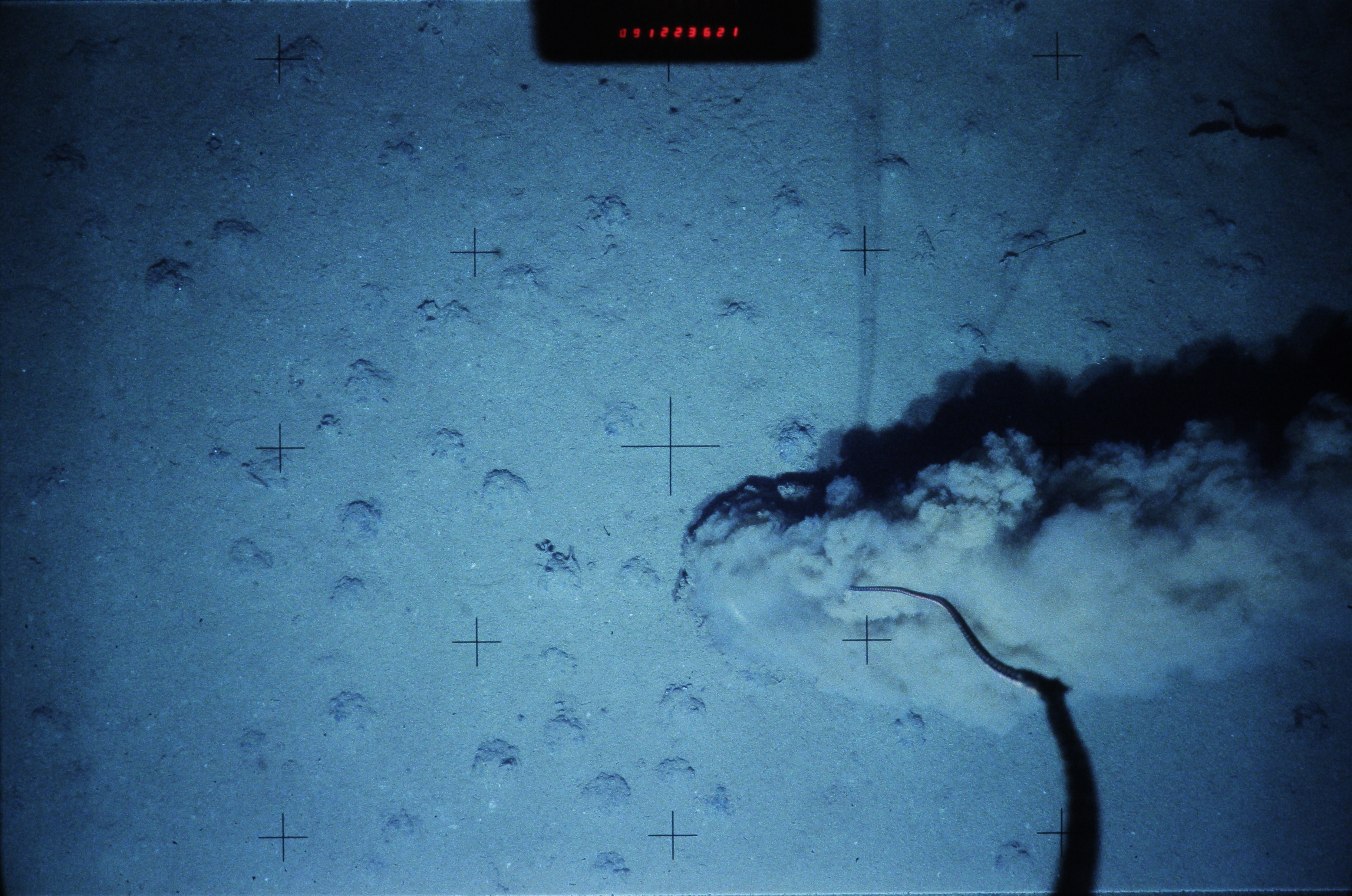}{Raw Input Image}{Including Annotations}{Raw example image without and with annotations. Every circle represents one annotation. }

We see that some items are only annotated once while others have multiple annotations.
We sorted all annotations in descending order of their size.
We merged all overlapping annotations with nearby centers. 
If at least three annotations were found, we used a cutout region around the largest found annotation as an image and the average across all annotations as the ground truth label.
The total number of annotations and the agreement with the majority vote per image can be seen as a histogram in \autoref{fig:benthic}. 
In agreement with the original authors, we simplified the original class with the following mapping:
\begin{verbatim}
"Anemone": "other_fauna",
"Coral": "coral",
"Crustacean": "crustacean",
"Epifauna": "other_fauna",
"Ipnops fish": "other_fauna",
"Jellyfish": "other_fauna",
"Litter": None,
"Mollusc": "other_fauna",
"Octopus / squid": None,
"Ophiuroid": "star",
"Other fauna": "other_fauna",
"Other fish": "other_fauna",
"Polychaete (sessile)": "worm",
"Sea Urchin": "other_fauna",
"Sea cucumber": "cucumber",
"Small encrusting": "encrusting",
"Spiral worm": "worm",
"Sponge": "sponge",
"Stalk (no head)": "other_fauna", 
"Stalked crinoid": "other_fauna",
"Stalked sponge": "other_fauna", 
"Starfish (not ophiuroid)": "star",
"polychaete (mobile)": "worm",
'lost+found': None,
\end{verbatim}

\picTwo{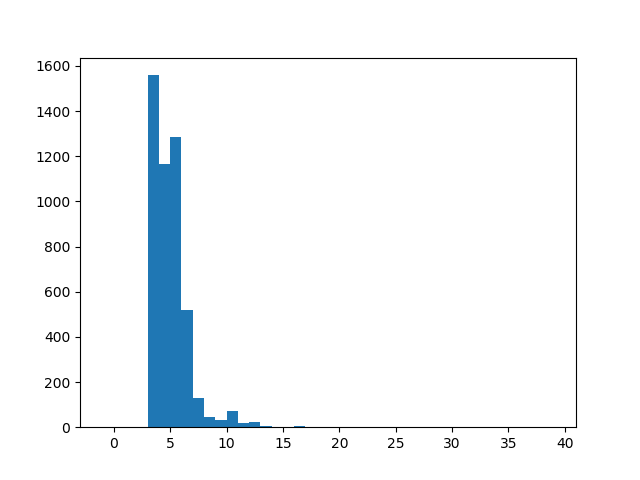}{Number of annotations}{Agreement}{Histograms for the Benthic Dataset}

\subsubsection{CIFAR-10H}

This dataset was originally described and created by \cite{cifar10h}.
The total number of annotations and the agreement with the majority vote per image can be seen as a histogram in \autoref{fig:cifar}. 

\picTwo{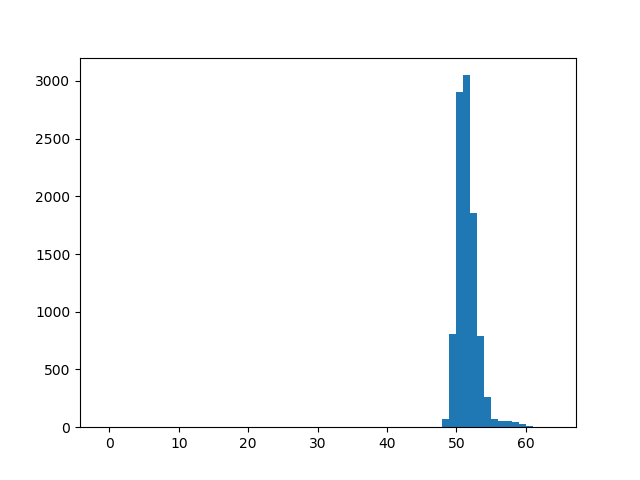}{Number of annotations}{Agreement}{Histograms for the CIFAR-10H Dataset}

\subsubsection{MiceBone}

We used the preprocessing described in \cite{dc3} and reproduced their description here with the authors permission:

The Mice Bone dataset is based on the raw data which was published in \cite{schmarje2019}.
The raw data are 3D scans from collagen fibers in mice bones.
The three proposed classes are similar ("g") and dissimilar ("ug") collagen fiber orientations and not relevant ("nr") regions due to noise or background.
We used the given segmentations to cut image regions from the original 2D image slices which mainly consist of one class.

We collected the annotations with four human workers.
Three of them were hired workers which replied to a local job advertisement and were paid 11€ per hour.
The last worker was a domain expert who desired no further payment than their contribution to science.
All workers received reference images as guidance from the domain expert and oral explanations of the different classes.
All reference images are shown in \autoref{fig:mice} with the written explanations.
The annotation process was structured in a partial first annotation of the data and multiple following annotations. 
The first partial annotation was used to get familiar with the annotation process and the classes.
No results from this first part were used in this publication.
We checked that all annotators achieved a similar classification performance than the domain expert with himself after this first part. 
All annotators annotated as much as possible during a fixed time frame of 15 hours.
We either annotated the full dataset or subset of 10\% multiple times.
The total number of annotations and the agreement with the majority vote per image can be seen as a histogram in \autoref{fig:mice1}. 

\picFourteenMice{mice}{Reference images for the annotation of the MiceBone Dataset}

\picTwo{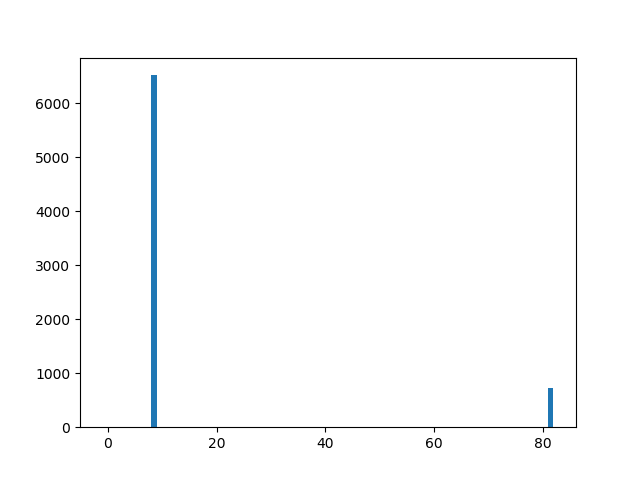}{Number of annotations}{Agreement}{Histograms for the MiceBone Dataset}

\subsection{Pig}

The raw data was collected at European farms for pigs. The animals were captured via a camera during their arrival and departure. 
We only received and worked with cropped regions of the tail provided to us by the University of Helsinki to protect data privacy and personal interest of the farms. 
We created a annotation catalog for the four classes, `no injury`, `shortened tail but healed`, `fresh wound at tail` and `tail not visible` as instruction. 
If annotators were uncertain they should decide themselves for the more probable class and in edge cases for the more severe class.
We tracked the consistency between the annotators and it stayed at around 50\% agreement over all annotations and annotators.
The data was annotated in multiple rounds and a subset of 20\% was annotated more often.
The 6 annotators were domain experts with one exception, a hired worker only for this task which was trained and instructed as the others. 
This hired worker was paid 11€ per hour and the domain expert did the work as part of their daily job.

\picTwo{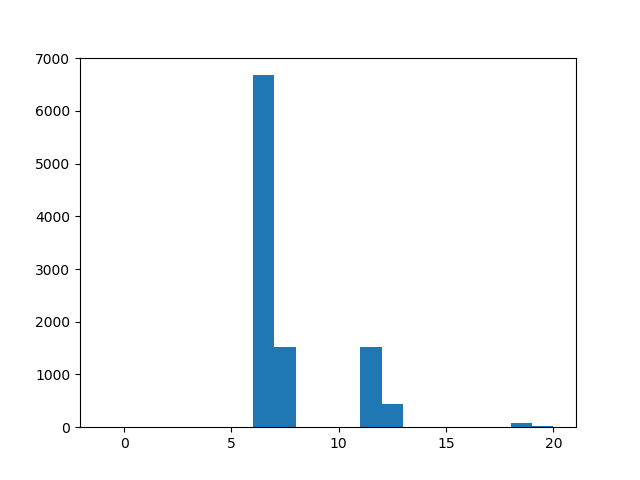}{Number of annotations}{Agreement}{Histograms for the Pig Dataset} 

\subsubsection{Plankton}

We used the preprocessing described in \cite{dc3} and reproduced their description here with the authors' permission.

The Plankton dataset was introduced in \cite{foc}. 
The dataset contains 10 plankton classes and has multiple labels per image due to the help of citizen scientists. 
In contrast to  \cite{foc}, we include ambiguous images in the training and validation set and do not enforce a class balance.

The mentioned citizen scientist used the platform PlanktonID \footnote{https://planktonid.geomar.de}.
The total number of annotations and the agreement with the majority vote per image can be seen as a histogram in \autoref{fig:plankton}. 

\picTwo{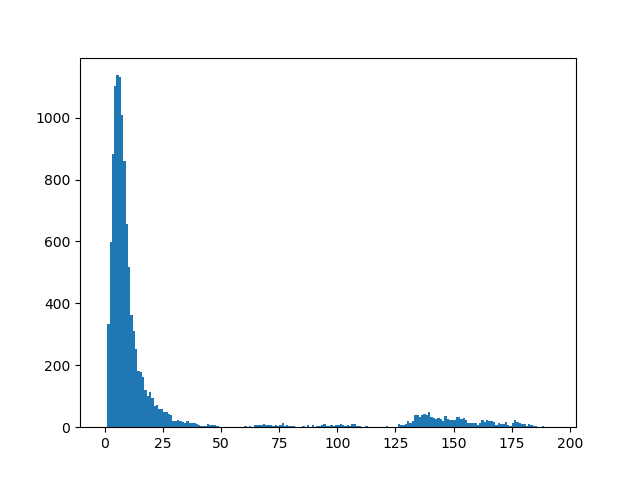}{Number of annotations}{Agreement}{Histograms for the Plankton Dataset}

\subsubsection{QualityMRI}

The QualityMRI dataset consists of human magnetic resonance images (MRI) with a varying quality and multiple subjective quality ratings  gathered in tests with radiologists.
	It was introduced and evaluated in \cite{obuchowicz2020qualityMRI,stepien2021cnnQuality}.
We received as raw data 70 images with around 31 annotations and 240 images with around 24 annotations.
The annotations had a score from 1 to 5. 
We simplified these labels into a continuous score between 0 and 1. 
We converted the given annotations into 4 new annotations for this continuous score. 
For example, we interpreted an original annotation of 1 as 4 annotations of 0 and an original annotation of 2 as 3 annotations 0 and one annotation 1.
The total number of annotations and the agreement with the majority vote per image can be seen as histogram in \autoref{fig:quality}. 

\picTwo{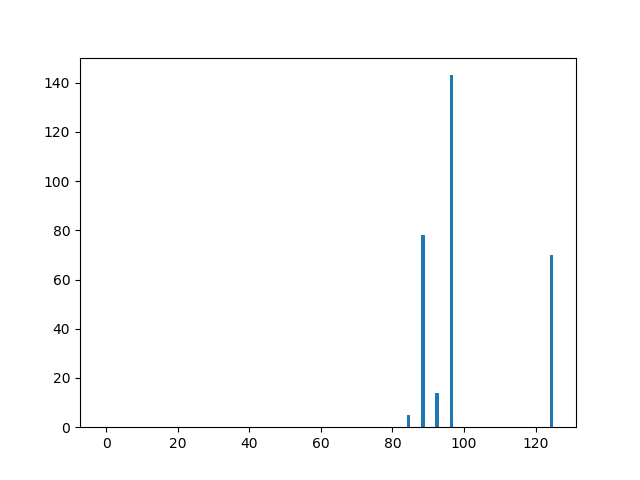}{Number of annotations}{Agreement}{Histograms for the QualityMRI Dataset}

\subsubsection{Synthetic}

This dataset consists of images that contain one blue, red, or green circle or ellipse on a black background.
We used a hue value of 0 (red), 120 (green) or 240 (blue) and the main axis ratio of 1 (circle) and 2 (ellipse).
The combination of all colors and ratios leads to six distinct classes. 
We generated 1200 images (200 per class) of all classes per fold.
The position and scale were determined randomly.
Additionally, we generated 1800 images with a random hue and main axis ratio.
Based on the random hue and ratio, we determined the class distribution of the six main classes.
For example, an image with the hue 60 and an axis ratio of 1.3 would have marginal probabilities of 50\% red, 50\% green, 30\% circle, and 70\% ellipse.
When we assume an independent combination, we arrive at the following ground truth distribution: red circle 0.15, red ellipse 0.35, blue circle 0.15, blue ellipse 0.35, green circle 0.0, green ellipse 0.0.
We generated 100 annotations based on these probability distributions for each image.

The total number of annotations and the agreement with the majority vote per image can be seen as a histogram in \autoref{fig:syn}. 

\picTwo{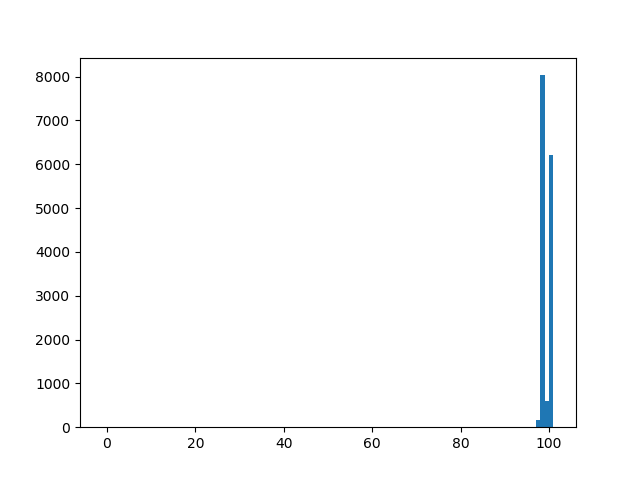}{Number of annotations}{Agreement}{Histograms for the Synthetic Dataset}

\subsubsection{Treeversity\#1 and Treeversity\#6}

Both datasets are based on a publicly available crowdsourced dataset of plant images from the Arnold Arboretum of Harvard University\footnote{\url{https://arboretum.harvard.edu/research/data-resources/}}.
The original task was to tag plant images with up 21 original classes.
The original crowd source page can be found at \url{https://www.zooniverse.org/projects/friedmaw/treeversity/about/research}.
We removed the original photographer and location at the bottom of each image to prevent any information leakage. 
We used all tags from the original research and interpreted them as annotations.
We simplified the original 22 tags to 6 classes as shown below.

\begin{verbatim}
    'FLWR': 'flower',
    'FLOWERINFLORESCENCE': 'flower',
    'LEAFNEEDLE': 'leaf',
    'LEAF': 'leaf', 'LF': 'leaf',
    'BARK': 'bark', 'BRK': 'bark',
    'BUDBREAK': 'bud',
    'BUD': 'bud', 'BD': 'bud',
    'FRUIT': 'fruit',
    'FRUITSEED': 'fruit', 'FRTSD': 'fruit',
    'WHOLEPLANT': 'whole_plant', 'WHLPLNT': 'whole_plant',
    'TRUNK': 'bark', 'TRNK': 'bark',
    'AUTUMNLEAFCOLOR': 'leaf', 'TMNLFCLR': 'leaf',
    'CONE': 'fruit', 'CN': 'fruit',
    'LEAFLESS': None, 'LFLSS': None', 
    'PLNTHLTHSS': None, 'PLANTHEALTHISSUE': None, 
    'BRNCH': 'bark', 'BRANCH': 'bark',
    'NSCTSPDRMT': None, 'INSECTSPIDERMITE': None,
     'PRICKLESPINETHORN': None, 'PRCKLSPNTHRN': None
    'LCHNMSSFNGS': None, 'LICHENMOSSFUNGUS': None
    'POORIMAGEQUALITYNOFEATURES': None
    'ROOT':'bark'
\end{verbatim}

We used a simplification due to overlapping classes because of the original tagging and small classes (e.g. health issues with 8 tags).
We counted multiple tags which we simplified to the same annotation class only once per image and user and used only images with at least three different annotations.
We saw that most people gave only one original tag per image (approx. 54\%) and the majority did not give more than three (approx. 98\%).
Due to the fact that tags are not the same as class annotations, we investigate two subsets.
In Treeversity\#1, we restrict the annotations based on original tags where exactly one tag was given.
In Treeversity\#6, we do not use this restriction.
The total number of annotations and the agreement with the majority vote per image can be seen as a histogram in \autoref{fig:tree1} and \autoref{fig:tree6}. 

\picTwo{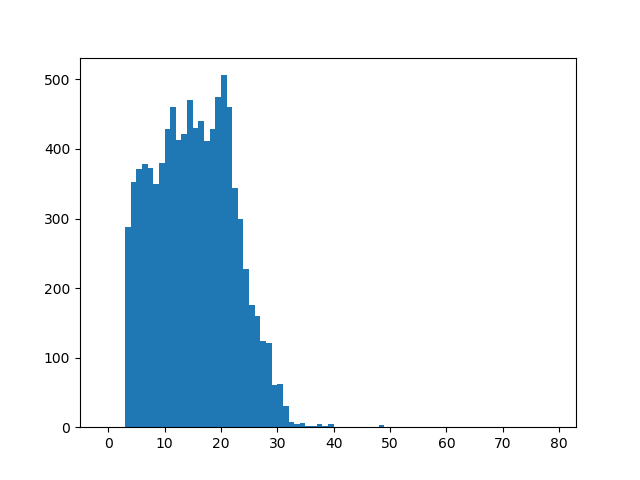}{Number of annotations}{Agreement}{Histograms for the TreeVersity\#1 Dataset}

\picTwo{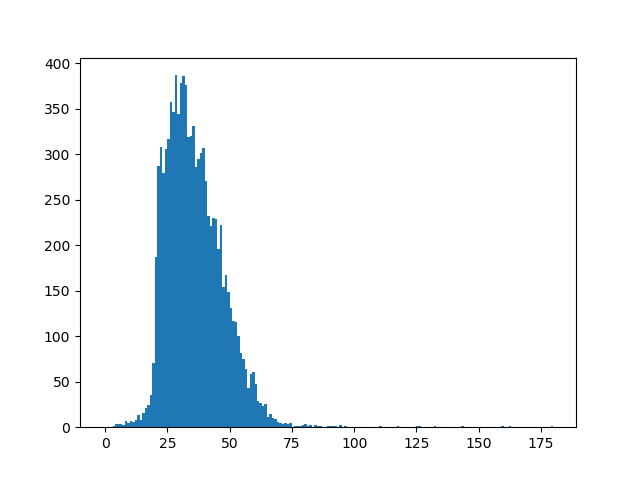}{Number of annotations}{Agreement}{Histograms for the TreeVersity\#6 Dataset}

\subsubsection{Turkey}

 The turkey dataset with images of turkeys and their injuries \cite{turkey_dataset,volkmann_turkeys}.
 We use three classes head injury, plumage injury, or no injury. 
 All injuries from the neck upwards to the eyes were counted as head injuries and the rest as plumage injuries.
 We used the original cross-analysis results between domain experts of the original papers.
 In addition, we collected additional annotations with a similar protocol as for the MiceBone dataset.
 The main difference is that reference images were not given but the distinction were discussed with all annotators and the domain expert multiple times.
 The total number of annotations and the agreement with the majority vote per image can be seen as a histogram in \autoref{fig:turkey}. 

\picTwo{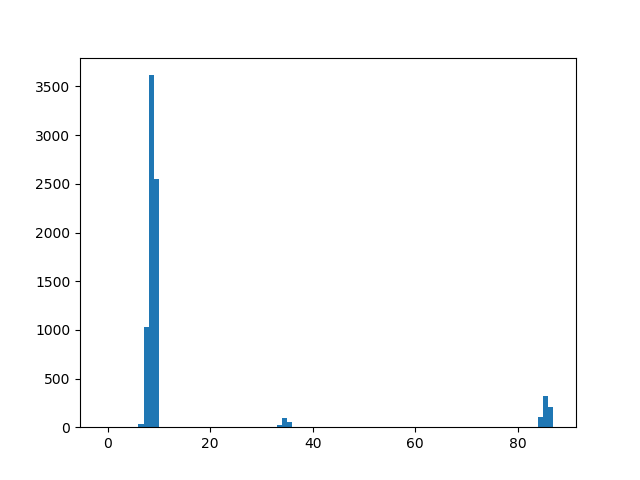}{Number of annotations}{Agreement}{Histograms for the Turkey Dataset}

\FloatBarrier

\subsection{Full result tables}

We give the full result tables including Median and Mean $\pm$ SEM  for all figures in the main paper. 
Additionally, we create an overview for $ACC$ and $KL$ for a budget of 10\%, 100\%, and 1000\%. 

\tblMoreAnnotationsZero
\tblMoreAnnotationsOne
\tblMoreAnnotationsTwo
\tblTransQualZero
\tblTransQualOne
\tblTransQualTwo

\tblImprovementKl
\tblImprovementAcc
\tblDetailsTenKl
\tblDetailsTenAcc
\tblDetailsHundretKl
\tblDetailsHundretAcc
\tblDetailsThousandKl
\tblDetailsThousandAcc

\FloatBarrier

\subsection{Raw Data}

In the following, we provide the raw data aggregated across the three slices as mean $\pm$ standard deviation (STD).
The content is best viewed digitally.
Some results are missing (N/A) due to hardware restrictions, degenerated training, or not applicable methods (e.g., the method Het does not work with only two classes as in QualityMRI).

\begin{table}[t]
\resizebox{\textwidth}{!}{
\begin{tabular}{l c c c c c c c c c c c c c}
Method                    & $m$ & $n$  &  $F1$                &  $ACC$               &  $KL$                &  $ECE$               &  $\hat{F1}$          &  $\hat{ACC}$         &  $\hat{KL}$          &  $\hat{ECE}$         &  $\kappa$            &  $b$                 &  $\hat{b}$          \\ 
 \midrule \\ 
BYOL                      & 01 & 0.10  & 0.5729 $\pm$ 0.0074 & 0.5561 $\pm$ 0.0061 & 0.9457 $\pm$ 0.0116 & 0.0286 $\pm$ 0.0057 & 0.5820 $\pm$ 0.0054 & 0.5654 $\pm$ 0.0036 & 1.0828 $\pm$ 0.0265 & 0.0997 $\pm$ 0.0128 & 0.5186 $\pm$ 0.0118 & 0.0999 $\pm$ 0.0001 & 0.0999 $\pm$ 0.0001\\ 
BYOL                      & 01 & 0.20  & 0.5742 $\pm$ 0.0145 & 0.5499 $\pm$ 0.0246 & 0.9061 $\pm$ 0.0395 & 0.0389 $\pm$ 0.0070 & 0.5865 $\pm$ 0.0291 & 0.5734 $\pm$ 0.0281 & 0.9676 $\pm$ 0.0476 & 0.0628 $\pm$ 0.0176 & 0.5730 $\pm$ 0.0078 & 0.1999 $\pm$ 0.0001 & 0.1999 $\pm$ 0.0001\\ 
BYOL                      & 01 & 0.50  & 0.6216 $\pm$ 0.0328 & 0.6033 $\pm$ 0.0349 & 0.8066 $\pm$ 0.0586 & 0.0672 $\pm$ 0.0169 & 0.6345 $\pm$ 0.0114 & 0.6253 $\pm$ 0.0166 & 0.8324 $\pm$ 0.0366 & 0.0317 $\pm$ 0.0102 & 0.6597 $\pm$ 0.0172 & 0.4999 $\pm$ 0.0001 & 0.4999 $\pm$ 0.0001\\ 
BYOL                      & 01 & 1.00  & 0.6615 $\pm$ 0.0098 & 0.6426 $\pm$ 0.0065 & 0.7352 $\pm$ 0.0281 & 0.0899 $\pm$ 0.0057 & 0.6563 $\pm$ 0.0089 & 0.6453 $\pm$ 0.0050 & 0.7630 $\pm$ 0.0169 & 0.0311 $\pm$ 0.0055 & 0.7192 $\pm$ 0.0174 & 1.0000 $\pm$ 0.0000 & 1.0000 $\pm$ 0.0000\\ 
DivideMix                 & 01 & 0.10  & 0.5597 $\pm$ 0.0105 & 0.5632 $\pm$ 0.0051 & 1.0669 $\pm$ 0.0452 & 0.1192 $\pm$ 0.0269 & 0.5811 $\pm$ 0.0129 & 0.5795 $\pm$ 0.0151 & 0.4633 $\pm$ 0.0295 & 0.1590 $\pm$ 0.0078 & 0.4951 $\pm$ 0.0155 & 0.0999 $\pm$ 0.0001 & 0.0999 $\pm$ 0.0001\\
\end{tabular}
}
\caption{Results for dataset Benthic}
\end{table}

\begin{table}[t]
\resizebox{\textwidth}{!}{
\begin{tabular}{l c c c c c c c c c c c c c}
Method                    & $m$ & $n$  &  $F1$                &  $ACC$               &  $KL$                &  $ECE$               &  $\hat{F1}$          &  $\hat{ACC}$         &  $\hat{KL}$          &  $\hat{ECE}$         &  $\kappa$            &  $b$                 &  $\hat{b}$          \\ 
 \midrule \\ 
BYOL                      & 01 & 0.10  & 0.1814 $\pm$ 0.0013 & 0.2095 $\pm$ 0.0086 & 1.9838 $\pm$ 0.0193 & 0.0304 $\pm$ 0.0025 & 0.1407 $\pm$ 0.0093 & 0.1849 $\pm$ 0.0012 & 2.0503 $\pm$ 0.0051 & 0.0392 $\pm$ 0.0037 & 0.4263 $\pm$ 0.1113 & 0.1000 $\pm$ 0.0000 & 0.1000 $\pm$ 0.0000\\ 
BYOL                      & 01 & 0.20  & 0.2465 $\pm$ 0.0726 & 0.2658 $\pm$ 0.0657 & 1.8161 $\pm$ 0.1234 & 0.0247 $\pm$ 0.0128 & 0.2005 $\pm$ 0.0501 & 0.2215 $\pm$ 0.0312 & 2.0070 $\pm$ 0.0523 & 0.0767 $\pm$ 0.0586 & 0.1160 $\pm$ 0.0202 & 0.2000 $\pm$ 0.0000 & 0.2000 $\pm$ 0.0000\\ 
BYOL                      & 01 & 0.50  & 0.3757 $\pm$ 0.0828 & 0.3895 $\pm$ 0.0770 & 1.5548 $\pm$ 0.2339 & 0.0630 $\pm$ 0.0224 & 0.2830 $\pm$ 0.0691 & 0.2933 $\pm$ 0.0650 & 1.8778 $\pm$ 0.1154 & 0.1051 $\pm$ 0.0380 & 0.2059 $\pm$ 0.0141 & 0.5000 $\pm$ 0.0000 & 0.5000 $\pm$ 0.0000\\ 
BYOL                      & 01 & 1.00  & 0.3708 $\pm$ 0.0562 & 0.3848 $\pm$ 0.0590 & 1.6074 $\pm$ 0.2167 & 0.0825 $\pm$ 0.0360 & 0.2730 $\pm$ 0.0526 & 0.2790 $\pm$ 0.0496 & 1.9395 $\pm$ 0.0343 & 0.0959 $\pm$ 0.0651 & 0.2056 $\pm$ 0.0387 & 1.0000 $\pm$ 0.0000 & 1.0000 $\pm$ 0.0000\\ 
DivideMix                 & 01 & 0.10  & 0.6608 $\pm$ 0.0073 & 0.6903 $\pm$ 0.0029 & 0.9671 $\pm$ 0.0359 & 0.1806 $\pm$ 0.0192 & 0.6264 $\pm$ 0.0078 & 0.6577 $\pm$ 0.0028 & 0.4765 $\pm$ 0.0264 & 0.1310 $\pm$ 0.0035 & 0.6540 $\pm$ 0.0205 & 0.1000 $\pm$ 0.0000 & 0.1000 $\pm$ 0.0000\\
\end{tabular}
}
\caption{Results for dataset CIFAR10H}
\end{table}

\begin{table}[t]
\resizebox{\textwidth}{!}{
\begin{tabular}{l c c c c c c c c c c c c c}
Method                    & $m$ & $n$  &  $F1$                &  $ACC$               &  $KL$                &  $ECE$               &  $\hat{F1}$          &  $\hat{ACC}$         &  $\hat{KL}$          &  $\hat{ECE}$         &  $\kappa$            &  $b$                 &  $\hat{b}$          \\ 
 \midrule \\ 
BYOL                      & 01 & 0.10  & 0.5097 $\pm$ 0.0345 & 0.5600 $\pm$ 0.0477 & 0.5300 $\pm$ 0.0460 & 0.1084 $\pm$ 0.0178 & 0.5154 $\pm$ 0.0629 & 0.5684 $\pm$ 0.0662 & 1.0798 $\pm$ 0.2157 & 0.2017 $\pm$ 0.0318 & 0.4100 $\pm$ 0.0375 & 0.0999 $\pm$ 0.0000 & 0.0999 $\pm$ 0.0000\\ 
BYOL                      & 01 & 0.20  & 0.5085 $\pm$ 0.0203 & 0.5536 $\pm$ 0.0260 & 0.5435 $\pm$ 0.0788 & 0.1300 $\pm$ 0.0269 & 0.5176 $\pm$ 0.0325 & 0.5689 $\pm$ 0.0197 & 1.0121 $\pm$ 0.0441 & 0.2048 $\pm$ 0.0276 & 0.4847 $\pm$ 0.0596 & 0.2000 $\pm$ 0.0000 & 0.2000 $\pm$ 0.0000\\ 
BYOL                      & 01 & 0.50  & 0.5871 $\pm$ 0.0271 & 0.6071 $\pm$ 0.0219 & 0.4270 $\pm$ 0.0290 & 0.0592 $\pm$ 0.0213 & 0.5418 $\pm$ 0.0218 & 0.5905 $\pm$ 0.0105 & 0.8816 $\pm$ 0.0470 & 0.1613 $\pm$ 0.0232 & 0.5125 $\pm$ 0.0473 & 0.4999 $\pm$ 0.0000 & 0.4999 $\pm$ 0.0000\\ 
BYOL                      & 01 & 1.00  & 0.5422 $\pm$ 0.0175 & 0.5444 $\pm$ 0.0256 & 0.4568 $\pm$ 0.0241 & 0.0802 $\pm$ 0.0179 & 0.5284 $\pm$ 0.0491 & 0.5647 $\pm$ 0.0259 & 1.0331 $\pm$ 0.0772 & 0.1688 $\pm$ 0.0112 & 0.5352 $\pm$ 0.0664 & 1.0000 $\pm$ 0.0000 & 1.0000 $\pm$ 0.0000\\ 
DivideMix                 & 01 & 0.10  & 0.4653 $\pm$ 0.0956 & 0.4871 $\pm$ 0.0991 & 0.4514 $\pm$ 0.0687 & 0.1050 $\pm$ 0.0681 & 0.5195 $\pm$ 0.0671 & 0.5389 $\pm$ 0.0752 & -0.0336 $\pm$ 0.0859 & 0.0636 $\pm$ 0.0416 & 0.3707 $\pm$ 0.1178 & 0.0999 $\pm$ 0.0000 & 0.0999 $\pm$ 0.0000\\
\end{tabular}
}
\caption{Results for dataset MiceBone}
\end{table}

\begin{table}[t]
\resizebox{\textwidth}{!}{
\begin{tabular}{l c c c c c c c c c c c c c}
Method                    & $m$ & $n$  &  $F1$                &  $ACC$               &  $KL$                &  $ECE$               &  $\hat{F1}$          &  $\hat{ACC}$         &  $\hat{KL}$          &  $\hat{ECE}$         &  $\kappa$            &  $b$                 &  $\hat{b}$          \\ 
 \midrule \\ 
BYOL                      & 01 & 0.10  & 0.2875 $\pm$ 0.0448 & 0.2961 $\pm$ 0.0328 & 0.7423 $\pm$ 0.0184 & 0.1621 $\pm$ 0.0108 & 0.2986 $\pm$ 0.0394 & 0.3026 $\pm$ 0.0356 & 1.2326 $\pm$ 0.1833 & 0.3006 $\pm$ 0.0290 & 0.0915 $\pm$ 0.0165 & 0.1000 $\pm$ 0.0000 & 0.1000 $\pm$ 0.0000\\ 
BYOL                      & 01 & 0.20  & 0.3160 $\pm$ 0.0007 & 0.3382 $\pm$ 0.0069 & 0.7102 $\pm$ 0.0229 & 0.1129 $\pm$ 0.0180 & 0.3333 $\pm$ 0.0239 & 0.3389 $\pm$ 0.0195 & 1.0877 $\pm$ 0.0458 & 0.2621 $\pm$ 0.0172 & 0.2018 $\pm$ 0.0347 & 0.2000 $\pm$ 0.0000 & 0.2000 $\pm$ 0.0000\\ 
BYOL                      & 01 & 0.50  & 0.3562 $\pm$ 0.0172 & 0.3665 $\pm$ 0.0111 & 0.6249 $\pm$ 0.0394 & 0.0615 $\pm$ 0.0343 & 0.3357 $\pm$ 0.0232 & 0.3407 $\pm$ 0.0211 & 1.0636 $\pm$ 0.1034 & 0.2562 $\pm$ 0.0147 & 0.1995 $\pm$ 0.0383 & 0.5000 $\pm$ 0.0000 & 0.5000 $\pm$ 0.0000\\ 
BYOL                      & 01 & 1.00  & 0.3300 $\pm$ 0.0221 & 0.3413 $\pm$ 0.0234 & 0.6425 $\pm$ 0.0144 & 0.0866 $\pm$ 0.0264 & 0.3437 $\pm$ 0.0120 & 0.3508 $\pm$ 0.0122 & 1.1124 $\pm$ 0.1066 & 0.2607 $\pm$ 0.0255 & 0.2149 $\pm$ 0.0189 & 1.0000 $\pm$ 0.0000 & 1.0000 $\pm$ 0.0000\\ 
DivideMix                 & 01 & 0.10  & 0.3570 $\pm$ 0.0284 & 0.3684 $\pm$ 0.0158 & 0.7348 $\pm$ 0.0669 & 0.1724 $\pm$ 0.0408 & 0.3946 $\pm$ 0.0442 & 0.4096 $\pm$ 0.0323 & 1.4538 $\pm$ 0.1111 & 0.3237 $\pm$ 0.0449 & 0.2433 $\pm$ 0.0123 & 0.1000 $\pm$ 0.0000 & 0.1000 $\pm$ 0.0000\\
\end{tabular}
}
\caption{Results for dataset Pig}
\end{table}

\begin{table}[t]
\resizebox{\textwidth}{!}{
\begin{tabular}{l c c c c c c c c c c c c c}
Method                    & $m$ & $n$  &  $F1$                &  $ACC$               &  $KL$                &  $ECE$               &  $\hat{F1}$          &  $\hat{ACC}$         &  $\hat{KL}$          &  $\hat{ECE}$         &  $\kappa$            &  $b$                 &  $\hat{b}$          \\ 
 \midrule \\ 
BYOL                      & 01 & 0.10  & 0.6635 $\pm$ 0.0649 & 0.6697 $\pm$ 0.0671 & 0.7899 $\pm$ 0.1436 & 0.0330 $\pm$ 0.0022 & 0.6072 $\pm$ 0.0638 & 0.6012 $\pm$ 0.0662 & 1.0759 $\pm$ 0.1352 & 0.1051 $\pm$ 0.0221 & 0.5512 $\pm$ 0.0128 & 0.1000 $\pm$ 0.0000 & 0.1000 $\pm$ 0.0000\\ 
BYOL                      & 01 & 0.20  & 0.6935 $\pm$ 0.0493 & 0.7035 $\pm$ 0.0509 & 0.7056 $\pm$ 0.1014 & 0.0215 $\pm$ 0.0027 & 0.6482 $\pm$ 0.0552 & 0.6477 $\pm$ 0.0588 & 0.9227 $\pm$ 0.1221 & 0.0729 $\pm$ 0.0201 & 0.6446 $\pm$ 0.0242 & 0.2000 $\pm$ 0.0000 & 0.2000 $\pm$ 0.0000\\ 
BYOL                      & 01 & 0.50  & 0.7254 $\pm$ 0.0511 & 0.7371 $\pm$ 0.0519 & 0.6367 $\pm$ 0.1294 & 0.0376 $\pm$ 0.0168 & 0.6605 $\pm$ 0.0574 & 0.6557 $\pm$ 0.0581 & 0.8635 $\pm$ 0.1673 & 0.0585 $\pm$ 0.0229 & 0.6299 $\pm$ 0.0250 & 0.5000 $\pm$ 0.0000 & 0.5000 $\pm$ 0.0000\\ 
BYOL                      & 01 & 1.00  & 0.7869 $\pm$ 0.0224 & 0.8002 $\pm$ 0.0215 & 0.5106 $\pm$ 0.0527 & 0.0241 $\pm$ 0.0044 & 0.7165 $\pm$ 0.0301 & 0.7165 $\pm$ 0.0299 & 0.7778 $\pm$ 0.0821 & 0.0772 $\pm$ 0.0050 & 0.7191 $\pm$ 0.0149 & 1.0000 $\pm$ 0.0000 & 1.0000 $\pm$ 0.0000\\ 
DivideMix                 & 01 & 0.10  & 0.7192 $\pm$ 0.0098 & 0.7653 $\pm$ 0.0136 & 0.6350 $\pm$ 0.0417 & 0.1127 $\pm$ 0.0155 & 0.7173 $\pm$ 0.0141 & 0.7579 $\pm$ 0.0173 & 0.3050 $\pm$ 0.0272 & 0.0609 $\pm$ 0.0099 & 0.8223 $\pm$ 0.0075 & 0.1000 $\pm$ 0.0000 & 0.1000 $\pm$ 0.0000\\
\end{tabular}
}
\caption{Results for dataset Plankton}
\end{table}

\begin{table}[t]
\resizebox{\textwidth}{!}{
\begin{tabular}{l c c c c c c c c c c c c c}
Method                    & $m$ & $n$  &  $F1$                &  $ACC$               &  $KL$                &  $ECE$               &  $\hat{F1}$          &  $\hat{ACC}$         &  $\hat{KL}$          &  $\hat{ECE}$         &  $\kappa$            &  $b$                 &  $\hat{b}$          \\ 
 \midrule \\ 
BYOL                      & 01 & 0.10  & 0.3876 $\pm$ 0.0252 & 0.4428 $\pm$ 0.0640 & 1.7667 $\pm$ 0.9465 & 0.3935 $\pm$ 0.1115 & 0.4123 $\pm$ 0.0444 & 0.4744 $\pm$ 0.0535 & 3.1910 $\pm$ 2.1229 & 0.3828 $\pm$ 0.1778 & -0.0210 $\pm$ 0.0297 & 0.0968 $\pm$ 0.0000 & 0.0968 $\pm$ 0.0000\\ 
BYOL                      & 01 & 0.20  & 0.4082 $\pm$ 0.0271 & 0.4755 $\pm$ 0.0172 & 2.2554 $\pm$ 0.8723 & 0.3587 $\pm$ 0.0396 & 0.4666 $\pm$ 0.0640 & 0.4829 $\pm$ 0.0532 & 4.7936 $\pm$ 0.3150 & 0.4619 $\pm$ 0.0756 & -0.1304 $\pm$ 0.0897 & 0.1976 $\pm$ 0.0000 & 0.1976 $\pm$ 0.0000\\ 
BYOL                      & 01 & 0.50  & 0.5523 $\pm$ 0.0681 & 0.5769 $\pm$ 0.0462 & 1.7833 $\pm$ 0.3197 & 0.2832 $\pm$ 0.0334 & 0.5125 $\pm$ 0.0567 & 0.5138 $\pm$ 0.0549 & 3.7317 $\pm$ 0.6659 & 0.4149 $\pm$ 0.0621 & -0.0159 $\pm$ 0.0216 & 0.5000 $\pm$ 0.0000 & 0.5000 $\pm$ 0.0000\\ 
BYOL                      & 01 & 1.00  & 0.4576 $\pm$ 0.0506 & 0.5045 $\pm$ 0.0188 & 1.0714 $\pm$ 0.1787 & 0.3317 $\pm$ 0.0358 & 0.5303 $\pm$ 0.0391 & 0.5308 $\pm$ 0.0384 & 3.4709 $\pm$ 0.4725 & 0.4088 $\pm$ 0.0486 & 0.1854 $\pm$ 0.0563 & 1.0000 $\pm$ 0.0000 & 1.0000 $\pm$ 0.0000\\ 
DivideMix                 & 01 & 0.10  & 0.5114 $\pm$ 0.1521 & 0.5618 $\pm$ 0.1065 & 0.2375 $\pm$ 0.0925 & 0.1301 $\pm$ 0.0352 & 0.5039 $\pm$ 0.0755 & 0.5504 $\pm$ 0.0361 & 0.1516 $\pm$ 0.4777 & 0.1659 $\pm$ 0.1432 & -0.0319 $\pm$ 0.0270 & 0.0968 $\pm$ 0.0000 & 0.0968 $\pm$ 0.0000\\
\end{tabular}
}
\caption{Results for dataset QualityMRI}
\end{table}

\begin{table}[t]
\resizebox{\textwidth}{!}{
\begin{tabular}{l c c c c c c c c c c c c c}
Method                    & $m$ & $n$  &  $F1$                &  $ACC$               &  $KL$                &  $ECE$               &  $\hat{F1}$          &  $\hat{ACC}$         &  $\hat{KL}$          &  $\hat{ECE}$         &  $\kappa$            &  $b$                 &  $\hat{b}$          \\ 
 \midrule \\ 
BYOL                      & 01 & 0.10  & 0.6060 $\pm$ 0.0060 & 0.6131 $\pm$ 0.0126 & 0.4437 $\pm$ 0.0154 & 0.0353 $\pm$ 0.0063 & 0.5715 $\pm$ 0.0230 & 0.5758 $\pm$ 0.0215 & 0.8207 $\pm$ 0.1118 & 0.1385 $\pm$ 0.0243 & 0.5073 $\pm$ 0.0134 & 0.1000 $\pm$ 0.0000 & 0.1000 $\pm$ 0.0000\\ 
BYOL                      & 01 & 0.20  & 0.6389 $\pm$ 0.0235 & 0.6452 $\pm$ 0.0224 & 0.3869 $\pm$ 0.0056 & 0.0391 $\pm$ 0.0176 & 0.5965 $\pm$ 0.0097 & 0.6010 $\pm$ 0.0074 & 0.6531 $\pm$ 0.0359 & 0.1052 $\pm$ 0.0061 & 0.6234 $\pm$ 0.0096 & 0.2000 $\pm$ 0.0000 & 0.2000 $\pm$ 0.0000\\ 
BYOL                      & 01 & 0.50  & 0.6513 $\pm$ 0.0161 & 0.6552 $\pm$ 0.0140 & 0.3823 $\pm$ 0.0123 & 0.0487 $\pm$ 0.0076 & 0.6037 $\pm$ 0.0086 & 0.6099 $\pm$ 0.0063 & 0.6672 $\pm$ 0.1158 & 0.0996 $\pm$ 0.0289 & 0.5928 $\pm$ 0.0237 & 0.5000 $\pm$ 0.0000 & 0.5000 $\pm$ 0.0000\\ 
BYOL                      & 01 & 1.00  & 0.6733 $\pm$ 0.0013 & 0.6773 $\pm$ 0.0024 & 0.3603 $\pm$ 0.0055 & 0.0581 $\pm$ 0.0139 & 0.6272 $\pm$ 0.0169 & 0.6305 $\pm$ 0.0180 & 0.6658 $\pm$ 0.0338 & 0.0959 $\pm$ 0.0133 & 0.6179 $\pm$ 0.0204 & 1.0000 $\pm$ 0.0000 & 1.0000 $\pm$ 0.0000\\ 
DivideMix                 & 01 & 0.10  & 0.8726 $\pm$ 0.0155 & 0.8717 $\pm$ 0.0152 & 0.2606 $\pm$ 0.0507 & 0.0537 $\pm$ 0.0257 & 0.8845 $\pm$ 0.0065 & 0.8840 $\pm$ 0.0063 & 0.1202 $\pm$ 0.0674 & 0.0154 $\pm$ 0.0039 & 0.8362 $\pm$ 0.0206 & 0.1000 $\pm$ 0.0000 & 0.1000 $\pm$ 0.0000\\
\end{tabular}
}
\caption{Results for dataset Synthetic}
\end{table}

\begin{table}[t]
\resizebox{\textwidth}{!}{
\begin{tabular}{l c c c c c c c c c c c c c}
Method                    & $m$ & $n$  &  $F1$                &  $ACC$               &  $KL$                &  $ECE$               &  $\hat{F1}$          &  $\hat{ACC}$         &  $\hat{KL}$          &  $\hat{ECE}$         &  $\kappa$            &  $b$                 &  $\hat{b}$          \\ 
 \midrule \\ 
BYOL                      & 01 & 0.10  & 0.5553 $\pm$ 0.0181 & 0.5709 $\pm$ 0.0138 & 0.9523 $\pm$ 0.0454 & 0.0847 $\pm$ 0.0097 & 0.5240 $\pm$ 0.0081 & 0.5305 $\pm$ 0.0103 & 1.4802 $\pm$ 0.0827 & 0.2180 $\pm$ 0.0111 & 0.4813 $\pm$ 0.0086 & 0.1000 $\pm$ 0.0000 & 0.1000 $\pm$ 0.0000\\ 
BYOL                      & 01 & 0.20  & 0.6068 $\pm$ 0.0352 & 0.6183 $\pm$ 0.0355 & 0.7976 $\pm$ 0.0417 & 0.0432 $\pm$ 0.0161 & 0.5536 $\pm$ 0.0244 & 0.5579 $\pm$ 0.0241 & 1.2286 $\pm$ 0.0252 & 0.1867 $\pm$ 0.0023 & 0.5127 $\pm$ 0.0282 & 0.1999 $\pm$ 0.0000 & 0.1999 $\pm$ 0.0000\\ 
BYOL                      & 01 & 0.50  & 0.6216 $\pm$ 0.0191 & 0.6265 $\pm$ 0.0166 & 0.7486 $\pm$ 0.0496 & 0.0315 $\pm$ 0.0054 & 0.5745 $\pm$ 0.0157 & 0.5782 $\pm$ 0.0154 & 1.1168 $\pm$ 0.0447 & 0.1584 $\pm$ 0.0131 & 0.5707 $\pm$ 0.0158 & 0.5000 $\pm$ 0.0000 & 0.5000 $\pm$ 0.0000\\ 
BYOL                      & 01 & 1.00  & 0.5882 $\pm$ 0.0451 & 0.5924 $\pm$ 0.0336 & 0.8186 $\pm$ 0.0909 & 0.0444 $\pm$ 0.0181 & 0.5320 $\pm$ 0.0281 & 0.5323 $\pm$ 0.0298 & 1.2814 $\pm$ 0.1346 & 0.1909 $\pm$ 0.0282 & 0.5197 $\pm$ 0.0412 & 1.0000 $\pm$ 0.0000 & 1.0000 $\pm$ 0.0000\\ 
DivideMix                 & 01 & 0.10  & 0.7348 $\pm$ 0.0077 & 0.7431 $\pm$ 0.0057 & 0.5952 $\pm$ 0.0267 & 0.0844 $\pm$ 0.0073 & 0.7399 $\pm$ 0.0080 & 0.7543 $\pm$ 0.0061 & 0.1689 $\pm$ 0.0172 & 0.0758 $\pm$ 0.0009 & 0.7755 $\pm$ 0.0067 & 0.1000 $\pm$ 0.0000 & 0.1000 $\pm$ 0.0000\\
\end{tabular}
}
\caption{Results for dataset Treeversity\#1}
\end{table}

\begin{table}[t]
\resizebox{\textwidth}{!}{
\begin{tabular}{l c c c c c c c c c c c c c}
Method                    & $m$ & $n$  &  $F1$                &  $ACC$               &  $KL$                &  $ECE$               &  $\hat{F1}$          &  $\hat{ACC}$         &  $\hat{KL}$          &  $\hat{ECE}$         &  $\kappa$            &  $b$                 &  $\hat{b}$          \\ 
 \midrule \\ 
BYOL                      & 01 & 0.10  & 0.4777 $\pm$ 0.0346 & 0.4945 $\pm$ 0.0396 & 0.7195 $\pm$ 0.0279 & 0.0659 $\pm$ 0.0304 & 0.4558 $\pm$ 0.0256 & 0.4685 $\pm$ 0.0283 & 1.1810 $\pm$ 0.0997 & 0.2045 $\pm$ 0.0320 & 0.3276 $\pm$ 0.0252 & 0.0999 $\pm$ 0.0000 & 0.0999 $\pm$ 0.0000\\ 
BYOL                      & 01 & 0.20  & 0.5155 $\pm$ 0.0183 & 0.5447 $\pm$ 0.0147 & 0.6462 $\pm$ 0.0119 & 0.0325 $\pm$ 0.0153 & 0.4947 $\pm$ 0.0129 & 0.5049 $\pm$ 0.0110 & 0.9723 $\pm$ 0.0755 & 0.1473 $\pm$ 0.0319 & 0.3911 $\pm$ 0.0087 & 0.1999 $\pm$ 0.0000 & 0.1999 $\pm$ 0.0000\\ 
BYOL                      & 01 & 0.50  & 0.5251 $\pm$ 0.0069 & 0.5517 $\pm$ 0.0069 & 0.6029 $\pm$ 0.0050 & 0.0183 $\pm$ 0.0064 & 0.4963 $\pm$ 0.0130 & 0.5044 $\pm$ 0.0154 & 0.8811 $\pm$ 0.0514 & 0.1177 $\pm$ 0.0204 & 0.4364 $\pm$ 0.0096 & 0.5000 $\pm$ 0.0000 & 0.5000 $\pm$ 0.0000\\ 
BYOL                      & 01 & 1.00  & 0.5687 $\pm$ 0.0235 & 0.5851 $\pm$ 0.0191 & 0.5770 $\pm$ 0.0283 & 0.0449 $\pm$ 0.0143 & 0.5102 $\pm$ 0.0251 & 0.5162 $\pm$ 0.0244 & 0.8958 $\pm$ 0.0457 & 0.1285 $\pm$ 0.0111 & 0.4638 $\pm$ 0.0156 & 1.0000 $\pm$ 0.0000 & 1.0000 $\pm$ 0.0000\\ 
DivideMix                 & 01 & 0.10  & 0.6410 $\pm$ 0.0465 & 0.6790 $\pm$ 0.0306 & 0.5851 $\pm$ 0.0299 & 0.0812 $\pm$ 0.0315 & 0.6764 $\pm$ 0.0361 & 0.7057 $\pm$ 0.0252 & 0.1183 $\pm$ 0.0185 & 0.0901 $\pm$ 0.0074 & 0.6584 $\pm$ 0.0384 & 0.0999 $\pm$ 0.0000 & 0.0999 $\pm$ 0.0000\\
\end{tabular}
}
\caption{Results for dataset Treeversity\#6}
\end{table}

\begin{table}[t]
\resizebox{\textwidth}{!}{
\begin{tabular}{l c c c c c c c c c c c c c}
Method                    & $m$ & $n$  &  $F1$                &  $ACC$               &  $KL$                &  $ECE$               &  $\hat{F1}$          &  $\hat{ACC}$         &  $\hat{KL}$          &  $\hat{ECE}$         &  $\kappa$            &  $b$                 &  $\hat{b}$          \\ 
 \midrule \\ 
BYOL                      & 01 & 0.10  & 0.6602 $\pm$ 0.0913 & 0.6356 $\pm$ 0.0857 & 0.3303 $\pm$ 0.0194 & 0.0556 $\pm$ 0.0200 & 0.5386 $\pm$ 0.0936 & 0.5193 $\pm$ 0.0830 & 0.5542 $\pm$ 0.1467 & 0.0813 $\pm$ 0.0367 & 0.4975 $\pm$ 0.0426 & 0.0999 $\pm$ 0.0000 & 0.0999 $\pm$ 0.0000\\ 
BYOL                      & 01 & 0.20  & 0.5243 $\pm$ 0.0596 & 0.5225 $\pm$ 0.0466 & 0.4098 $\pm$ 0.0937 & 0.0724 $\pm$ 0.0137 & 0.4959 $\pm$ 0.0388 & 0.4877 $\pm$ 0.0334 & 0.5160 $\pm$ 0.1512 & 0.0892 $\pm$ 0.0448 & 0.4644 $\pm$ 0.0588 & 0.1999 $\pm$ 0.0000 & 0.1999 $\pm$ 0.0000\\ 
BYOL                      & 01 & 0.50  & 0.5913 $\pm$ 0.0084 & 0.6047 $\pm$ 0.0096 & 0.3375 $\pm$ 0.0387 & 0.0489 $\pm$ 0.0148 & 0.5597 $\pm$ 0.0370 & 0.5525 $\pm$ 0.0283 & 0.4688 $\pm$ 0.0942 & 0.0669 $\pm$ 0.0316 & 0.4690 $\pm$ 0.0436 & 0.5000 $\pm$ 0.0000 & 0.5000 $\pm$ 0.0000\\ 
BYOL                      & 01 & 1.00  & 0.6256 $\pm$ 0.0191 & 0.6245 $\pm$ 0.0445 & 0.3369 $\pm$ 0.0400 & 0.0224 $\pm$ 0.0064 & 0.5683 $\pm$ 0.0287 & 0.5635 $\pm$ 0.0294 & 0.5758 $\pm$ 0.0768 & 0.1033 $\pm$ 0.0238 & 0.4679 $\pm$ 0.0535 & 1.0000 $\pm$ 0.0000 & 1.0000 $\pm$ 0.0000\\ 
DivideMix                 & 01 & 0.10  & 0.4193 $\pm$ 0.1057 & 0.4330 $\pm$ 0.0887 & 0.5163 $\pm$ 0.0954 & 0.1386 $\pm$ 0.0165 & 0.4275 $\pm$ 0.0843 & 0.4396 $\pm$ 0.0728 & 0.0590 $\pm$ 0.0396 & 0.0290 $\pm$ 0.0131 & 0.5994 $\pm$ 0.0708 & 0.0999 $\pm$ 0.0000 & 0.0999 $\pm$ 0.0000\\
\end{tabular}
}
\caption{Results for dataset Turkey}
\end{table}

\end{document}